%% file: main.tex
\pdfoutput=1
\documentclass[10pt,journal,compsoc]{IEEEtran} %

\usepackage{cite}
\usepackage{amsmath,amssymb,amsfonts}

\usepackage{booktabs}

\usepackage{siunitx}
\DeclareSIUnit{\nothing}{\relax}
\usepackage{algpseudocode,algorithm}
\usepackage{xcolor}
\usepackage{array}

\usepackage{graphicx}
\usepackage{grffile} %
\usepackage{float}
\usepackage{caption}
\usepackage{subcaption}
\usepackage{makecell}
\newcolumntype{P}[1]{>{\centering\arraybackslash}p{#1}}
\newcolumntype{M}[1]{>{\centering\arraybackslash}m{#1}}
\usepackage{adjustbox} %
\usepackage{enumitem} %
\usepackage{multirow}
\usepackage{mathrsfs}
\usepackage{empheq}
\usepackage{rotating}
\usepackage{soul}

\usepackage{etoolbox} %

\usepackage{url}
\def\MYTITLE{Multi-Event-Camera Depth Estimation and Outlier Rejection by Refocused Events Fusion}

\newcommand\MYhyperrefoptions{bookmarks=true,bookmarksnumbered=true,
pdfpagemode={UseOutlines},plainpages=false,pdfpagelabels=true,
colorlinks=true,breaklinks=true,
pdftitle={\MYTITLE},%
pdfsubject={Computer Vision, Depth Estimation, Robotics},%
pdfauthor={S. Ghosh, G. Gallego},%
pdfkeywords={Event camera, Asynchronous sensor, 3D Reconstruction, Space Sweep, Stereo}}%
\usepackage[\MYhyperrefoptions,pdftex]{hyperref}

\usepackage[capitalize]{cleveref}
\newif\ifaisy
\aisytrue %
\ifaisy
\crefname{section}{Section}{Sections}
\crefname{table}{Table}{Tables}
\crefname{figure}{Figure}{Figures}
\else 
\crefname{section}{Sec.}{Secs.}
\crefname{table}{Tab.}{Tabs.} 
\crefname{figure}{Fig.}{Figs.}
\fi
\Crefname{section}{Section}{Sections}
\Crefname{table}{Table}{Tables}
\Crefname{figure}{Figure}{Figures}

\pdfpxdimen=\dimexpr 1 in/72\relax
\usepackage{calc}

\newif\ifclearsectionlook
\clearsectionlookfalse

\input{chapters/macros}

\begin{document}

\title{\MYTITLE}

\author{Suman Ghosh and Guillermo Gallego%
\IEEEcompsocitemizethanks{\IEEEcompsocthanksitem S. Ghosh and G. Gallego are with the Technische Universität Berlin, Berlin, Germany.
G. Gallego is with the Einstein Center Digital Future and the Science of Intelligence Excellence Cluster, Berlin, Germany.
\IEEEcompsocthanksitem Preprint of paper accepted at Advanced Intelligent Systems, 2022.
\hfil\break doi: 10.1002/aisy.202200221.
}%
}

\IEEEtitleabstractindextext{%
\input{chapters/00_abstract}

}

\maketitle

\IEEEdisplaynontitleabstractindextext

\IEEEpeerreviewmaketitle

\input{chapters/01_introduction}
\input{chapters/02_related_work}

\input{chapters/03_method}

\input{chapters/04_experiments}

\input{chapters/05_conclusion}

\vspace{-1ex}
\appendix

\input{chapters/07_suppl_mat}

\bibliographystyle{IEEEtran}
\input{main.bbl}

\input{chapters/biographies}

\end{document}

%% file: chapters/macros.tex
\def\Lum{L}
\def\pol{p} %
\def\cE{\mathcal{E}} %
\def\numEvents{N_e} %
\def\Warp{\mathbf{W}}

\def\bx{\mathbf{x}}
\def\be{\mathbf{e}}

\def\bt{\mathbf{t}}

\def\mP{\mathtt{P}}
\def\Rot{\mathtt{R}}

\def\bzero{\mathbf{0}}

\def\bX{\mathbf{X}}

\def\mId{\mathtt{I}}
\def\mH{\mathtt{H}}

\def\GTS{GTS}
\def\rpgreader{\emph{rpg\_reader}}

\def\rpgbox{\emph{rpg\_box}}

\def\rpgmonitor{\emph{rpg\_monitor}}
\def\upennflyOne{\emph{upenn\_flying1}}

\def\upennflyThree{\emph{upenn\_flying3}}

\robustify \bfseries
\robustify \underline
\newcommand{\unum}[1]{\underline{\num{#1}}}
\newcommand{\bnum}[1]{\bfseries #1}

\definecolor{light-gray}{gray}{0.75}
\newcommand\gframe[1]{\color{light-gray}\frame{#1}}

%% file: chapters/00_abstract.tex
\begin{abstract}
Event cameras are bio-inspired sensors that offer advantages over traditional cameras. 
They operate asynchronously, sampling the scene at microsecond resolution and producing a stream of brightness changes.
This unconventional output has sparked novel computer vision methods to unlock the camera's potential.
Here, the problem of event-based stereo 3D reconstruction for SLAM is considered.
Most event-based stereo methods attempt to exploit the high temporal resolution of the camera and the simultaneity of events across cameras to establish matches and estimate depth.
By contrast, this work investigates how to estimate depth without explicit data association by fusing Disparity Space Images (DSIs) originated in efficient monocular methods. 
Fusion theory is developed and applied to design multi-camera 3D reconstruction algorithms that produce state-of-the-art results, as confirmed by comparisons with four baseline methods and tests on a variety of available datasets.
\end{abstract}

%% file: chapters/01_introduction.tex
\section*{Video and Code}
Project page:  
\url{https://github.com/tub-rip/dvs_mcemvs}

\section{Introduction}
\label{sec:intro}

Intelligent navigation in our complex 3D world relies on robust and efficient visual perception, which is challenging for autonomous robots. However, humans use vision very efficiently to navigate 3D environments, even in novel scenarios.
Inspired by human vision, neuromorphic spike-based sensing and processing has been recently investigated for 
robot vision \cite{Berco2021,Gallego20pami} and retinal implants \cite{Berco2019}.

Event cameras, such as the Dynamic Vision Sensor \cite{Lichtsteiner08ssc,Finateu20isscc,Suh20iscas} (DVS), are neuromorphic sensors that acquire visual information very differently from traditional cameras. 
They sample the scene asynchronously, producing a stream of spikes, called ``events'', that encode the time, location and sign of per-pixel brightness changes.
Event cameras possess outstanding properties compared to traditional cameras:
very high dynamic range (HDR), high temporal resolution ($\approx$ \si{\micro\second}), 
temporal redundancy suppression and low power consumption.
These properties offer potential to tackle challenging scenarios for standard cameras (high speed and/or HDR) \cite{Delbruck13fns,Matsuda15iccp,Amir17cvpr,Rebecq18ijcv,Wang2020}.
However, this calls for novel methods to process the unconventional output of event cameras in order to unlock their capabilities~\cite{Gallego20pami}.

In this work, we tackle the problem of event-based stereo 3D reconstruction for Visual Odometry (VO) and Simultaneous Localization and Mapping (SLAM).
An efficient SLAM system is critical for the navigation of autonomous intelligent agents (like field robots) in challenging unstructured environments, especially in extreme ones like offshore drilling, nuclear power plants, etc. \cite{Sayed2021}. 
An event-based SLAM system has the potential to efficiently overcome difficult scenarios in these applications \cite{Rebecq17ral,Kim16eccv,Zhou20tro}.
For example the HDR advantages of event cameras translate into high-fidelity depth maps in difficult lighting conditions, as demonstrated by a broad variety of works in VO/SLAM \cite[Fig.~12]{Rebecq17ral}, \cite[Fig.~8]{Kim16eccv}, \cite[Fig.~15]{Zhou20tro}.

\input{floats/fig_summary}

Our work is inspired by EVO~\cite{Rebecq17ral}, which is %
the state of the art in event-based monocular VO. 
The effectiveness of EVO is largely due to its mapping module, Event-based Multi-View Stereo (EMVS)~\cite{Rebecq18ijcv}, 
which enables 3D reconstruction without the need to recover image intensity, 
without having to explicitly solve for data association between events, 
and without the need of a GPU (it is fast on a standard CPU --e.g., speed of \SI{1.2}{\mega ev/\second/core} \cite{Rebecq18ijcv}). 
Additionally, EMVS admits an interpretation in terms of event refocusing %
or event alignment (contrast maximization)~\cite{Gallego18cvpr}, 
which is the state of the art framework to tackle other vision problems~\cite{Gallego19cvpr,Stoffregen19cvpr,Stoffregen19iccv,Wang20cvpr,Nunes21pami,Peng21pami,Shiba22eccv,Paredes21neurips,Shiba22sensors}.
Our goal is to extend EMVS to the multi-camera setting (i.e., two or more event cameras in a multi-view configuration sharing a common clock), and in particular to the stereo setting, 
in order to benefit from these advantages and connections (\cref{fig:summary:quadrants}). 
In the process, we revisit the event simultaneity assumption used in stereo depth estimation and develop a theory of fusion of refocused events, which could be useful in other problems, such as feature or camera tracking \cite{Seok20wacv}.

In summary, our contributions are:
\begin{itemize}
    \item Simple, efficient and extensible solutions to the problem of event-based stereo 3D reconstruction for SLAM using a correspondence-free approach.
    We investigate early event data fusion strategies in \emph{two orthogonal directions}: 
    between cameras (``spatial stereo'', \cref{sec:method:fusion_across_cameras}) and along time (``temporal stereo'', \cref{sec:method:fusion_time}).
    
    \item The investigation of several functions to fuse refocused events (using e.g., Generalized means, \cref{sec:method:fusion}) and its application to the two mentioned directions.
    
    \item A comprehensive experimental evaluation on five publicly available datasets and comparing against several baseline methods, producing state-of-the-art results (\cref{sec:experim}).
    We also show how the method can naturally handle multi-event-camera setups with \emph{linear} complexity.

\end{itemize}

This research aims at developing robust multi-camera visual perception systems for the navigation of artificial intelligent systems in challenging environments, like stereo depth perception for SLAM and attention in robots \cite{Gao2021,Ghosh123}. 

%% file: floats/fig_summary.tex
\global\long\def\figWidth{0.49\columnwidth}
\begin{figure}[t]
	\centering
    {\small
    \setlength{\tabcolsep}{1pt}
	\begin{tabular}{
	>{\centering\arraybackslash}m{\figWidth} 
	>{\centering\arraybackslash}m{\figWidth}}
		Monocular & Stereo (Ours) \\\addlinespace[1ex]
		\includegraphics[width=\linewidth, trim={2.6cm 3cm 2.5cm 2cm},clip]{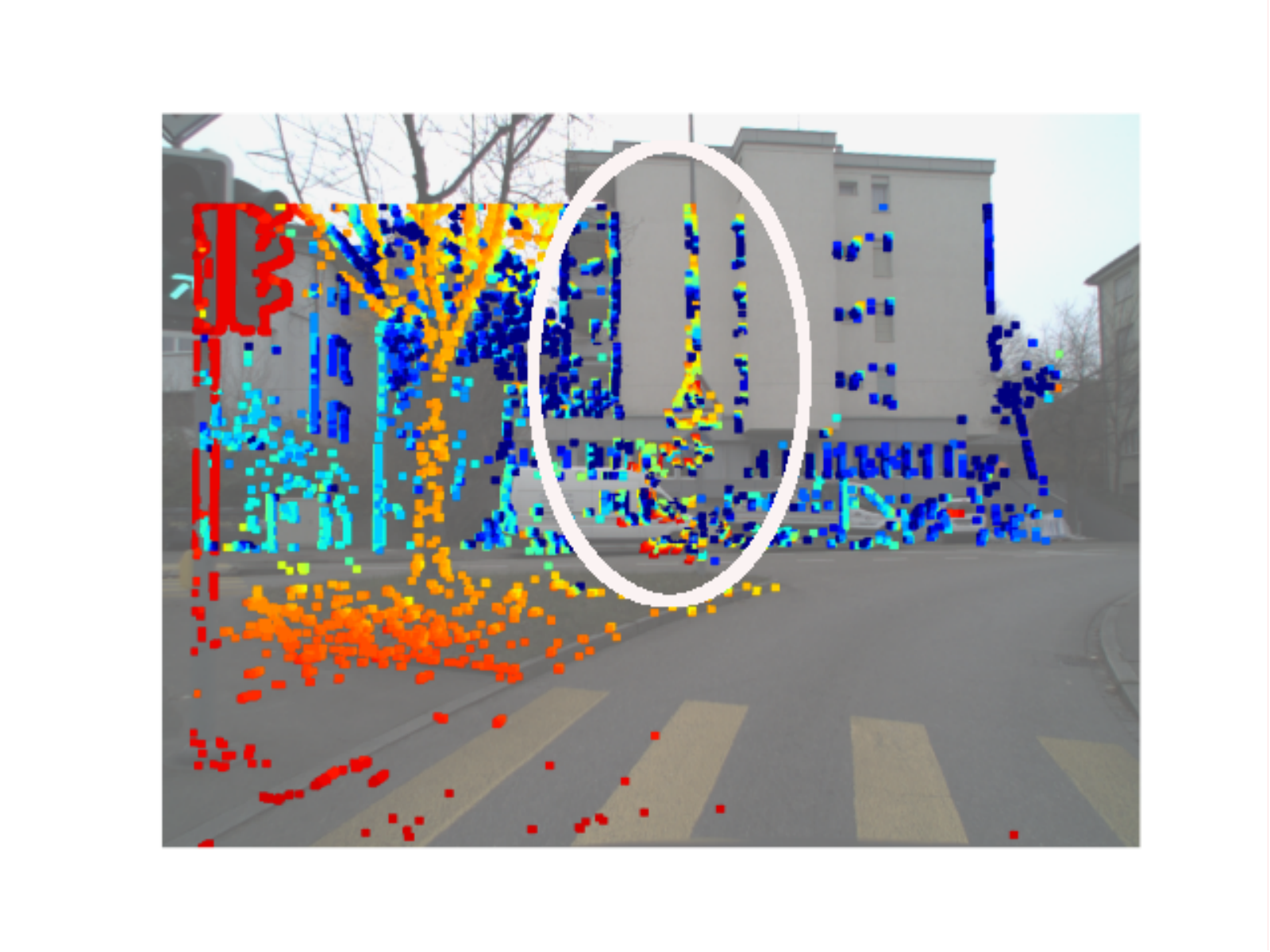}
		&\includegraphics[width=\linewidth, trim={2.6cm 3cm 2.5cm 2cm},clip]{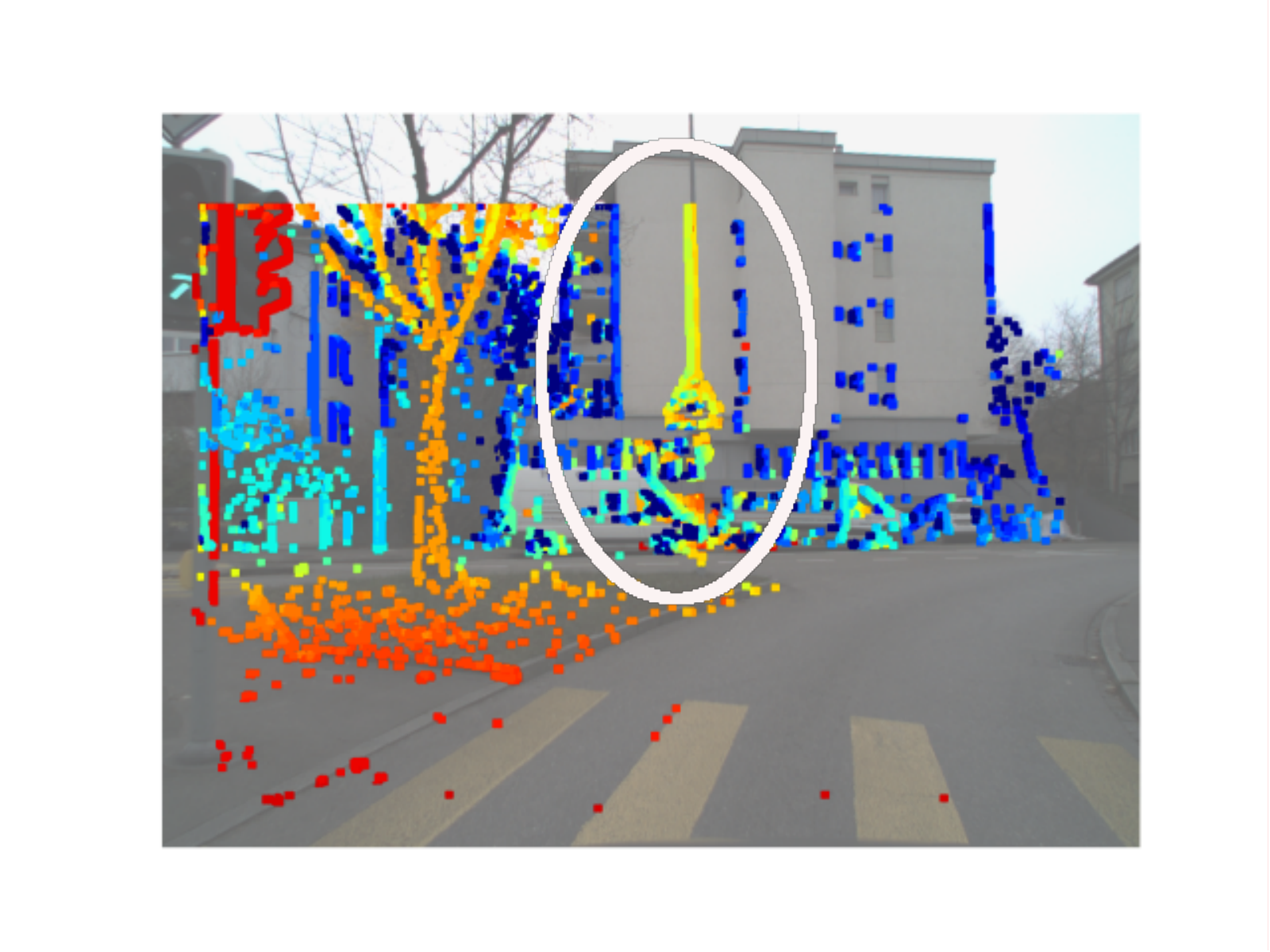}
		\\
	\end{tabular}
	}
	\caption{\label{fig:summary:quadrants}
	Semi-dense depth maps estimated by event-based monocular~\cite{Rebecq18ijcv} and stereo methods.
	Stereo is beneficial for more accurate estimation 
	and outlier removal compared to monocular depth estimation (e.g., traffic sign in the center).
	Depth is pseudo-colored, from red (close) to blue (far).
	Color frames are only shown for visualization.
	Data from \cite{Gehrig21ral}.
	}
\end{figure}

%% file: chapters/02_related_work.tex
\section{Related Work}
\label{sec:relatedwork}

\textbf{What}: Stereo depth estimation using event cameras has been an interesting problem %
ever since the first event camera
was invented by Mahowald and Mead in the 1990s~\cite{Mahowald92thesis}.
As such, they simultaneously designed a stereo chip~\cite{Mahowald94book} to implement Marr and Poggio's cooperative stereo algorithm \cite{Marr76Science}. %
This approach has inspired a lot of literature that focuses on 3D reconstruction over short time intervals (``instantaneous stereo'') \cite{Piatkowska13iccvw,Firouzi16npl,Osswald17srep}. 
These methods work well with stationary cameras in uncluttered scenes (where events are caused only by few moving objects), thus enabling 3D reconstruction of sparse, dynamic scenes.
For a detailed survey on these stereo methods we refer to \cite{Steffen19fnbot,Furmonas22sensors}.
In contrast, stereo event-based 3D reconstruction for VO/SLAM has been addressed recently \cite{Zhou18eccv,Zhou20tro}. 
It assumes a static world and known camera motion (e.g., from a tracking method) to assimilate events over longer time intervals, 
so as to increase parallax and produce more accurate semi-dense depth maps.
Some other works estimate depth by combining an event camera with other devices, such as light projectors \cite{Matsuda15iccp,Martel18iscas,Muglikar21threedv} or a motorized focal lens \cite{Haessig19srep}, which are different from our hardware setup and application.

\textbf{How}: 
Depth estimation with stereo event cameras is predominantly based on exploiting the epipolar constraint %
and the assumption of \emph{temporal coincidence} of events across retinas, namely that a moving object produces events of same timestamps on both cameras \cite{Carneiro13nn,Martel18iscas,Ieng18fnins}.
This aims at exploiting the high temporal resolution and redundancy suppression of event cameras to establish event matches across image planes and then triangulate.
It is also known as \emph{event simultaneity} or \emph{temporal consistency} \cite{Zhou18eccv}, and it is analogous to photometric consistency in traditional cameras. 
This assumption does not strictly hold \cite{Benosman12nn,Piatkowska14msci}, and so it is relaxed to account for temporal noise (jitter and delay).
Essentially event simultaneity is exploited to solve the data association problem (establishing event matches), which is a well-known difficult problem due to the little information carried by each event and their dependency with motion direction (changing ``appearance'' of events~\cite{Gehrig19ijcv,Gallego20pami}). 

The above ideas are used in the mapping module of~\cite{Zhou20tro}, the state-of-the-art stereo 3D reconstruction method for VO/SLAM. 
In this method temporal consistency is measured across space-time neighborhoods of events by first converting the events into time surfaces (TSs) \cite{Lagorce17pami} and then comparing their spatial neighborhoods. 
Stereo point matches are established and provide depth estimates which are fused in a probabilistic way using multiple TSs to produce a more accurate semi-dense inverse depth map.
\input{floats/fig_dsi_monitor}
\input{floats/fig_pipeline}

In contrast, we investigate \emph{a new way of doing stereo, without explicitly using event simultaneity} 
and hence without establishing event matches.
Therefore, to the best of our knowledge, we completely depart from previous event-based \emph{stereo} methods.
Correspondence-free approaches for depth estimation have been proposed for frame-based cameras \cite{Collins96cvpr,Yuan2003,Lehmann2007}. 
For example, \cite{Lehmann2007} proposes a method to estimate affine fundamental matrices without explicit correspondences, but it does not solve the full problem of depth estimation. 
On the other hand, \cite{Yuan2003} aims to solve the problem of stereo depth estimation, but for the special case of planar or quasi-planar scenes. 
It is an extension of \cite{Aloimonos1990} for arbitrary camera positions. 
In contrast, our method solves the full depth estimation problem and is not limited to the planar case, 
following the seminal idea of \cite{Collins96cvpr} to sweep space and exploit the sparsity of scene edges.
Inspired by \cite{Rebecq18ijcv,Collins96cvpr}, we circumvent the data association task by leveraging the sparsity of events (event cameras naturally highlight edges, which are sparse, in hardware) 
and by exploiting the continuous set of camera viewpoints at which events are available.
This provides a rich collection of back-projected rays through the events to estimate scene structure. 
Our contributions pertain to the processing (e.g., fusion) of such back-projected rays or ``refocused events'', 
which has not been considered before (since \cite{Rebecq18ijcv} and newer approaches \cite{Cho21ral} do not consider data fusion, e.g., across cameras).

%% file: floats/fig_dsi_monitor.tex
\begin{figure}[t]
\centering
    \includegraphics[width=.8\linewidth]{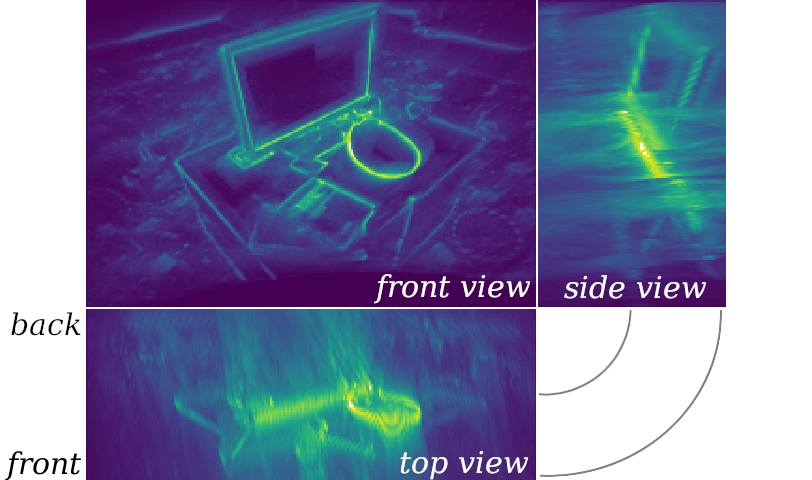}    
    \caption{\emph{DSI projections} of a non-planar scene (\rpgmonitor{} data from \cite{Zhou18eccv}, also in \cref{fig:output:stereo}).
    Disparity Space Image (DSI) values (i.e., ray counts) are pseudo-colored, from blue (low) to yellow (high).
    The DSI has dimensions $w\times h \times N_Z$, 
    according to the resolution of the event camera DAVIS240 ($w=240, h=180$ pix) and the number of \emph{inverse-depth} planes used, $N_Z = 100$. 
    The three DSI $\max$-projections are: 
    front view (top left, $240 \times 180$ pix), top view (bottom left, $240 \times 100$ pix) and side view (top right, $100 \times 180$ pix).
    }
    \label{fig:dsiproj:monitor}
\end{figure}

%% file: floats/fig_pipeline.tex
\begin{figure*}[t]
\centering
    {\includegraphics[trim={0 11.2cm 0 0.7cm},clip,width=0.9\linewidth]{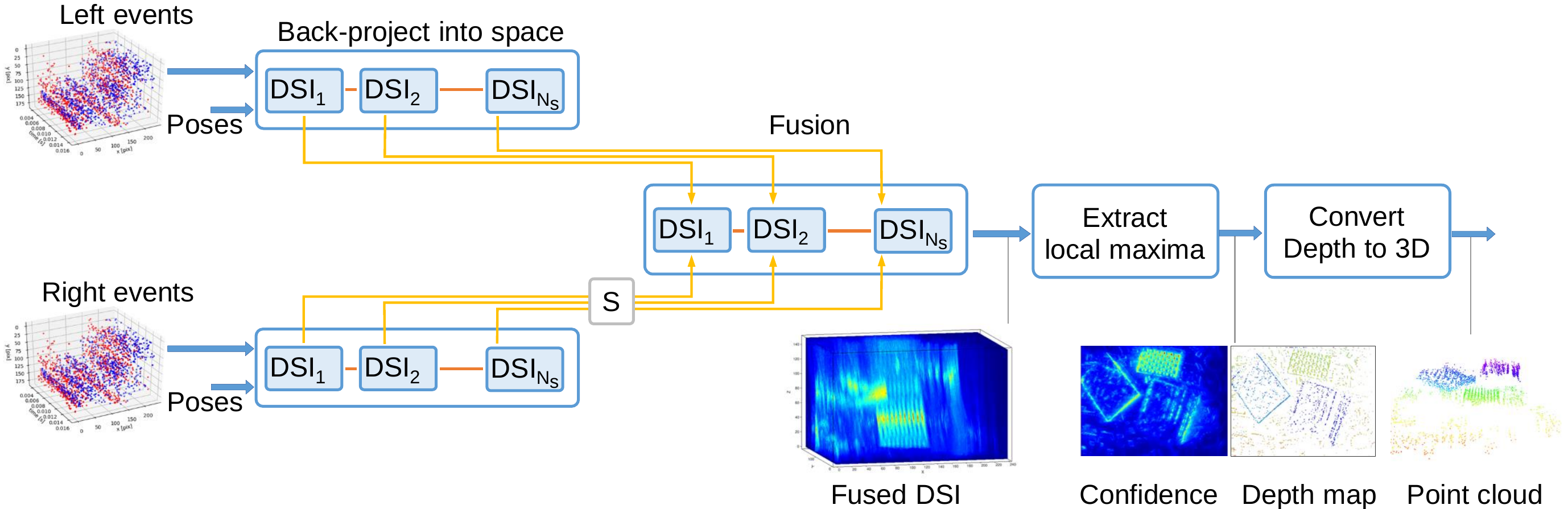}}
    \caption{Our method takes as input the events from two or more synchronized, rigidly attached event cameras and their poses, and estimates the scene depth.
    Using Space Sweeping, it builds ray density Disparity Space Images (DSIs) from each camera data and fuses them into one DSI (\cref{sec:method:fusion}), from which the 3D structure of the scene is extracted in the form of a semi-dense depth map, which may be cast into a point could. 
    Fusion across cameras (\cref{sec:method:fusion_across_cameras}) is represented with yellow lines,
    and temporal fusion (\cref{sec:method:fusion_time}) with red lines. 
    The optional shuffling block (``S'') is presented in \cref{sec:method:shuffling}.
    }
    \label{fig:block-diagram}
\end{figure*}

%% file: chapters/03_method.tex
\section{Event-based Stereo Depth Estimation}
\label{sec:method:stereodepth}

This section reviews how an event camera works (\cref{sec:method:eventcam}) 
and the monocular method EMVS (\cref{sec:method:emvs})
before presenting our stereo depth estimation approach.
Two main event fusion directions are presented: 
fusion of camera views (\cref{sec:method:fusion_across_cameras}) 
using one of several functions (\cref{sec:method:fusion}),
and fusion of multiple time intervals (\cref{sec:method:fusion_time}).
Then, we revisit the event simultaneity assumption (\cref{sec:method:shuffling}) and analyze the computational complexity of the approach (\cref{sec:complexity}).

\subsection{How an Event Camera Works}
\label{sec:method:eventcam}
Event cameras, such as the Dynamic Vision Sensor (DVS) \cite{Lichtsteiner08ssc}, are bio-inspired sensors that capture pixel-wise \emph{brightness changes}, called events, instead of brightness images. 
An event $e_k \doteq (\bx_k, t_k, \pol_{k})$ is triggered when the logarithmic brightness $\Lum$ at a pixel exceeds a contrast sensitivity $\theta>0$, 
\begin{equation}
\label{eq:generativeEventCondition}
\Lum(\bx_k,t_k) - \Lum(\bx_k, t_k-\Delta t_k) = \pol_k \, \theta,
\end{equation} 
where $\bx_k\doteq (x_k, y_k)^{\top}$, $t_k$ (in \si{\micro\second}) and $\pol_{k} \in \{+1,-1\}$
are the spatio-temporal coordinates and polarity of the brightness change, respectively,
and $t_k-\Delta t_k$ is the time of the previous event at the same pixel $\bx_k$.
Hence, each pixel has its own sampling rate, which depends on the visual input. 
Assuming constant illumination, pixels produce events proportionally to the amount of scene motion and texture.

\subsection{EMVS: Monocular 3D Reconstruction}
\label{sec:method:emvs}

The problem of \emph{monocular} depth estimation with an event camera consists of estimating the 3D structure of the scene given the events and the camera poses (i.e., position and orientation) as the sensor moves through the scene. 
The method in~\cite{Rebecq18ijcv} solves this problem, called EMVS, in two main steps: 
it builds a Disparity Space Image (DSI) using a space sweeping approach~\cite{Collins96cvpr} and then detects local maxima of the DSI.
The key idea is that, as the camera moves, events are triggered at an almost continuous set of viewpoints, which are used to back-project events into space in the form of rays (a DSI). 
The local maxima of the ray density (where many rays intersect, as shown in \cref{fig:dsiproj:monitor}) are candidate locations for the 3D edges that produce the events.
Specifically, the DSI is discretized on a projective voxel grid defined at a reference view, 
and local maxima are detected along viewing rays, thus producing a semi-dense depth map. %
Events are processed in packets %
of about \SIrange{0.2}{1}{\mega~events}.
The key benefits of EMVS are its simplicity, accuracy, efficiency (real-time, with $\sim$1.2Mev/s throughput per CPU core \cite{Rebecq18ijcv}) and that it estimates depth without explicit data association. %

\subsection{Fusion Across Cameras}
\label{sec:method:fusion_across_cameras}

We consider the problem of depth estimation from two synchronized and calibrated event cameras rigidly attached. 
Hence, the input consists of a stereo event stream and poses, and the desired output is a (semi-dense) depth map or, equivalently, a 3D point cloud with the scene structure (\cref{fig:block-diagram}).

\input{floats/alg_fusion_cameras}
\input{floats/fig_mono_vs_stereo}

\input{floats/fig_show_output}

\subsubsection{Challenges and Proposed Architecture}
A naive solution to the problem consists of running two instances of EMVS, one per camera, and fusing the resulting point clouds into a single one, including post-processing to mitigate redundant 3D points.
This is a late-fusion approach, which greatly ignores the benefits that arise from having two cameras observing the same scene.
By contrast, we seek to perform fusion earlier in the processing pipeline: at the DSI stage. %
Hence the first technical challenge is to define the DSI. 
EMVS defines a DSI per camera, located at a reference view (RV) along the camera's trajectory. 
However, the fusion of DSIs at two different RVs is prone to resampling errors.
Thus it is key to define a common DSI location for both cameras.

The second challenge is to investigate sensible fusion strategies.
Our approach includes, as a particular case, that of back-projecting the events from both cameras into a single DSI and simply counting rays.
This is equivalent to the scenario of a single event camera that moves twice through the scene, with different motions, but uses the same DSI to aggregate rays.
It doubles the ray count in the DSI, %
but summation discards valuable information for fusion, 
such as how many rays originate in each camera: 
given 8 rays at a point, it is preferable to have 4 rays from each camera than an unbalanced situation (a 3D edge seen only by one camera).

To deal with the above challenges we define two DSIs at a common reference view:
having one DSI per camera allows us to preserve the origin of the event data, 
and having \emph{geometrically aligned} DSIs avoids resampling errors during fusion.
Without loss of generality, let the RV be a point along the trajectory of the left camera.
We investigate how to compare and fuse the ray densities from each event camera.
\Cref{fig:block-diagram} shows the block diagram of our stereo approach.
For now, assume there is $N_s=1$ DSI per camera ($N_s>1$ is presented in \cref{sec:method:fusion_time}).
First, the aligned DSIs are populated with back-projected events from each camera, 
then they are \emph{fused (combined)} into a single one (e.g., using a voxel-wise harmonic mean or other similarity score (\cref{sec:method:fusion})), 
and finally local maxima are extracted to produce a semi-dense depth map.
The steps are specified in Alg.~\ref{alg:fusion:stereo}.

\subsubsection{Intuitive Example}
To illustrate key differences between EMVS (monocular) and the stereo Alg.~\ref{alg:fusion:stereo} we use a sequence acquired with two event cameras~\cite{Brandli14ssc} performing a 1D motion (translation along the $X$ axis, using a linear slider).
As input to EMVS we use the data from the left camera.
\Cref{fig:experim:monovsstereo} shows the evolution of the DSIs as time progresses, i.e., as the camera rig moves and more events are acquired and back-projected onto the DSIs (and fused in the stereo case). 
We plot projections of the DSI along its three coordinate axes (like in \cref{fig:dsiproj:monitor}) at the reference viewpoint RV.

As \cref{fig:experim:monovsstereo} shows, the stereo DSI (bottom row) \emph{converges faster} to the 3D structure of the scene than the monocular DSI (top row).
This is specially noticeable in the top views: 
only one set of nearly parallel rays, poor for triangulation, is visible in the monocular case. 
By contrast, the top views of the stereo DSIs show two sets of rays, one from each camera: 
the rays from the left camera are nearly straight, whereas the rays from the right camera are curved due to inverse depth parametrization of the DSI grid and the fact that the DSI is projective.
In both scenarios, the rays intersect at multiple voxels and as time progresses the true intersection locations dominate over others, i.e., the 3D structure emerges by event refocusing \cite{Rebecq18ijcv,Gallego19cvpr}.
Additionally, \emph{in the stereo case refocusing is combined with the proposed fusion functions (\cref{sec:method:fusion}) to speed-up the emergence} and better highlight 3D structure.
In Alg.~\ref{alg:fusion:stereo} this is achieved by the harmonic mean, which deemphasizes the non-intersecting parts of the rays. %

\subsubsection{Output of the Stereo Method}
\Cref{fig:output:stereo} shows the output of Alg.~\ref{alg:fusion:stereo} on two scenes.
After DSI fusion, Alg.~\ref{alg:fusion:stereo} extracts a depth map by locating the DSI maxima along each viewing ray (through RV pixel $\bx = (x,y)^\top$).
Letting $f:\mathbb{R}^3 \to \mathbb{R}_{\geq 0}$ be the fused DSI, 
its maxima provide the confidence or ``contrast'' map $c(\bx) = f(\bx,Z^\star(\bx))$ and the depth map $Z^\star(\bx)$. 
Adaptive Gaussian Thresholding (AGT) selects the pixels with highest local value, thus making the depth maps semi-dense.
A median filter is applied to remove isolated points.
The front-view projection of the DSI in \cref{fig:dsiproj:monitor} corresponds to the confidence map, 
which is called this way because it is used in AGT to ``select the most confident pixels in the depth map'' \cite{Rebecq18ijcv} 
(since voxels with many ray intersections are more likely to capture true 3D points than voxels with few ray intersections).

\subsection{DSI Fusion Functions}
\label{sec:method:fusion}
DSI fusion is the central part of our method (\cref{fig:block-diagram}).
It takes two ray density DSIs on the same region of space as input (one per camera) and produces a merged DSI, which is then used to extract depth information (candidate locations of 3D edges).
So, \emph{what are sensible ways to compare two DSIs?}
FMRay density DSIs have very different statistics from natural images, hence standard similarity metrics for image patches may not be the most appropriate ones \cite{Deledalle12ijcv}.

Formally, let $\cE_l=\{e^l_{k}\}_{k=1}^{\numEvents^l}$ and $\cE_r=\{e^r_{k}\}_{k=1}^{\numEvents^l}$ be stereo events over some time interval $[0,T]$, 
and ${f}_l,{f}_r:V\subset\mathbb{R}^3 \to \mathbb{R}_{\geq 0}$ be the ray densities (DSIs) defined over a volume $V$:
\begin{equation}
\label{eq:LeftDSI}
{f}_l(\bX) = \sum_{k=1}^{\numEvents^l}\delta\bigl(\bX-\bX'_k(e^l_{k})\bigr),
\end{equation}
where $\bX'_k(e^l_{k}) = (\bx^{l\prime \top}_k,Z)^\top$ is a 3D point on the back-projected ray through event $e^l_{k}$, at depth $Z$ with respect to the reference view (RV). 
Events are transferred to RV using the continuous motion of the cameras and candidate depth values $Z\in [Z_{\min},Z_{\max}]$:
\begin{equation}
\label{eq:WarpThatTransfersPointsTwoCameras}
\bx^{l\prime}_k = \Warp\bigl(e^l_{k}, \mP^l(t_k),\mP_{v}, Z \bigr),
\end{equation}
where $\mP^l(t)$ is the pose of the left event camera at time $t$ 
and $\mP_v$ is the pose of RV.
The warp $\Warp$ corresponds to the planar homography induced by a plane parallel to the image plane of RV and at the given depth $Z$.
In a coordinate system adapted to RV (i.e., $\mP_v=(\mId, \bzero)$), the planar homography is given by the $3\times 3$ homogeneous matrix
\begin{equation}
\mH_\Warp \sim \bigl(\Rot + \frac1{Z}\bt \be_3^\top\bigr)^{-1},
\end{equation}
where $\mP^l(t_k)=(\Rot,\bt)$ and $\be_3=(0,0,1)^\top$.
A similar formula applies to compute ${f}_r$ from $\cE_r$ and the corresponding camera poses.
In practice, DSIs are discretized over a projective voxel grid with $N_Z$ depth planes in $[Z_{\min},Z_{\max}]$, 
and the Delta $\delta$ in \eqref{eq:LeftDSI} is approximated by bilinear voting \cite{Gallego17ral,Rebecq18ijcv}.
Hence, each voxel counts the number of event rays that pass through it.

Next, letting $u=f_l(\bX)$, $v=f_r(\bX)$ be the values of the DSIs at a 3D point $\bX$, we seek to define a fused value $g(\bX)$. %
For simplicity, we consider metrics operating in a point-wise (i.e., voxel-wise) manner, i.e., with a slight abuse of notation:
\begin{equation}
\label{eq:DSIfusion}
g(\bX) \equiv g\bigl(f_l(\bX),f_r(\bX) \bigr) = g(u,v).
\end{equation}

The metrics considered are the following:
\begin{align}
    A(u,v) & \doteq (u+v)/2 & \text{Arithmetic mean}\label{eq:fusionfunc:A}\\
    G(u,v) & \doteq \sqrt{uv} & \text{Geometric mean}\label{eq:fusionfunc:G}\\
    H(u,v) & \doteq 2/(u^{-1}+v^{-1}) & \text{Harmonic mean}\label{eq:fusionfunc:H}\\
    \text{RMS}(u,v) & \textstyle \doteq \sqrt{\frac1{2}(u^2+v^2)} & \text{Quadratic mean}\label{eq:fusionfunc:RMS}\\
    \min (u,v) & & \text{Minimum}\label{eq:fusionfunc:min}\\
    \max (u,v) & & \text{Maximum}\label{eq:fusionfunc:max}
\end{align}

They are special cases of the Generalized mean (power mean or H\"older mean) and satisfy an order 
(see \cref{tab:fusionmetrics}):
\begin{equation}
    \label{eq:metrics:order}
\min \leq H \leq G %
\leq A \leq \operatorname{RMS} %
\leq \max,
\end{equation}
where the equal sign holds if and only if $u=v$.
Eq.~\eqref{eq:metrics:order} also establishes a qualitative order of the depth maps obtained after DSI fusion with the corresponding function.
The arithmetic mean $A$ (i.e., averaging ray densities) corresponds to the above-mentioned particular case of counting the back-projected stereo events on a single DSI.
Hence, functions performing worse than this case are not pursued.

\input{floats/tab_fusion_funcs}

\subsubsection{What makes a good fusion function?}
Intuitively, given two ray densities defined on the same volume, a fusion function should emphasize the regions of high ray density on both DSIs and deemphasize the rest.
It is not sufficient for one of the two densities to be large at a point $\bX$ to signal the presence of a 3D edge; \emph{both} densities have to be similar and large at $\bX$.
The arithmetic mean $A$ and its dominant functions (e.g., quadratic mean and $\max$ in \eqref{eq:metrics:order}) do not satisfy this ``AND'' logic, whereas the geometric mean, harmonic mean and $\min$ functions do satisfy it (e.g., a large value $G(u,v)$ can only be achieved if both $u$ and $v$ are large).

Mathematically, this requirement is well described by the \emph{concavity} properties of the function 
(plot in \cref{tab:fusionmetrics}).
The arithmetic mean $A$ and functions below it are concave (assuming non-negative inputs). 
Further, $G$, $H$ and $\min$ are strictly concave.
As the experiments will show, fusion using $G$ still lets notorious outliers pass.
Functions $H$ and $\min$ deemphasize considerably more than $G$.
The $\min$ function saturates strictly, treating values $u>v$ as $u=v$, hence clipping and discarding potentially beneficial information about ray density values. 
The harmonic mean $H$ shows strong concavity and varies smoothly with both input arguments, without discarding information.
$H$ is dominated by the minimum of its arguments, 
\begin{equation}
\label{eq:Hmean:bounds}
\min(u,v)\leq H(u,v)\leq 2\min(u,v)    
\end{equation}
(in terms of the plot in \cref{tab:fusionmetrics} ($v=1$), the green curve is bounded: $H(u,1)\leq 2$).
The goal of the present work is to introduce and study fusion functions rather than to select a single ``best one''. 
To narrow the discussion we often use a subset of the fusion functions.

\subsubsection{Interpretation in terms of Contrast Maximization}
\label{sec:method:contrastmax}
The proposed stereo fusion method is related to contrast/focus maximization \cite{Gallego18cvpr,Gallego19cvpr}.
The depth slices of the DSIs count refocused events (warped by back-projection),
i.e., they constitute so-called images of warped events (IWEs) \cite{Gallego18cvpr}. 
The fused DSI can be interpreted as a similarity score between refocused events.
Since fusion functions such as $H$ try to emphasize DSI regions with large and similar values, 
stereo Alg.~\ref{alg:fusion:stereo} tries to maximize the similarity score between refocused events (in-focus effect) on both cameras, jointly.
The confidence map registers the maximum focus similarity score at each viewing ray of the fused DSI.

\subsubsection{More fusion functions}

Additional means exist beyond those in \cref{tab:fusionmetrics}, such as the contraharmonic mean, logarithmic mean, quasi-arithmetic mean, arithmetic-geometric mean, Heronian mean and weighted generalized means. 
However they are not covered for the sake of brevity and to avoid clutter.
In some cases, the order relation~\eqref{eq:metrics:order} can be extended to justify their limited practical interest.
Seeking more fusion functions, one could combine functions in \cref{tab:fusionmetrics} with non-linear transformations of the input DSIs, in a homomorphic filtering fashion.
For example, the $A$-mean of the log-DSIs is related to the $G$-mean of the DSIs, which has a stronger concavity than the $A$-mean of the DSIs. %
The same idea can be applied to other functions to increase concavity: the $G$-mean of the log-DSIs, $G(\log(1+u),\log(1+v)))$, has stronger concavity than the $G$-mean of the original DSIs.
The logarithm plays down large DSI values, thus deemphasizing differences between corresponding DSIs before fusion.
For simplicity, we restrict the study to the functions in \cref{tab:fusionmetrics}.

\subsubsection{Loose connection with prior fusion work}
A method for fusing 3D representations called ``temporally synchronized event disparity volumes'' was proposed in \cite{Zhu18eccv},
where two binary data volumes $I_L, I_R$ were fused using an intersection-over-union (IoU) cost.
It resembles the $H$ mean:
\begin{equation}
\label{eq:iouvsHmean}
\text{IoU}=\frac{\sum_{\bx \in W} I_L(\bx,d) \cap I_R(\bx,d)}{\sum_{\bx \in W} I_L(\bx,d) \cup I_R(\bx,d)}
\;\text{ vs. }\; H=2\frac{uv}{u+v}, 
\end{equation}
where the product in the numerator (``intersection'') acts as an ``AND'' condition  
and the sum in the denominator (``union'') acts as a normalization factor.
However, note that the IoU in \cite{Zhu18eccv} is computed by aggregating binary data $I_L, I_R$ over spatial windows $W$ (of $32\times 32$ pixels), whereas $H$ (\cref{tab:fusionmetrics}) is computed voxel-wise, without spatial aggregation (i.e., it has higher spatial resolution), on continuous ray densities.

\subsection{Temporal Fusion}
\label{sec:method:fusion_time}
The functions presented in \cref{sec:method:fusion} can be used to fuse any pair of aligned DSIs.
Moreover, the functions can be extended to handle more than two inputs: 
they allow us to \emph{fuse an arbitrary number of registered DSIs}, with a complexity that is \emph{linear} in the number of DSIs.
The DSIs may be populated by events from different cameras or, 
as we also investigate, from different time intervals.
The main idea is to split an interval into multiple sub-intervals, build the DSI for each of them and fuse all DSIs into a single one (\cref{fig:block-diagram}).
This strategy can be applied regardless of the number of cameras in the system, hence it represents an independent axis of variation. 
Moreover, \emph{the same technique enables camera- and time- fusion},
which we collectively call Alg.~\ref{alg:fusion:time:stereo}.
The key lines of Alg.~\ref{alg:fusion:time:stereo} that change with respect to Alg.~\ref{alg:fusion:stereo} are lines 3 and 4.

Given $N$ fusion functions ($N=6$ in \cref{tab:fusionmetrics}) there are $2N^2$ possible fusion schemes considering the choice of temporal fusion function, across-camera fusion function and the order of application (Alg.~\ref{alg:fusion:time:stereo}).
For brevity we reduce the analysis to the comparison of $N=2$ fusion functions: $A$ and $H$, which yield 8 possible fusion schemes.

Let $A_t$ denote the fusion operation along the time ($t$) axis using the arithmetic mean ($A$).
Likewise, $H_c$ is the fusion operation along the camera ($c$) axis using the harmonic mean ($H$).
Then, $A_t \circ H_c$ first applies $H_c$ (producing as many DSIs as sub-intervals) and then $A_t$.
Out the of 8 possibilities, there are only 6 distinct ones due to commutativity:
\begin{equation}
\label{eq:AlgTwoAAHH}
A_c \circ A_t = A_t \circ A_c \;\text{ and }\; H_c \circ H_t = H_t \circ H_c.
\end{equation}
The four remaining fusion combinations are: 
\begin{equation}
\label{eq:AlgTwoOtherFour}
A_t\circ H_c,\;\; A_c\circ H_t,\;\; H_t\circ A_c\;\; \text{ and }\; H_c\circ A_t.
\end{equation}
Clearly, $A_c\circ A_t$ is equivalent to the approach of summing all stereo events into a single DSI, 
and $H_t \circ H_c$ is very restrictive because only edges seen by all cameras in all subintervals will survive.
Alg.~\ref{alg:fusion:stereo} is also a particular case of Alg.~\ref{alg:fusion:time:stereo} ($H_c\circ A_t$).

\input{floats/alg_fusion_time_stereo}

\input{chapters/03_time_fusion_shuffling}
\input{chapters/03_complexity}

%% file: floats/alg_fusion_cameras.tex
\begin{algorithm}[t!]
  \caption{Stereo event fusion across cameras}
  \label{alg:fusion:stereo}
  {\small
  \begin{algorithmic}[1]
    \State \emph{Input}: stereo events in interval $[0,T]$, camera trajectory, camera calibration (intrinsic and extrinsic).
    \State Define a single reference view (RV) for both DSIs, coinciding with the left camera pose at say $t=T/2$.
    \State Create 2 DSIs by back-projecting events from each camera.
    \State Fusion: compute the pointwise harmonic mean of the DSIs.
    \State Extract depth and confidence maps from the fused DSI $f$:
    $Z^\ast(x,y) \doteq \arg\max f(\bX(x,y))$, $c^\ast(x,y) \doteq \max f(\bX(x,y))$.
  \end{algorithmic}
  }
\end{algorithm}

%% file: floats/fig_mono_vs_stereo.tex
\def\figWidth{0.31\linewidth}
\begin{figure}[t]
	\centering
\begin{subfigure}{0.5\linewidth}
    \includegraphics[width=\linewidth]{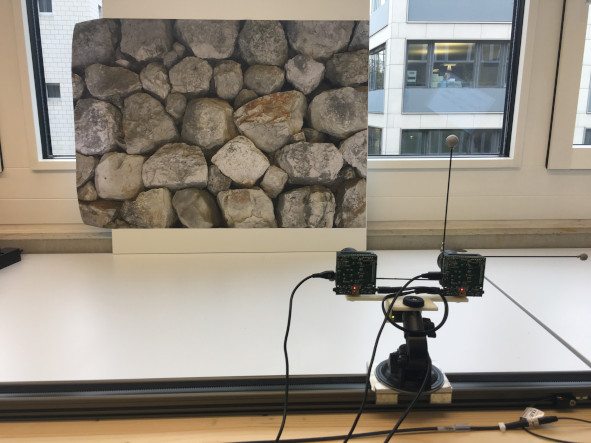}%
    \caption{\label{fig:experim:monovsstereo:setup}Event cameras and scene.}
\end{subfigure}\\[1ex]
\begin{subfigure}{\linewidth}
    {\small
    \setlength{\tabcolsep}{1pt}
	\begin{tabular}{
	>{\centering\arraybackslash}m{0.35cm} 
	>{\centering\arraybackslash}m{\figWidth}
	>{\centering\arraybackslash}m{\figWidth} 
	>{\centering\arraybackslash}m{\figWidth}}
		& \SI{0.05}{\second} & \SI{0.2}{\second} & \SI{1}{\second} \\\addlinespace[0.2ex]

		\rotatebox{90}{\makecell{Monocular~\cite{Rebecq18ijcv}}}
		&\includegraphics[width=\linewidth]{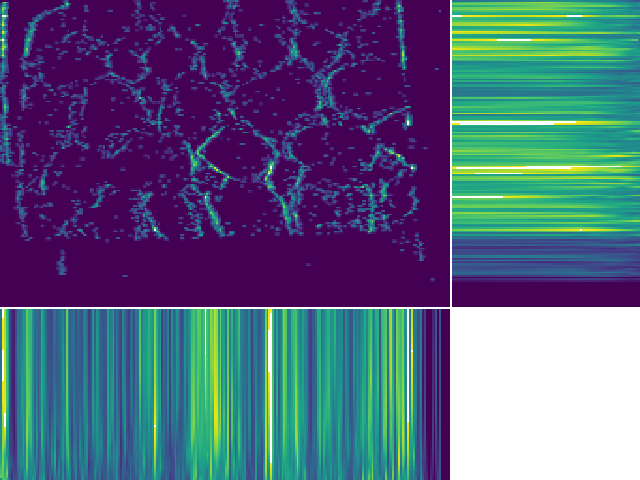}
		&\includegraphics[width=\linewidth]{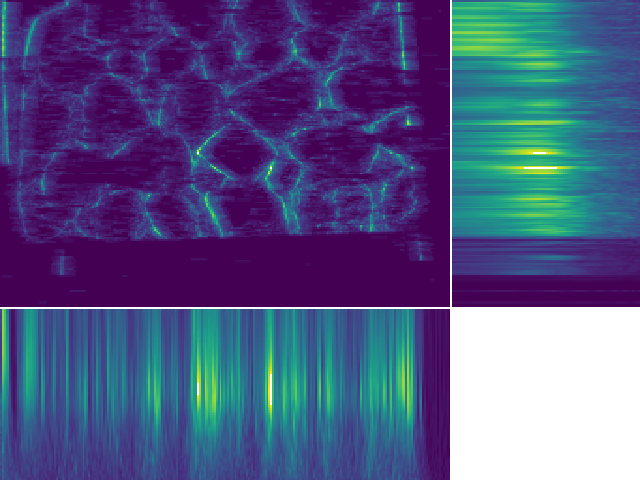}
		&\includegraphics[width=\linewidth]{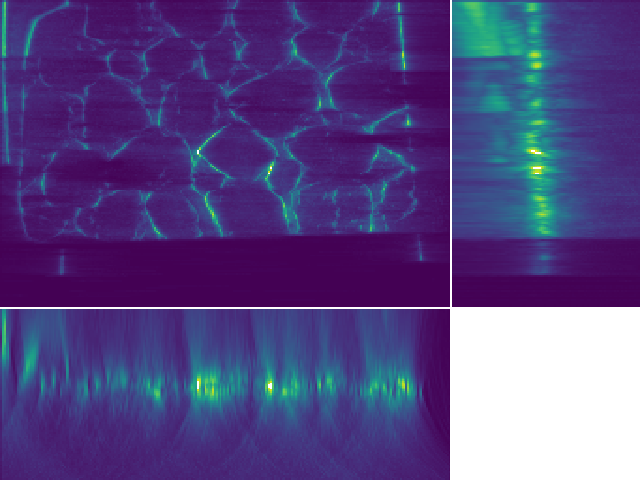}
		\\\addlinespace[-0.3ex]

		\rotatebox{90}{\makecell{Stereo, Alg.~\ref{alg:fusion:stereo}}}
		&\includegraphics[width=\linewidth]{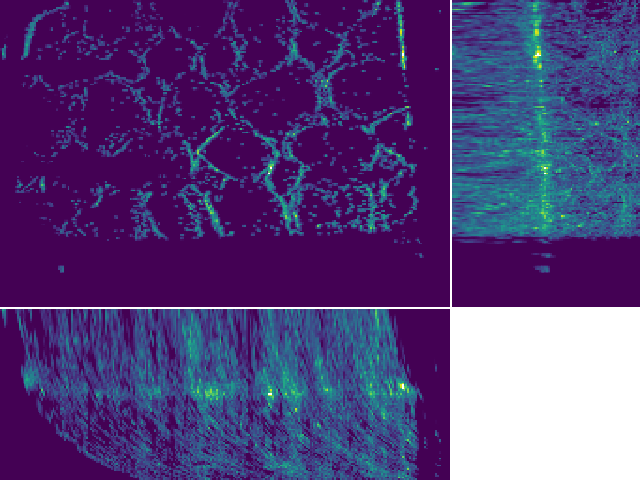}
		&\includegraphics[width=\linewidth]{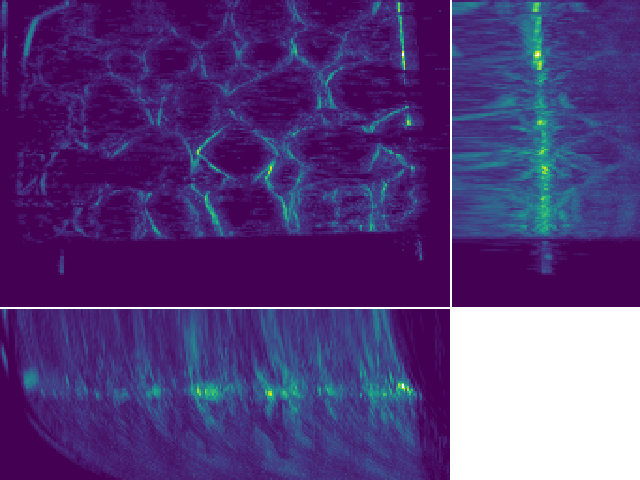}
		&\includegraphics[width=\linewidth]{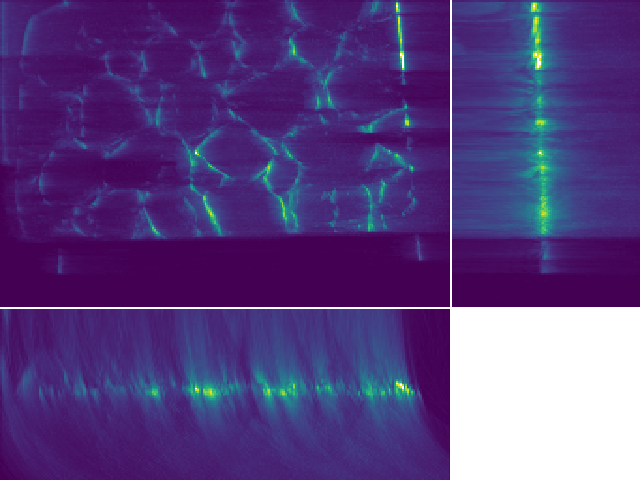}
		\\
	\end{tabular}
	}
    \caption{\label{fig:experim:monovsstereo:dsi}Evolution of DSI projections.}
\end{subfigure}
	\caption{\label{fig:experim:monovsstereo}\emph{Intuitive Example: Monocular vs.~Stereo} method on planar rock scene and 1D motion along the camera's $X$ axis.
	Plots of the evolution of the DSI projections (see text) for different methods (rows) as time increases (columns).
	The 3D edge patterns in the DSI (in yellow) are less localized in EMVS (top row) than in stereo Alg.~\ref{alg:fusion:stereo} (bottom).
	}
\end{figure}

%% file: floats/fig_show_output.tex
\def\figWidth{0.46\columnwidth}
\begin{figure}[t]
	\centering
    {\small
    \setlength{\tabcolsep}{2pt}
	\begin{tabular}{
	>{\centering\arraybackslash}m{0.3cm} 
	>{\centering\arraybackslash}m{\figWidth}
	>{\centering\arraybackslash}m{\figWidth} 
	>{\centering\arraybackslash}m{\figWidth}}
		& \emph{slider\_plane} & \rpgmonitor{}
		\\\addlinespace[0.2ex]
	
		\rotatebox{90}{\makecell{Depth map on frame}}
		&\gframe{\includegraphics[width=\linewidth]{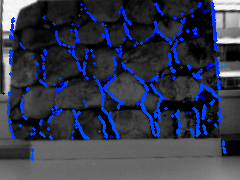}}
		&\gframe{\includegraphics[width=\linewidth]{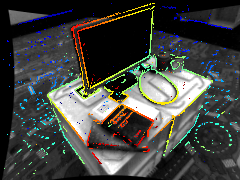}}
		\\
		
		\rotatebox{90}{\makecell{Confidence map}}
        &\gframe{\includegraphics[width=\linewidth]{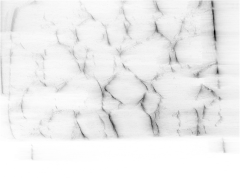}}
        &\gframe{\includegraphics[width=\linewidth]{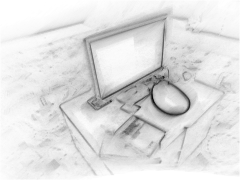}}
		\\
	\end{tabular}
	}
	\caption{\label{fig:output:stereo}
	\emph{Output of Stereo Alg.~\ref{alg:fusion:stereo} on two scenes}.
	Top: our method produces a semi-dense depth map of the scene 
	(color coded from red (close) to blue (far), overlaid on a grayscale frame of the DAVIS~\cite{Brandli14ssc}),
	and a confidence map (Bottom) with the maximum DSI value along each reference view pixel, 
	in negated scale (bright = small; dark = large).
	}
\end{figure}

%% file: floats/tab_fusion_funcs.tex
\begin{figure}[t]
\centering
{\includegraphics[trim={1.8cm 0 2.6cm 0},clip, width=0.5\columnwidth]{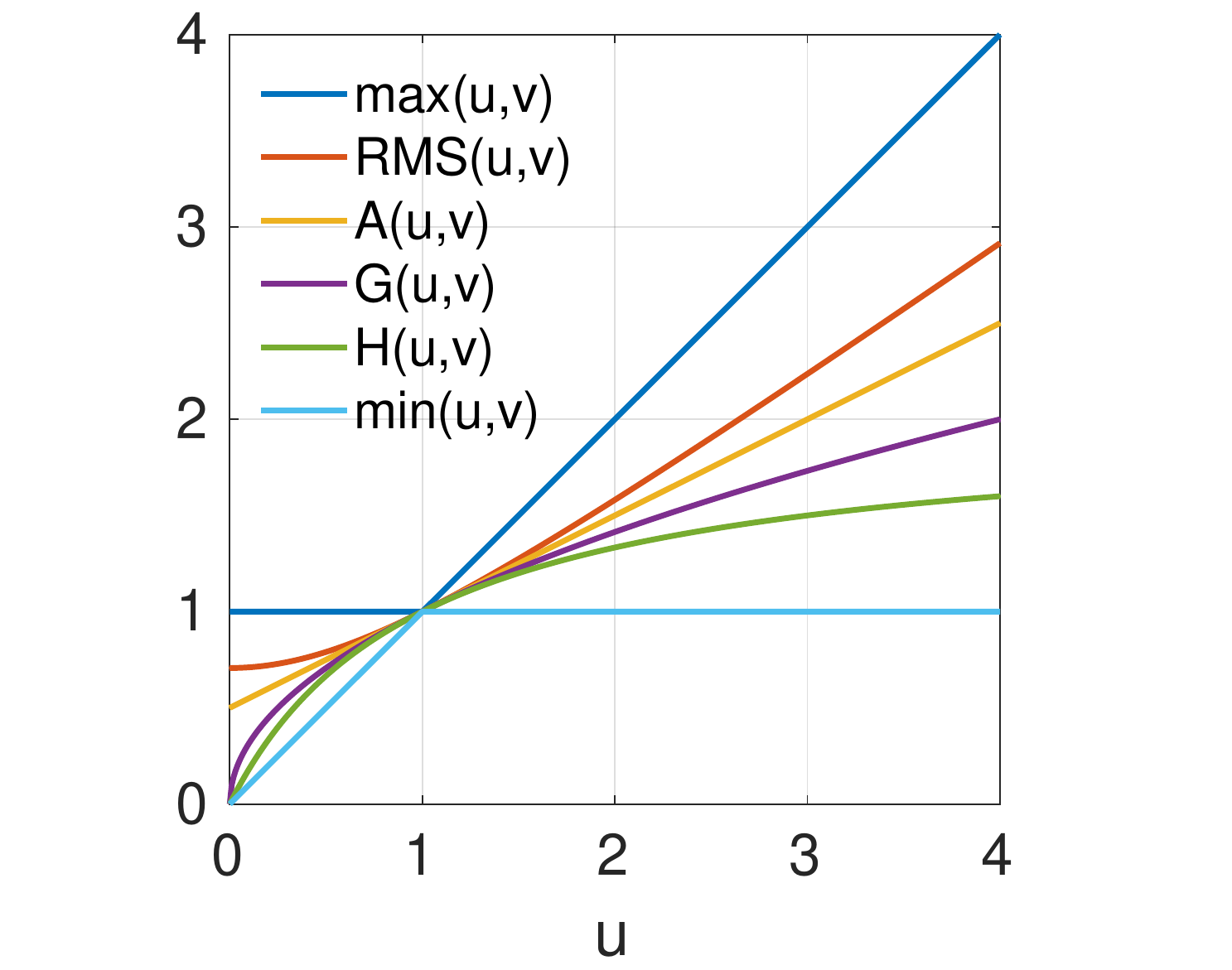}}
\caption{\label{tab:fusionmetrics}Fusion functions considered, for $u=[0,4]$, $v=1$.}
\end{figure}

%% file: floats/alg_fusion_time_stereo.tex
\begin{algorithm}[t]
  \caption{Stereo event fusion across cameras and time}
  \label{alg:fusion:time:stereo}
  {\small
  \begin{algorithmic}[1]
    \State \emph{Input}: stereo events in interval $[0,T]$, camera trajectory, camera calibration (intrinsic and extrinsic).
    \State Define a single reference view (RV) for all DSIs. %
    \State Divide the interval $[0,T]$ into $N_s$ sub-intervals (of equal size or equal number of events). Create $2N_s$ DSIs by back-projecting events from each subinterval and camera.
    \State $S_2 \circ S_1$: 
    Two fusion axes (cameras and time). 
    If $S_2\equiv A_t$ and $S_1 \equiv H_c$, 
    compute first the $S_1$ fusion ($H$-mean of two corresponding DSIs, on the same sub-interval);
    then compute the $S_2$ fusion ($A$-mean of all sub-interval DSIs).
    \State Extract depth and confidence maps from the fused DSI.
  \end{algorithmic}
  }
\end{algorithm}

%% file: chapters/03_time_fusion_shuffling.tex
\subsection{Is Event Simultaneity Needed in Stereo?}
\label{sec:method:shuffling}

Data association is a fundamental problem in event-based vision \cite{Gallego20pami}.
In stereo, event simultaneity is a cornerstone assumption to resolve data association (i.e., find corresponding points) and subsequently infer depth.
A thought-provoking discovery made while developing our method is that across-camera fusion does not need to be done on corresponding intervals (step 4 in Alg.~\ref{alg:fusion:time:stereo}).
We tested our method with the shuffling block in \cref{fig:block-diagram} enabled and it still produced good results (see \cref{sec:experim:shuffled}).
Hence, the proposed stereo method foregoes the event simultaneity assumption.
The explanation is that given the camera poses, events are transformed into a representation (i.e., the DSI) where event simultaneity is not as critical as in the instantaneous stereo problem (\cref{sec:relatedwork}).
The camera poses serve as a proxy allowing us to reproject stereo events to a common DSI and fuse them, even if the DSIs are well separated in time.
The DSI representation is sufficient to produce 3D reconstructions.
Stereo is not solved by matching events, but by comparing possibly non-simultaneous DSIs (each DSI spans several thousands of events).

%% file: chapters/03_complexity.tex
\subsection{Complexity Analysis}
\label{sec:complexity}

Let us analyze the complexity of the proposed stereo methods in comparison with the monocular case.
The main steps of the methods are: 
DSI creation (event back-projection), 
DSI fusion, 
maxima detection along viewing rays of the DSI,
and thresholding (AGT).
If $N_e$ is the number of events, $N_p$ is the number of pixels in the reference view, $N_Z$ is the number of depth planes in the DSI, and $N_k$ is the number of pixels in the AGT kernel (e.g., $5\times 5$), then the complexity of \cite{Rebecq18ijcv} is 
\begin{equation}
    \label{eq:complexityEMVS}
    O(\underbrace{N_e N_Z}_{\text{DSI creation}} + \underbrace{N_Z N_p}_{\text{arg max}} + \underbrace{N_p N_k}_{\text{AGT}}).
\end{equation}

In the case of Alg.~\ref{alg:fusion:stereo} with $N_c$ cameras, assuming that each camera produces $N_e$ events, there are $N_c$ DSIs to build and fuse.
Hence, the complexity is:
\begin{equation}
    \label{eq:complexityAlgOne}
    O(\underbrace{N_c N_e N_Z}_{\text{DSI creation}} + \underbrace{N_c N_Z N_p}_{\text{DSI fusion}} + \underbrace{N_Z N_p}_{\text{arg max}} + \underbrace{N_p N_k}_{\text{AGT}}).
\end{equation}

In the case of Alg.~\ref{alg:fusion:time:stereo} with $N_s$ subintervals, there are $N_s$ DSIs per camera, but each one has $N_e/N_s$ events, and so the complexity of DSI creation does not change.
Only the fusion step becomes more expensive: %
\begin{equation}
    \label{eq:complexityAlgTwo}
    O(\underbrace{N_c N_e N_Z}_{\text{DSI creation}} + \underbrace{N_s N_c N_Z N_p}_{\text{DSI fusion}} + \underbrace{N_Z N_p}_{\text{arg max}} + \underbrace{N_p N_k}_{\text{AGT}}).
\end{equation}

%% file: chapters/04_experiments.tex
\section{Experiments}
\label{sec:experim}
\input{floats/fig_fusion_functions}

To assess the performance of our method we test on a wide variety of real-world and synthetic sequences,
which are introduced in \Cref{sec:experim:datasets}. %
\Cref{sec:experim:fusionfunctions} compares functions for fusion across cameras.
\Cref{sec:experim:methods} compares our method with three state-of-the-art methods on MVSEC and UZH data.
\Cref{sec:experim:timefusion} evaluates temporal fusion and sub-interval shuffling.
Then, we evaluate on higher resolution data: driving dataset DSEC (\cref{sec:experim:dsec}), 
1Mpixel VIO dataset TUM-VIE (\cref{sec:experim:tumvie}), 
and analyze the sensitivity with respect to the camera's spatial resolution (\cref{sec:experim:multires}).
We also present trinocular examples (\cref{sec:experim:morethantwocams}), analyze runtime (\cref{sec:experim:runtime}) 
and sensitivity with respect to the 
camera's contrast threshold (\cref{sec:experim:varyingcontrastthreshold}). 
Finally, \cref{sec:experim:summary} summarizes the findings and \cref{sec:limitations} discusses limitations of the method.

\input{chapters/04_datasets}

\ifclearsectionlook\cleardoublepage\fi
\input{chapters/04_experim_fusion_funcs}

\ifclearsectionlook\cleardoublepage\fi
\input{chapters/04_experim_sota}

\ifclearsectionlook\cleardoublepage\fi
\input{chapters/04_experim_time_fusion}

\ifclearsectionlook\cleardoublepage\fi
\input{chapters/04_experim_dsec}

\ifclearsectionlook\cleardoublepage\fi
\input{chapters/04_experim_tumvie}

\ifclearsectionlook\cleardoublepage\fi
\input{chapters/04_experim_multires}

\ifclearsectionlook\cleardoublepage\fi
\input{chapters/04_experim_analysis}

\subsection{Discussion}
\label{sec:experim:summary}
Let us summarize some of our findings.
In the UZH dataset (\cref{fig:mapping:depthmaps-grid}, top three rows), our method achieves best results compared to the state of the art (SOTA). 
\Cref{fig:mapping:depthmaps-grid} also shows that our stereo method, in its different variations, recovers depth at more fine-detailed structures than SOTA method~ESVO.

In the indoor MVSEC dataset, \cref{tab:sota:mvsec} shows that our stereo method outperforms SOTA quantitatively across multiple standard metrics. 
We also show that the effect of time fusion, while being significant because it speeds up structure convergence and cleans up depth maps, has a smaller effect than that of fusion induced by parallax from an additional sensor (i.e., switching from monocular to stereo) (\cref{tab:sota:mvsec,tab:mvsec:timefusion}).
Remarkably, both strategies are unified in the same theory of fusion of refocused events that we propose.

The outdoor driving MVSEC sequences have very small stereo baseline, which makes them poor for 3D reconstruction \cite{Zhu18rss}, hence we test on the recent DSEC dataset. Here, our method also outperforms the SOTA method ESVO (\cref{tab:dsec}).

The TUM-VIE dataset allowed us to demonstrate experiments on high event camera resolution (1Mpix) and robustness to errors in the camera poses. 
The EVIMO2 dataset allowed us to establish multi-camera (trinocular) depth estimation and during high-speed motion (which blurs regular frames). 

Throughout the experiments (MVSEC, DSEC, TUM-VIE, time fusion, etc.), we have shown the gains with respect to the monocular method; the main advantages of stereo are: higher accuracy, outlier rejection, and faster convergence (due to the additional parallax).

Additionally, we have analyzed the sensitivity with respect to the camera's spatial resolution and contrast threshold: 
the higher the resolution or the lower the threshold, the more accurate our method becomes, at the expense of computational burden due to the larger number of input events.
We also analyzed the computational performance, showing the agreement between complexity (theory) and runtime (practice).

Most interestingly, we have analyzed the effect of shuffling events: our method does not need event simultaneity. 
It can fuse DSIs even if they are built from events well separated in time. 
The best results are obtained fusing identical intervals.

\ifclearsectionlook\cleardoublepage\fi
\input{chapters/04_limitations}

%% file: floats/fig_fusion_functions.tex
\def\figWidth{0.134\linewidth}
\begin{figure*}[t]
	\centering
    {\small
    \setlength{\tabcolsep}{1pt}
	\begin{tabular}{
	>{\centering\arraybackslash}m{0.5cm} 
	>{\centering\arraybackslash}m{\figWidth} 
	>{\centering\arraybackslash}m{\figWidth}
	>{\centering\arraybackslash}m{\figWidth} 
	>{\centering\arraybackslash}m{\figWidth}
	>{\centering\arraybackslash}m{\figWidth}
	>{\centering\arraybackslash}m{\figWidth}
	>{\centering\arraybackslash}m{\figWidth} 
	>{\centering\arraybackslash}m{\figWidth}}
		& Scene & $\min$ & $H$ & $G$ & $A$ & RMS & $\max$ \\\addlinespace[.3ex]

		\rotatebox{90}{\makecell{Depth}}
		&\includegraphics[width=\linewidth]{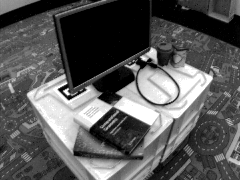}
		&\gframe{\includegraphics[width=\linewidth]{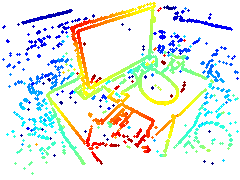}}
		&\gframe{\includegraphics[width=\linewidth]{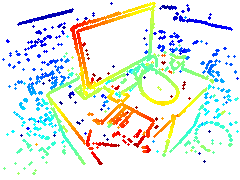}}
		&\gframe{\includegraphics[width=\linewidth]{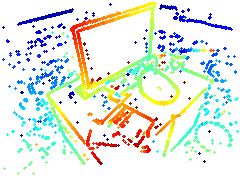}}
		&\gframe{\includegraphics[width=\linewidth]{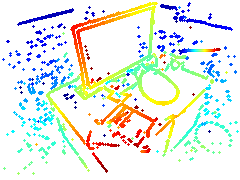}}
		&\gframe{\includegraphics[width=\linewidth]{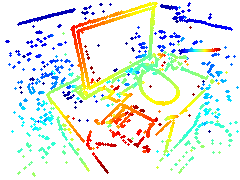}}
		&\gframe{\includegraphics[width=\linewidth]{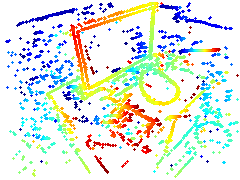}}
		\\
		
		\rotatebox{90}{\makecell{Depth (zoom)}}
		&\includegraphics[width=\linewidth]{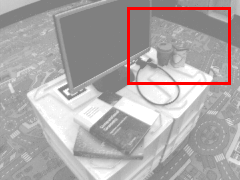}
		&\gframe{\includegraphics[width=\linewidth]{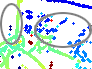}}
		&\gframe{\includegraphics[width=\linewidth]{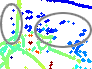}}
		&\gframe{\includegraphics[width=\linewidth]{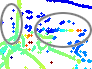}}
		&\gframe{\includegraphics[width=\linewidth]{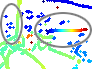}}
		&\gframe{\includegraphics[width=\linewidth]{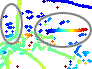}}
		&\gframe{\includegraphics[width=\linewidth]{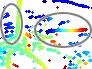}}
		\\
		
		\rotatebox{90}{\makecell{Conf. map}}
		&
		&\gframe{\includegraphics[width=\linewidth]{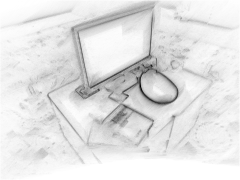}}
		&\gframe{\includegraphics[width=\linewidth]{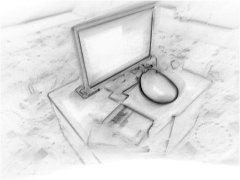}}
		&\gframe{\includegraphics[width=\linewidth]{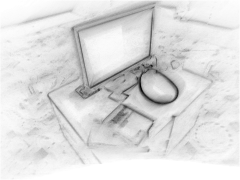}}
		&\gframe{\includegraphics[width=\linewidth]{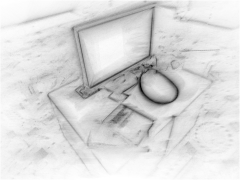}}
		&\gframe{\includegraphics[width=\linewidth]{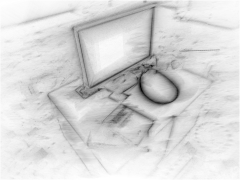}}
		&\gframe{\includegraphics[width=\linewidth]{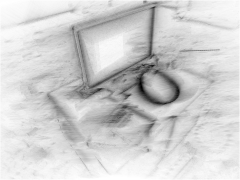}}
		\\
		
	\end{tabular}
	}
	\caption{\label{fig:fusion:functions}
	\emph{Fusion Functions}.
	Semi-dense depth maps (top rows) and confidence maps (bottom row) produced by Alg.~\ref{alg:fusion:stereo} using the fusion functions in \cref{tab:fusionmetrics} on data from~\cite{Zhou18eccv}.
	The columns (fusion functions) follow the order in~\eqref{eq:metrics:order}.
	The differences are subtle: the depth maps on the left columns have fewer outliers than those on the right columns (zoomed-in insets). 
	The differences are more noticeable in the confidence maps, whose sharpness increase from right to left.
	Depth is color coded, from red (close) to blue (far), 
	in the range \SIrange{0.55}{6.25}{\meter}.
	Confidence maps are colored as in \cref{fig:output:stereo}.
	}
\end{figure*}

%% file: chapters/04_datasets.tex
\subsection{Datasets and Evaluation Metrics}
\label{sec:experim:datasets}

\input{floats/tab_stereo_rig}
\input{floats/fig_across_cam_fusion_stack}

\subsubsection{Datasets}
We evaluate our stereo methods on sequences from five publicly available datasets~\cite{Zhou18eccv,Zhu18ral,Mitrokhin19iros,Gehrig21ral,Klenk21iros} and a simulator.
Sequences from \cite{Zhou18eccv,Mitrokhin19iros} were acquired with a hand-held stereo or trinocular event camera in indoor environments.
Sequences in the MVSEC dataset~\cite{Zhu18ral} were acquired with a stereo event camera mounted on a drone while flying indoors.
The sequences in the DSEC dataset \cite{Gehrig21ral} were recorded with event cameras on a car that drove through Zurich's surroundings.
The TUM-VIE dataset was recorded with the sensor rig mounted on a helmet, and its sequences contain indoor and outdoor scenes.
The simulator~\cite{Rebecq18corl,Mueggler17ijrr} provides synthetic sequences using an ideal event camera model and scenes built using CAD models.

\emph{Ground Truth}.
Some datasets contain ground truth \emph{poses} from a motion-capture system, which we use as input to all tested methods. 
If camera poses are not available (e.g., TUM-VIE), we compute them using data from the sensor rig (e.g., a visual-inertial odometry algorithm).
Some datasets, such as MVSEC and DSEC, contain ground truth \emph{depth} for quantitative assessment of the 3D reconstruction methods. 
Depth is given by a LiDAR operating at \SIrange{10}{20}{\hertz}. 
The event camera pixels corresponding to points outside the LiDAR's field of view (FOV) or points close to the sensor rig may not have a LiDAR depth value.

\emph{Rigs and Calibration}.
The main geometric parameters of the event cameras used in the above datasets are summarized in \cref{tab:stereo-rig-params}.
The stereo rigs in~\cite{Zhou18eccv,Zhu18ral} consist of two Dynamic and Active Pixel Vision Sensors (DAVIS)~\cite{Brandli14ssc}. 
The DAVIS comprises a frame-based and an event-based sensor on the same pixel array, thus calibration (intrinsic and extrinsic) 
is achieved using the intensity frames, and then it is applied to the events.
The datasets whose cameras output only events (EVIMO2, DSEC and TUM-VIE), are calibrated by converting events to frames and calibrating the latter (e.g., using \cite{Muglikar21cvprw}).
All methods work on undistorted coordinates.

\subsubsection{Metrics}
The performance of the proposed method is quantitatively characterized using several standard metrics on the datasets with ground truth depth (i.e., MVSEC and DSEC).
We provide mean and median errors between the estimated depth and the ground truth one 
(median errors are more robust to outliers than mean errors).
We also report the number of reconstructed points, the number of outliers (bad-pix \cite{Geiger12cvpr}), 
the scale invariant depth error (SILog Err), 
the sum of absolute value of relative differences in depth (AErrR), 
and $\delta$-accuracy values on the percentage of points whose depth ratio with respect to ground truth is within some threshold (see \cite{Ye19arxiv}).
We also provide precision, recall and F1-score curves \cite{Pizzoli14icra}.
\emph{Precision} is the percentage of estimations that are within a certain error from the ground truth.
\emph{Recall} (e.g., completeness or reconstruction density) is the percentage of ground truth points that are within a certain error from the estimations.
The \emph{F1 score} is the harmonic mean of precision and recall, %
which is dominated by the smallest of them. 
Since the depth maps obtained are semi-dense (while the ground truth is often more dense), recall often dominates.

The method in \cref{sec:method:stereodepth} is presented for the events in a time window.
To apply the method to a whole sequence, we split the latter into non-overlapping time windows 
and apply the method to each of them.
When thresholding to obtain semi-dense depth maps (AGT step), we normalize by a robust maximum DSI value obtained over the sequence, 
which makes the comparisons more stable. %

%% file: floats/tab_stereo_rig.tex
\begin{table}[t]
\centering
    \caption{\label{tab:stereo-rig-params}Parameters of stereo or trinocular event-camera rigs used in the experiments.}
    
    \begin{adjustbox}{max width=\linewidth}
    \setlength{\tabcolsep}{4pt}
    \begin{tabular}{@{}lllll@{}}
    \toprule
    Dataset           & Cameras & Resolution [pix] & Baseline [\si{\centi\meter}] & FOV [\si{\degree}] %
    \\
    \midrule
    ECCV18 \cite{Zhou18eccv} & DAVIS240C & $240 \times 180$ & 14.7 & 62.9 \\[0.5ex]
    MVSEC \cite{Zhu18ral}   & DAVIS346  & $346 \times 260$ & 10.0 & 74.8  \\[0.5ex] %
    EVIMO2 \cite{Mitrokhin19iros} & Samsung Gen3  & $640 \times 480$ & trinocular & 75 \\
                      & $2\times$ Prophesee Gen3  & $640 \times 480$ & trinocular & 70 \\[0.5ex]
    DSEC \cite{Gehrig21ral}  & Prophesee Gen3  & $640 \times 480$  & 60 & 60.1   \\[0.5ex]
    TUM-VIE \cite{Klenk21iros}  & Prophesee Gen4  & $1280 \times 720$ & 11.84 & 90 \\[0.5ex]
    ESIM \cite{Rebecq18corl} & Simulator & up to $1280 \times 960$ & 20 & 77.3 \\[0.5ex]
    \bottomrule
    \end{tabular}
    \end{adjustbox}
\end{table}

%% file: floats/fig_across_cam_fusion_stack.tex
\def\figWidth{0.24\linewidth}
\begin{figure*}[t]
	\centering
	\captionof{table}{\label{tab:mvsec:across-cam-small-data}
	Depth errors for stereo DSI fusion across cameras (\cref{sec:experim:fusionfunctions}).
	Experiments on 200~\si{\s} (110 million events) of the three indoor flying sequences from MVSEC \cite{Zhu18ral}.
	The maximum scene depth is \SI{8.4}{\meter}.
	}
    \input{floats/tab_fusion_mvsec}
\end{figure*}

%% file: floats/tab_fusion_mvsec.tex
\centering
\begin{adjustbox}{width=.93\textwidth}
\setlength{\tabcolsep}{5pt}
\begin{tabular}[t]{l*{12}{S[table-format=2.2,table-number-alignment=left]}}
\toprule
 & \multicolumn{3}{c}{Mean Abs Error {[}cm{]} $\downarrow$} & \multicolumn{3}{c}{Median Abs Error {[}cm{]} $\downarrow$} & \multicolumn{3}{c}{bad-pix {[}\%{]} $\downarrow$} & \multicolumn{3}{c}{\#Points {[}million{]} $\uparrow$}\\
\cmidrule(l{1mm}r{1mm}){2-4} \cmidrule(l{1mm}r{1mm}){5-7} \cmidrule(l{1mm}r{1mm}){8-10} \cmidrule(l{1mm}r{1mm}){11-13}
Sequence~\cite{Zhu18ral} & \text{flying1} & \text{flying2} & \text{flying3} & \text{flying1} & \text{flying2} & \text{flying3}& \text{flying1} & \text{flying2} & \text{flying3} & \text{flying1} & \text{flying2} & \text{flying3}\\
\midrule 
$\min_c\circ A_t$  & 58.13254379 & 68.79095868 & 49.96271304 & 24.072123 & 39.20403 & 20.34686208 & 17.40382153 & 38.82488066 & 12.39623001 & 6.127776 &	12.722571	& 5.827461\\
Alg.~\ref{alg:fusion:stereo} ($H_c\circ A_t$)\; & 60.16473811 & 68.82614491 & 51.94287048 & 25.45290589 & 39.62654173 & 21.14963531 & 18.57710592 & 38.90740741 & 13.69819373 & 6.612663	& 13.543398	& 6.27293\\
$G_c\circ A_t$ & 63.56692372 & 70.48220226 & 55.2165504 & 28.172183 & 41.509962 & 22.97324 & 20.30386645 & 39.68478983 & 15.29768101 & 6.89786 &	14.175827 &	6.346041\\
$A_c\circ A_t$  & 78.77535369 & 79.94705882 & 60.35128938 & 38.1496489 & 47.6252079 & 25.02818 & 27.195554 & 44.64192641 & 17.55990533 & 5.965837	& 14.532117	& 4.497053\\
$\text{RMS}_c\circ A_t$  & 99.14739544 & 88.46178431 & 88.20503585 & 61.728716 & 55.7063818 & 47.45121 & 36.83509371 & 48.88969607 & 29.85994717 & 7.646645	& 17.475859 &	5.412051\\
$\max_c\circ A_t$ & 109.3017139 & 93.81060186 & 104.5152567 & 75.438595 & 61.71233654 & 66.564536 & 41.31220227 & 51.64094436 & 36.27356056 & 9.24995	& 19.985385	& 6.917156\\
\bottomrule
\end{tabular}

\end{adjustbox}

%% file: chapters/04_experim_fusion_funcs.tex
\subsection{Comparison of Across-Camera Fusion Functions}
\label{sec:experim:fusionfunctions}

\input{floats/fig_experiments_large}
\input{floats/tab_sota_mvsec}

We first evaluate Alg.~\ref{alg:fusion:stereo} using the fusion functions in \cref{tab:fusionmetrics}.
\Cref{fig:fusion:functions} shows qualitatively the corresponding depth- and confidence maps for a sample sequence. 
The columns follow the order in~\eqref{eq:metrics:order}.
The arithmetic mean ($A$) corresponds to counting the back-projected event rays from both cameras (left/right) on a single DSI. 
It has not been proposed before in the literature and is a particular case of our fusion methods.
It provides moderate results by conveying that a 3D point is detected if enough rays are counted at a voxel, regardless of which camera the event ray originated from. 
However, it does not filter out spurious ray intersections (which do not correspond to actual 3D points) until there is sufficient evidence. 
Spurious ray intersections are more common in the stereo case than in the monocular one because rays are originated from two moving sources in stereo instead of just one.
The columns in \cref{fig:fusion:functions} to the right of $A$ (i.e., quadratic mean and $\max$) produce worse results than $A$.
They convey that 3D points are detected if enough rays from at least one camera are counted at a voxel. 
This strategy might be good to mitigate occlusions (edges seen in only one camera), but it does not produce optimal results if the edge is visible from both event cameras.
Finally, the columns to the left of $A$ (i.e., $G,H,\min$) produce better results than $A$.
This is due to the fact that they implement a more conservative strategy: 3D points are detected only if enough rays from both cameras are counted at a voxel.
The results are more noticeable in the confidence maps; these are sharper in the first columns than in the last ones.

\Cref{tab:mvsec:across-cam-small-data} quantitatively compares the six fusion functions on the three indoor flying sequences from~\cite{Zhu18ral}.
The experiment consists of running stereo Alg.~\ref{alg:fusion:stereo} on 200~\si{\s} of data (110 million events), at 20~\si{\Hz}. 
The estimated depth produces $\approx$28.4 million points on $\approx$4000 ground truth snapshots for each fusion function.
The differences between fusion functions are most noticeable when less data is available, and so we use events packets of 0.1s for this experiment.
\Cref{tab:mvsec:across-cam-small-data} reports mean and median depth errors, bad-pixel percentage and number of reconstructed points.
Errors and bad-pix follow a clear trend, decreasing towards the top rows. 
The number of reconstructed points also decreases, but non-monotonically in sequences 1 and 3.
Median errors are considerably smaller than mean errors, signaling the presence of outliers (points with large depth errors).
We observe an accuracy-completion trade-off: the top rows are more accurate than the bottom rows, but the bottom rows provide more reconstructed points. 
Comparing the top two rows (highest accuracy), the differences in accuracy are small (mean: \SI{2.41}{\percent}, median: \SI{3.43}{\percent}) while the difference in number of points is larger: \SI{6.83}{\percent}. 
Hence these results indicate, together with theoretical aspects (\cref{sec:method:fusion}), that the $H$-mean is advantageous to fuse across cameras.

%% file: floats/fig_experiments_large.tex
\def\figWidth{0.136\linewidth}
\begin{figure*}[t!]
	\centering
    {\small
    \setlength{\tabcolsep}{1pt}
	\begin{tabular}{
	>{\centering\arraybackslash}m{0.3cm} 
	>{\centering\arraybackslash}m{\figWidth} 
	>{\centering\arraybackslash}m{\figWidth}
	>{\centering\arraybackslash}m{\figWidth}
	>{\centering\arraybackslash}m{\figWidth}
	>{\centering\arraybackslash}m{\figWidth}
	>{\centering\arraybackslash}m{\figWidth}
	>{\centering\arraybackslash}m{\figWidth}}
		& Scene 
		& \GTS~\cite{Ieng18fnins} 
		& SGM~\cite{Hirschmuller08pami} 
		& ESVO \cite{Zhou20tro} 
		& Alg.~\ref{alg:fusion:stereo}
		& $A_t\circ H_c$
		& $A_c\circ A_t$
		\\\addlinespace[.3ex]

		\rotatebox{90}{\makecell{\rpgreader{}}}
		&\includegraphics[width=\linewidth]{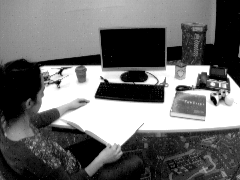}
		&\gframe{\includegraphics[width=\linewidth]{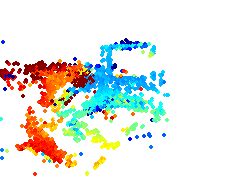}}
		&\gframe{\includegraphics[width=\linewidth]{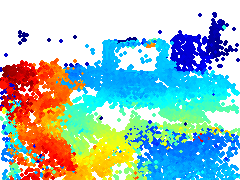}}
		&\gframe{\includegraphics[width=\linewidth]{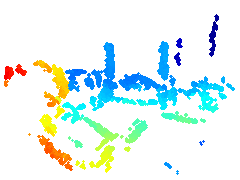}}
        &\gframe{\includegraphics[width=\linewidth]{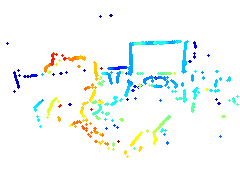}}
        &\gframe{\includegraphics[width=\linewidth]{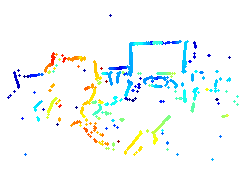}}
        &\gframe{\includegraphics[width=\linewidth]{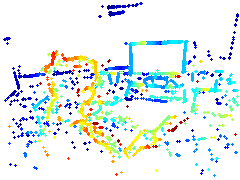}}
		\\\addlinespace[-0.3ex]

		\rotatebox{90}{\makecell{\rpgbox{}}}
		&\includegraphics[width=\linewidth]{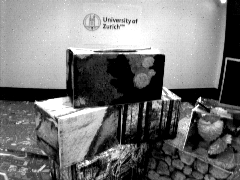}
		&\gframe{\includegraphics[width=\linewidth]{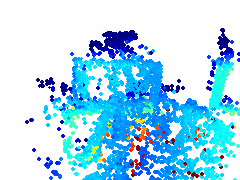}}
		&\gframe{\includegraphics[width=\linewidth]{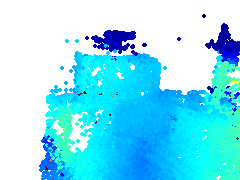}}
		&\gframe{\includegraphics[width=\linewidth]{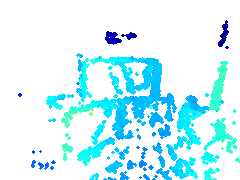}}
		&\gframe{\includegraphics[width=\linewidth]{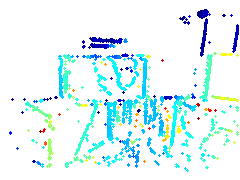}}
		&\gframe{\includegraphics[width=\linewidth]{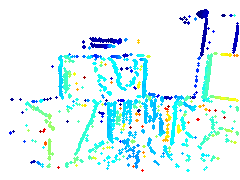}}
		&\gframe{\includegraphics[width=\linewidth]{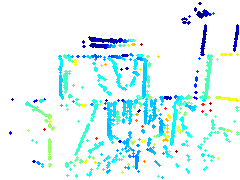}}
		\\\addlinespace[-0.3ex]
		
		\rotatebox{90}{\makecell{\rpgmonitor{}}}
		&\includegraphics[width=\linewidth]{images/exp_mapping/monitor2/rpg_eccv2018_monitor2.png}
		&\gframe{\includegraphics[width=\linewidth]{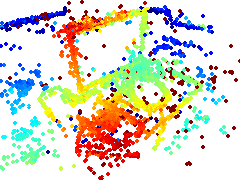}}
		&\gframe{\includegraphics[width=\linewidth]{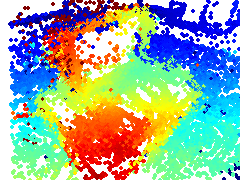}}
		&\gframe{\includegraphics[width=\linewidth]{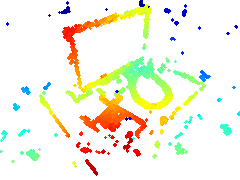}}
        &\gframe{\includegraphics[width=\linewidth]{images/rpg_monitor/14.000000inv_depth_colored_dilated_fused_2_w.png}}
		&\gframe{\includegraphics[width=\linewidth]{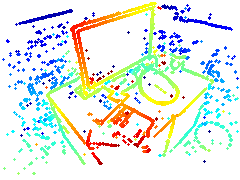}}
        &\gframe{\includegraphics[width=\linewidth]{images/rpg_monitor/14.000000inv_depth_colored_dilated_fused_4_w.png}}
		\\\addlinespace[-0.3ex]
		
		\rotatebox{90}{\makecell{\emph{upenn\_fly1}}}
		&\includegraphics[width=\linewidth]{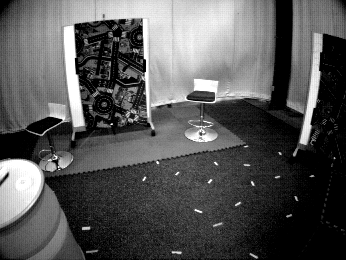}
		&\gframe{\includegraphics[width=\linewidth]{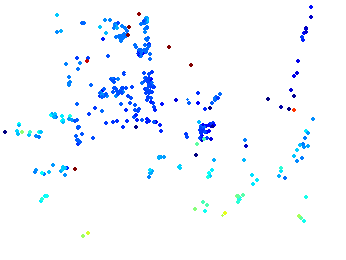}}
		&\gframe{\includegraphics[width=\linewidth]{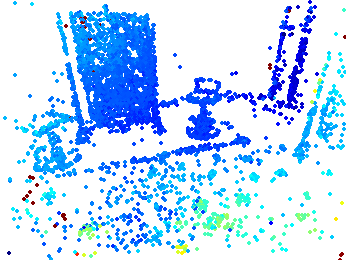}}
		&\gframe{\includegraphics[width=\linewidth]{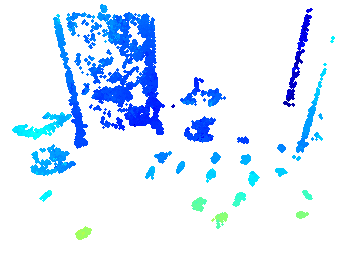}}
		&\gframe{\includegraphics[width=\linewidth]{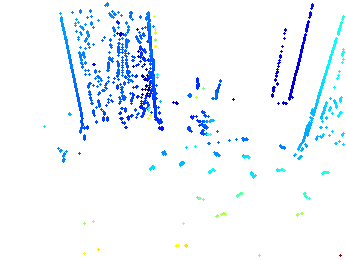}}
		&\gframe{\includegraphics[width=\linewidth]{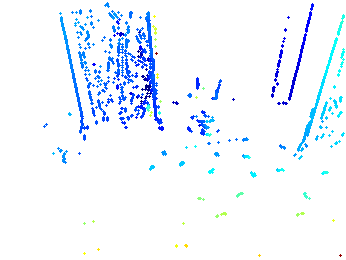}}
		&\gframe{\includegraphics[width=\linewidth]{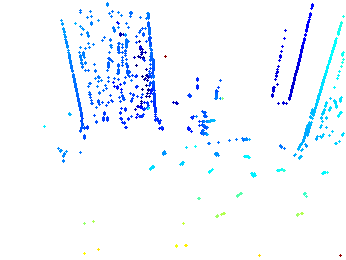}}
		\\\addlinespace[-0.3ex]

		\rotatebox{90}{\makecell{\emph{upenn\_fly3}}}
		&\includegraphics[width=\linewidth]{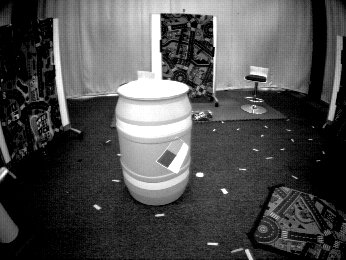}
		&\gframe{\includegraphics[width=\linewidth]{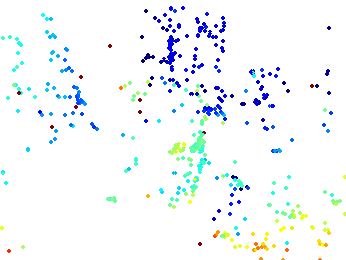}}
		&\gframe{\includegraphics[width=\linewidth]{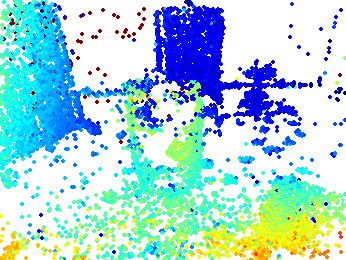}}
		&\gframe{\includegraphics[width=\linewidth]{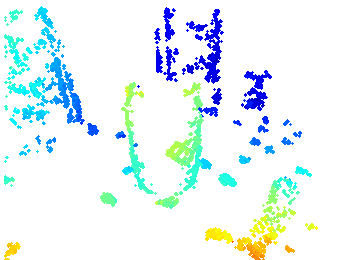}}
		&\gframe{\includegraphics[width=\linewidth]{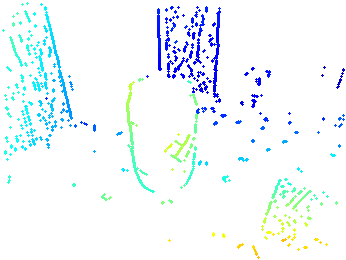}}
		&\gframe{\includegraphics[width=\linewidth]{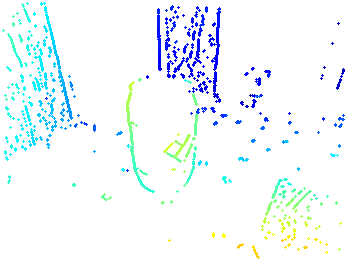}}
		&\gframe{\includegraphics[width=\linewidth]{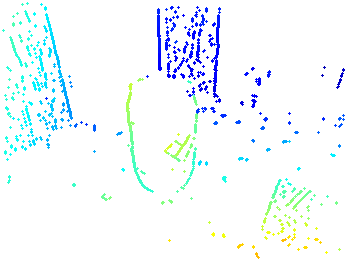}}\\\addlinespace[-0.3ex]
	\end{tabular}
	}
	\caption{\label{fig:mapping:depthmaps-grid}\emph{Event-based Stereo 3D Reconstruction}.
	Comparison of depth estimation results on several sequences using various stereo methods.
	For visualization purposes, the first column depicts intensity frames from the DAVIS camera (not used by any method).
	Columns 2 to 7 show semi-dense inverse depth maps produced by \GTS~\cite{Ieng18fnins}, SGM~\cite{Hirschmuller08pami}, ESVO~\cite{Zhou20tro}, our Alg.~\ref{alg:fusion:stereo} and Alg.~\ref{alg:fusion:time:stereo} ($A_t\circ H_c$ and $A_c\circ A_t$, with $N_s=2$), respectively.
	Depth maps are pseudo-colored, from red (close) to blue (far),
	in the range \SIrange{0.55}{6.25}{\meter} for the \emph{rpg} sequences~\cite{Zhou18eccv} and 
	in the range \SIrange{1}{6.5}{\meter} for the \emph{upenn} MVSEC sequences~\cite{Zhu18ral}.
	}
\end{figure*}

%% file: floats/tab_sota_mvsec.tex
\begin{table*}[t]
\centering
\begin{adjustbox}{width=\textwidth}
\setlength{\tabcolsep}{3pt}
\begin{tabular}{ll*{1}{S[table-format=2.2,table-number-alignment=center]}
*{4}{S[table-format=1.2,table-number-alignment=center]}
*{4}{S[table-format=2.2,table-number-alignment=center]}
*{1}{S[table-format=1.2,table-number-alignment=center]}}
\toprule 
&Algorithm & \text{Mean Err} & \text{Median Err} & \text{bad-pix} & \text{SILog Err} & \text{AErrR} & \text{log RMSE} & \text{$\delta < 1.25$} & \text{$\delta < 1.25^2$} & \text{$\delta < 1.25^3$} & \text{\#Points}\\
& & \text{[cm] $\downarrow$} & \text{[cm] $\downarrow$} & \text{[\%] $\downarrow$} & \text{$\times 100 \downarrow$} & \text{[\%] $\downarrow$} & \text{$\times 100 \downarrow$} & \text{[\%] $\uparrow$} & \text{[\%] $\uparrow$} & \text{[\%] $\uparrow$} & \text{[million] $\!\uparrow$}\\
\midrule
\multirow{5}{*}{\begin{turn}{90}SOTA\end{turn}} 
& EMVS \cite{Rebecq18ijcv} (monocular)  & 33.7751741 & 14.35062018 & 3.838182793 & 4.204367203 & 12.73857883 & 20.7224234 & 84.75111662 & 94.86663563 & 97.98934193 & 1.27\\
& ESVO \cite{Zhou20tro} & 25.00149683 & 10.59420305 & 3.35436039 & 3.483746516 & 10.19387852 & 18.83386882 & 90.43747571 & 95.76258152 & 97.97747808 & 2.04\\
& ESVO indep. 1s    & 22.70020432 & 9.833073427 & 2.831496404 & 3.026721445 & 9.585589113 & 17.53474656 & 91.82213853 & 96.49844849 & 98.38060529 & 1.56\\
& SGM indep. 1s     & 35.41582815 & 12.34814531 & 6.385815911 & 8.454076416 & 16.16815315 & 29.48681941 & 85.3444745 & 93.05255453 & 96.03251214 & \bnum{14.46}\\
& GTS indep. 1s     & 388.9995217 & 45.42733961 & 38.44853556 & 74.47240605 & 102.9226872 & 89.07963838 & 49.55505336 & 62.1883512 & 69.36289429 & 0.06\\
\midrule
\multirow{8}{*}{\begin{turn}{90}Ours\end{turn}} 
& $H_c\circ A_t$ (Alg~\ref{alg:fusion:stereo}) & \bnum{20.0720842} & \bnum{9.530901963} & \bnum{1.353268285} & \bnum{1.718787127} & \bnum{7.795808274} & \bnum{13.23558561} & \bnum{95.03581089} & \bnum{98.07547472} & \bnum{99.21066927} & 0.81\\
& $H_c\circ A_t$ (Alg~\ref{alg:fusion:stereo}) + MF & 20.64393831 & 9.720921988 & \unum{1.426366492} & \unum{1.796872049} & 7.936701079 & \unum{13.54005872} & \unum{94.73737768} & 97.95204147 & 99.16869564 & \unum{3.00}\\
& $H_c \circ H_t$   & 23.44557407 & 10.97523011 & 1.893246143 & 2.175004606 & 8.862131285 & 14.93215503 & 92.99506475 & 97.49319475 & 99.02736031 & 1.47\\
& $H_t \circ A_c$   & 22.60895386 & 10.68238428 & 1.669937341 & 2.032274988 & 8.613914361 & 14.43864459 & 93.48946726 & 97.75805881 & 99.12208482 & 1.25\\
& $A_c\circ H_t$    & 23.35069354 & 10.93604449 & 1.838998406 & 2.150553916 & 8.833090906 & 14.84722116 & 93.07219297 & 97.53080778 & 99.04124585 & 1.42\\
& $A_c \circ A_t$   & \unum{20.38238841} & \underline{\num{9.59756254}} & 1.50851983 & 1.797244927 & \unum{7.926398252} & 13.54701863 & 94.66690436 & \unum{98.00510168} & \unum{99.20335952} & 0.99\\
& $A_t \circ H_c$   & 20.91578705 & 9.759206544 & 1.662014008 & 1.8631732 & 8.045587963 & 13.81175316 & 94.3947297 & 97.80300225 & 99.14006845 & 1.14\\
& $A_t \circ H_c$ + shuffling   & 22.60447717 & 10.70503599 & 1.684106626 & 2.114156596 & 8.578011168 & 14.277874 & 93.49026928 & 97.76303041 & 99.13257469 & 1.24\\
\bottomrule
\end{tabular}
\end{adjustbox}
\caption{\label{tab:sota:mvsec}Quantitative evaluation and comparison of our proposed method with the state of the art. 
All metrics are averaged over the three indoor MVSEC sequences (flying 1, 2, 3), where the maximum ground truth depth is \SI{8.4}{\meter}. 
The methods are evaluated on 200~\si{\s} of data (110 million events and 4000 ground truth depth maps).
Each estimated depth map is computed using 1~\si{\s} of event data ($\approx0.55$ million events).
MF: morphological filter (see text).
Per-sequence results are in \cref{tab:sota:mvsec:one,tab:sota:mvsec:two,tab:sota:mvsec:three}.
}
\end{table*}

%% file: chapters/04_experim_sota.tex
\subsection{Comparison with Stereo State of the Art}
\label{sec:experim:methods}

We assess the performance of our methods in comparison to several event-based stereo methods, 
in \cref{fig:mapping:depthmaps-grid} and \cref{tab:sota:mvsec}.

\subsubsection{Baseline Methods}
The Generalized Time-Based Stereovision method (\GTS{})~\cite{Ieng18fnins} follows a classical two-step approach: stereo matching plus triangulation. 
Matching is based on a per-event time-based consistency score.
The Semi-Global Matching (SGM) method~\cite{Hirschmuller08pami} is adapted to event data by feeding time images~\cite{Lagorce17pami} and masking the produced depth map at recent event locations \cite{Zhou18eccv}.
We also compare against the mapping module of Event-based Stereo Visual Odometry (ESVO)~\cite{Zhou20tro}, 
which fuses multiple depth estimates using Student-$t$ filters, with each estimate and its uncertainty obtained by maximizing spatio-temporal consistency between patches of stereo time images.
\GTS{} and SGM are also endowed with depth propagation-and-fusion filters, as implemented in~\cite{Zhou20tro}.
Finally, we also compare stereo against the monocular method EMVS \cite{Rebecq18ijcv}, 
which has not been carried out before.
All baseline methods produce depth maps at the LiDAR rate (20~\si{\Hz}) and use ground truth poses to propagate depth estimates in time, if needed.
ESVO on MVSEC data works by fusing 20 depth maps generated at 20~\si{\Hz}, i.e., 1~\si{\second} of data.
EMVS works on event packets of 1~\si{\second}, shifted by 50~\si{\milli\second} (20~\si{\Hz}). 
For a sensible comparison with EMVS, baseline stereo methods are also run on event packets of 1~\si{\second}, shifted by 50~\si{\milli\second};
this is highlighted as ``indep 1s'' in \cref{tab:sota:mvsec}.

\subsubsection{Results}
\cref{fig:mapping:depthmaps-grid} compares qualitatively the inverse depth maps produced by the above stereo methods.
To illustrate the appearance of the scenes, the first column shows grayscale frames from the DAVIS~\cite{Brandli13fns}. 
The remaining columns show the output of \GTS{}, SGM, ESVO and our methods.
Because event cameras naturally respond to the apparent motion of edges, 
which occupy only a small portion of the image plane, 
most stereo methods produce semi-dense depth maps representing 3D scene edges.
\GTS{} produces modest results, albeit with many outliers.
SGM has the most dense results because its regularizer fills in depth estimates in regions where the data fidelity term (time-image consistency) is not dominant.
ESVO gives remarkable results in terms of accuracy and completeness, thus showing the effectiveness of its probabilistic inverse depth filters.
Finally, our methods produce the best results: visually similar to ESVO but with finer details, thus being able to resolve more and finer edges in the scene. 

\Cref{tab:sota:mvsec} summarizes the quantitative performance with the metrics defined in \cref{sec:experim:datasets} and on the same MVSEC sequences as~\cite{Zhou18eccv,Zhu18eccv,Tulyakov19iccv,Zhou20tro} (with ground truth depth).
The best result per column is highlighted in bold, and the second best is underlined. 
Detailed, per-sequence tables are provided in the Supplementary Material.
Contrary to previous works, we test on the entire sequences, consisting of 200~\si{\second} (110 M events).
The top part of \cref{tab:sota:mvsec} reports the results of the baseline methods, where ESVO is a top performer.
EMVS is slightly worse than ESVO, consistently in most metrics, which shows that sensible stereo depth estimation (ESVO's Student-$t$ filters) is beneficial to gain accuracy and reduce the number of outliers with respect to the monocular case.
The bottom part of \cref{tab:sota:mvsec} reveals the results of several variations of Alg.~\ref{alg:fusion:time:stereo} (\eqref{eq:AlgTwoAAHH}-\eqref{eq:AlgTwoOtherFour}) with $N_s=2$ sub-intervals.
All accuracy and outlier metrics of Alg.~\ref{alg:fusion:stereo} are significantly better than those of ESVO and EMVS, demonstrating the effectiveness of our fusion approaches: outperforming the state of the art and quantifying the gap between monocular and stereo methods.

Regarding completion, Alg.~\ref{alg:fusion:stereo} recovers fewer points than ESVO.
This has a natural explanation: ESVO generates several depth estimates per second that are propagated and fused. 
Even using ground truth poses, the estimates are noisy, and so they transfer to the fused image plane producing thick edges (\cref{fig:mapping:depthmaps-grid}).
By contrast, our stereo method generates depth estimates at the AGT thresholding step, which is called just once, and therefore generates thinner edges than ESVO (finer details and well distributed at scene edges, as shown in \cref{fig:mapping:depthmaps-grid}).
To justify that lower completion values are not an issue, we applied a 4-neighbor morphological filter (MF) to dilate the mask of the depth map produced by AGT, and filled in the depth values using the center pixels. 
This almost quadrupled the number of reconstructed points (from 0.81M to 3M) while had a minimal effect on accuracy (row ``MF'' in \cref{tab:sota:mvsec}), and therefore reduces the importance of comparing completion values given by very different algorithms.
This also aligns with ideas in semi-dense and sparse SLAM, where fewer but more accurate points are preferred for several reasons: 
to have better distributed points on the image plane and to reduce the computational load (i.e., increase efficiency and speed) \cite{Forster17troSVO}.

Variants $A_t\circ H_c$ and $A_c\circ A_t$ are also top performing. 
Despite the abundance of events accumulated in the DSI, \cref{tab:sota:mvsec} quantifies a gap between Alg.~\ref{alg:fusion:stereo} and $A_c\circ A_t$ 
(the gap between the arithmetic mean and more concave fusion functions like $H$ would be most noticeable if less data was used, as in \cref{tab:mvsec:across-cam-small-data}).
The last row of \cref{tab:sota:mvsec} is discussed in \cref{sec:experim:shuffled}.

\emph{Driving sequences}.
The tests on the outdoor MVSEC sequences did not give good results because the camera baseline is very small compared to the scene depth, so geometrically the data is poor for 3D reconstruction. 
As noticed in \cite{Zhu18eccv}, most points in those sequences are beyond the depth resolved by a disparity of 1 pix.
Instead, we show results of our method on driving sequences from the DSEC dataset (\cref{sec:experim:dsec}), which has a larger baseline of 60~\si{\centi\meter}.

%% file: chapters/04_experim_time_fusion.tex
\subsection{Temporal Fusion Experiments}
\label{sec:experim:timefusion}

\subsubsection{Stereo}
\input{floats/tab_time_fusion}

\Cref{fig:mapping:depthmaps-grid} and \cref{tab:mvsec:timefusion} show the effect of temporal fusion.
Using as few as $N_s=2$ sub-intervals in Alg.~\ref{alg:fusion:time:stereo} already delivers gains compared to the state of the art.
While the optimal choice of the number of subintervals $N_s$ depends on many factors, such as the number of events processed (duration of the intervals), the camera motion, etc. we found that using $N_s \in [2, 8]$ gives satisfactory results. 
This is reported in \cref{tab:mvsec:timefusion}, which is an ablation study of Alg.~\ref{alg:fusion:time:stereo} $A_t\circ H_c$ with respect to $N_s$.
There is a clear trend: accuracy and completion values increase with $N_s$. 
However, memory and complexity also increases (linearly with $N_s$, see \eqref{eq:complexityAlgTwo}). 
Comparing the values in \cref{tab:sota:mvsec} and \cref{tab:mvsec:timefusion}, we notice that the improvement due to temporal fusion is not as pronounced as that due to stereo parallax (for example, the median error improves \SI{32}{\percent} from EMVS to stereo $A_t\circ H_c$ ($N_s=2$), and \SI{5.02}{\percent} from $A_t\circ H_c$ $N_s=2$ to $N_s=8$).

\input{floats/fig_emvs_vs_timeemvs}

\subsubsection{Monocular}
\label{sec:experim:monofusion}

As a by-product, temporal fusion using the $H$-mean is a simple modification that can be applied to lightly improve the monocular method \cite{Rebecq18ijcv}, specially when little data is available.
While this is not the focus of the paper, we provide an example: \cref{fig:experim:mono_time_fusion} shows the effect of monocular temporal fusion on the same slider rock-plane sequence as \cref{fig:experim:monovsstereo}.
It compares the evolution of DSIs with and without temporal fusion.
As expected, with $H$-mean fusion the DSI converges faster to the 3D structure and deemphasizes the locations of spurious ray intersections visible in the unfused DSI.

The effect of temporal fusion is dramatic if we compare the depth maps obtained from each subinterval with the depth map obtained after fusion.
This is illustrated in \cref{fig:temporalfusion}, where an interval of the \upennflyThree{} sequence \cite{Zhu18ral} is divided into $N_s=4$ sub-intervals. 
The first four columns depict depth maps produced by EMVS (row 1) and Alg.~\ref{alg:fusion:stereo} (row 2) applied to individual sub-intervals (each with $\approx$56k events per camera). 
These depth maps are noisy; however, when the DSIs are temporally fused and depth is extracted (Alg.~\ref{alg:fusion:time:stereo}), the final depth maps are remarkably cleaner (last column).
Also, the fused stereo depth map has fewer outliers than the monocular~one.
\input{floats/fig_temporal_fusion}

\input{chapters/04_shuffled}

%% file: floats/tab_time_fusion.tex
\begin{table*}[t]
\centering
\begin{adjustbox}{width=.8\textwidth}
\setlength{\tabcolsep}{3pt}
\begin{tabular}{l*{1}{S[table-format=2.2,table-number-alignment=center]}
*{4}{S[table-format=1.2,table-number-alignment=center]}
*{4}{S[table-format=2.2,table-number-alignment=center]}
*{1}{S[table-format=1.2,table-number-alignment=center]}}
\toprule 
$N_s$ & \text{Mean Err} & \text{Median Err} & \text{bad-pix} & \text{SILog Err} & \text{AErrR} & \text{log RMSE} & \text{$\delta < 1.25$} & \text{$\delta < 1.25^2$} & \text{$\delta < 1.25^3$} & \text{\#Points}\\
& \text{[cm] $\downarrow$} & \text{[cm] $\downarrow$} & \text{[\%] $\downarrow$} & \text{$\times 100 \downarrow$} & \text{[\%] $\downarrow$} & \text{$\times 100 \downarrow$} & \text{[\%] $\uparrow$} & \text{[\%] $\uparrow$} & \text{[\%] $\uparrow$} & \text{[million] $\!\uparrow$}\\
\midrule
2   & 20.91578705 & 9.759206544 & 1.662014008 & 1.8631732 & 8.045587963 & 13.81175316 & 94.3947297 & 97.80300225 & 99.14006845 & 1.14\\
4 & 20.49214892 &	9.576383795 &	1.628666443 & 1.94281938	& 7.882431548 &	13.65393651 & 94.64342103 &	97.81668021 &	99.15083254 & 1.19\\
8 & \bnum{19.79617585}	& \bnum{9.273273736} &	\bnum{1.563687566} & \bnum{1.836569347} &	\bnum{7.631628374}	& \bnum{13.31470568} & \bnum{94.99030025}	& \bnum{97.89968493}	& \bnum{99.19392224} & \bnum{1.20}\\
\bottomrule
\end{tabular}
\end{adjustbox}
\caption{\label{tab:mvsec:timefusion}Sensitivity of $A_t \circ H_c$ (Alg.~\ref{alg:fusion:time:stereo}) with respect to the number of subintervals $N_s$. 
Continuation of \cref{tab:sota:mvsec}.
}
\end{table*}

%% file: floats/fig_emvs_vs_timeemvs.tex
\def\figWidth{0.31\columnwidth}
\begin{figure}[t]
	\centering
    {\small
    \setlength{\tabcolsep}{1pt}
	\begin{tabular}{
	>{\centering\arraybackslash}m{0.35cm} 
	>{\centering\arraybackslash}m{\figWidth}
	>{\centering\arraybackslash}m{\figWidth} 
	>{\centering\arraybackslash}m{\figWidth}}
		& \SI{0.1}{\second} & \SI{0.2}{\second} & \SI{1.0}{\second} \\\addlinespace[0.2ex]

		\rotatebox{90}{\makecell{w/o fusion \cite{Rebecq18ijcv}}}
		&\includegraphics[width=\linewidth]{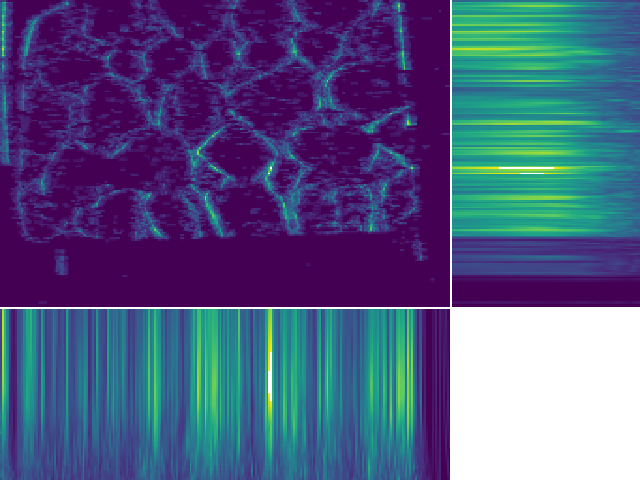}
		&\includegraphics[width=\linewidth]{images/slider_plane/dsi200ms_mono.png}
		&\includegraphics[width=\linewidth]{images/slider_plane/dsi1s_mono.png}
		\\\addlinespace[-0.3ex]

		\rotatebox{90}{\makecell{w/ fusion}}
		&\includegraphics[width=\linewidth]{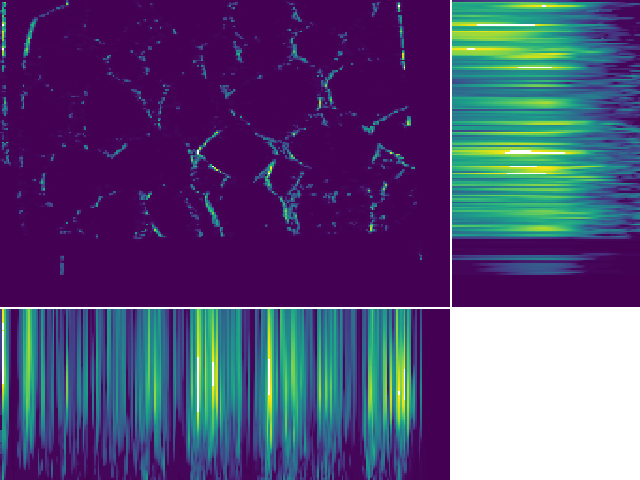}
		&\includegraphics[width=\linewidth]{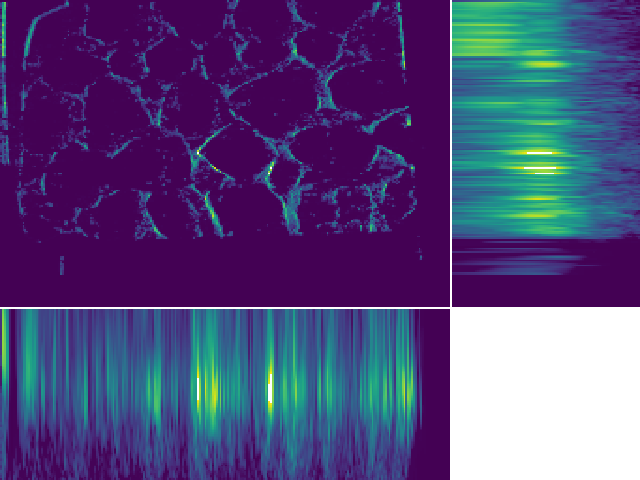}
		&\includegraphics[width=\linewidth]{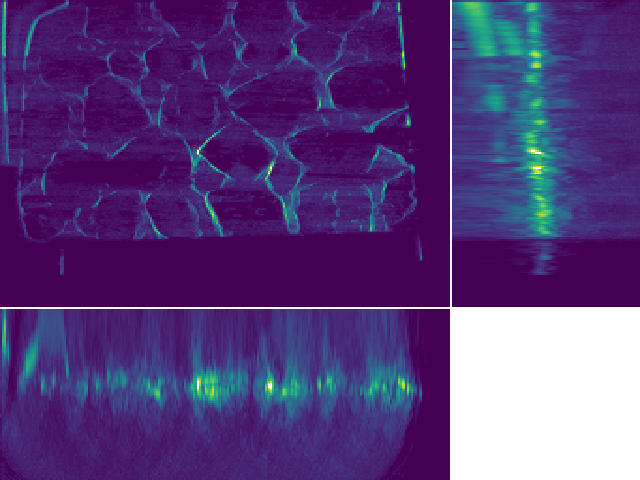}
		\\\addlinespace[-0.3ex]
		
	\end{tabular}
	}
	\caption{\label{fig:experim:mono_time_fusion}%
	\emph{Effect of temporal fusion} ($N_s=4$ sub-intervals) on monocular 3D reconstruction.
	Rocks scene in \cref{fig:experim:monovsstereo}.
	}
\end{figure}

%% file: floats/fig_temporal_fusion.tex
\def\figWidth{0.19\linewidth}
\begin{figure*}[t]
	\centering
    {\small
    \setlength{\tabcolsep}{1pt}
	\begin{tabular}{
	>{\centering\arraybackslash}m{0.4cm} 
	>{\centering\arraybackslash}m{\figWidth}
	>{\centering\arraybackslash}m{\figWidth} 
	>{\centering\arraybackslash}m{\figWidth} 
	>{\centering\arraybackslash}m{\figWidth}
	>{\centering\arraybackslash}m{\figWidth}}
		& Sub-interval 1 & Sub-interval 2 & Sub-interval 3 & Sub-interval 4 & \textbf{Fused} \\\addlinespace[.3ex]

		\rotatebox{90}{\makecell{Monocular}}
		&\gframe{\includegraphics[width=\linewidth]{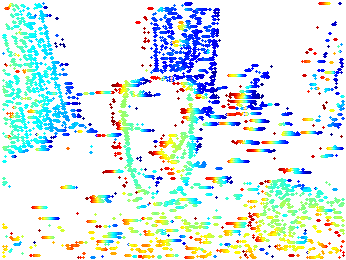}}
		&\gframe{\includegraphics[width=\linewidth]{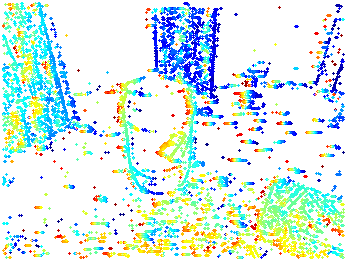}}
		&\gframe{\includegraphics[width=\linewidth]{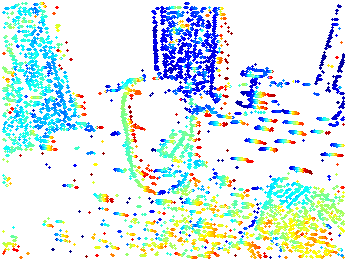}}
		&\gframe{\includegraphics[width=\linewidth]{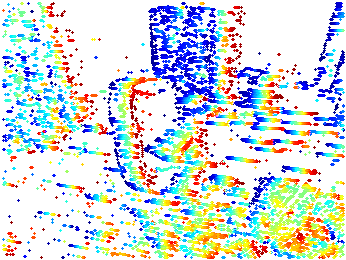}}
		&\gframe{\includegraphics[width=\linewidth]{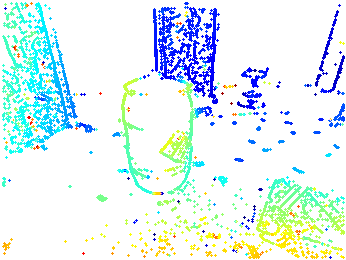}}
		\\
		
		\rotatebox{90}{\makecell{Stereo $A_t\circ H_c$}}
		&\gframe{\includegraphics[width=\linewidth]{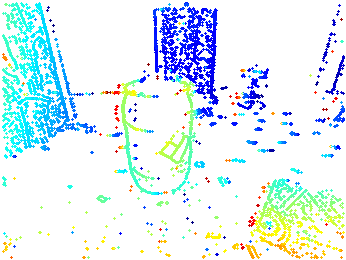}}
		&\gframe{\includegraphics[width=\linewidth]{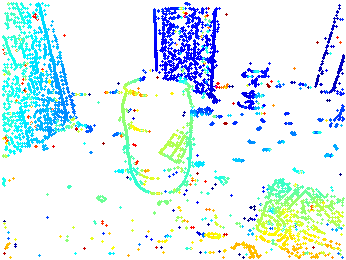}}
		&\gframe{\includegraphics[width=\linewidth]{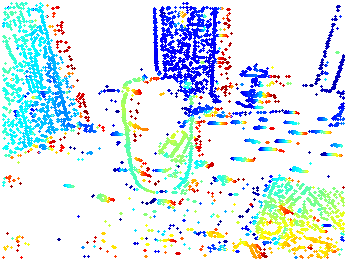}}
		&\gframe{\includegraphics[width=\linewidth]{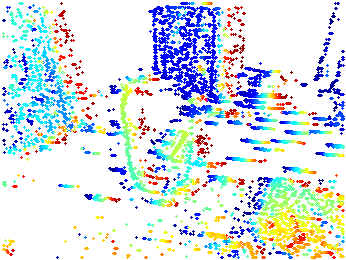}}
		&\gframe{\includegraphics[width=\linewidth]{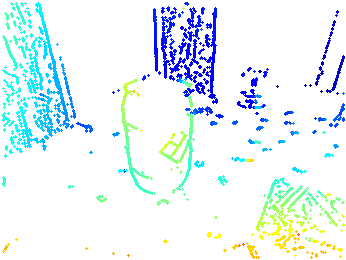}}
		\\

	\end{tabular}
	}
	\caption{\label{fig:temporalfusion}
	\emph{Effect of temporal fusion} on the obtained depth maps.
 	Depth is color coded as in \cref{fig:mapping:depthmaps-grid}.
	}
\end{figure*}

%% file: chapters/04_shuffled.tex
\subsubsection{Stereo Fusion from Shuffled Sub-intervals}
\label{sec:experim:shuffled}

\input{floats/fig_results_shuffled}

Arranging events in time subintervals allows us to question the event simultaneity assumption for depth estimation.
\Cref{fig:shuffled} shows the results of applying the system in \cref{fig:block-diagram} with the shuffling block enabled to the MVSEC and UZH sequences.
The shuffling block modifies line~4 in Alg~\ref{alg:fusion:time:stereo} to use \emph{different} subintervals for the $S_1$ fusion. 
The results are surprising: despite using non-corresponding subintervals (i.e., non-simultaneous events) for DSI fusion across cameras, the obtained depth- and confidence maps are very similar to those obtained with corresponding subintervals, with some added noise.
Hence, event simultaneity is not needed for stereo depth estimation with our system (\cref{fig:block-diagram}).
Only the similarity between the DSIs to be fused (intermediate ray density representations built by combining events and camera poses) is required.

Quantitatively, the last row of \cref{tab:sota:mvsec} informs about the depth errors and outliers incurred by shuffling. 
The differences with respect to the unshuffled case are small (slightly higher errors and outliers), which is remarkable given the quite diverse input events.

%% file: floats/fig_results_shuffled.tex
\def\figWidth{0.19\linewidth}
\begin{figure*}[t]
	\centering
    {\small
    \setlength{\tabcolsep}{1pt}
	\begin{tabular}{
	>{\centering\arraybackslash}m{0.4cm} 
	>{\centering\arraybackslash}m{\figWidth} 
	>{\centering\arraybackslash}m{\figWidth}
	>{\centering\arraybackslash}m{\figWidth}
	>{\centering\arraybackslash}m{\figWidth}
	>{\centering\arraybackslash}m{\figWidth}}
	
        & \rpgreader{} & \rpgbox{} & \rpgmonitor{} & \upennflyOne{} & \upennflyThree{}
        \\\addlinespace[0.3ex]

		\rotatebox{90}{\makecell{Depth}}
		&\gframe{\includegraphics[width=\linewidth]{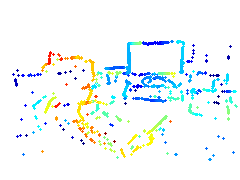}}
		&\gframe{\includegraphics[width=\linewidth]{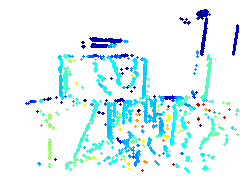}}
		&\gframe{\includegraphics[width=\linewidth]{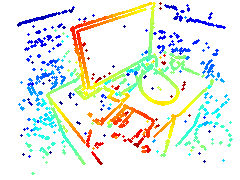}}
		&\gframe{\includegraphics[width=\linewidth]{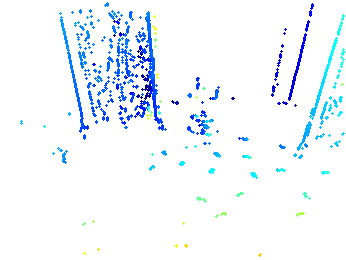}}
		&\gframe{\includegraphics[width=\linewidth]{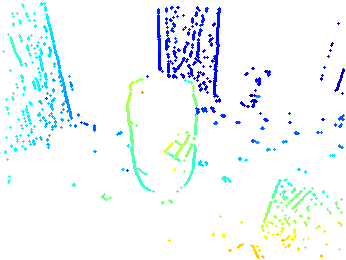}}
		\\
		
		\rotatebox{90}{\makecell{Confidence map}}
		&\gframe{\includegraphics[width=\linewidth]{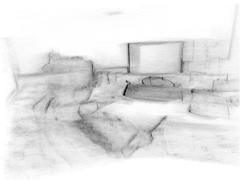}}
		&\gframe{\includegraphics[width=\linewidth]{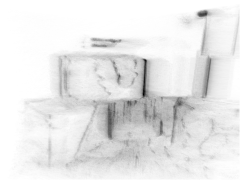}}	
		&\gframe{\includegraphics[width=\linewidth]{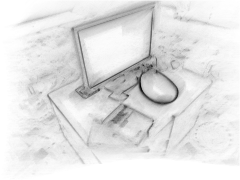}}
		&\gframe{\includegraphics[width=\linewidth]{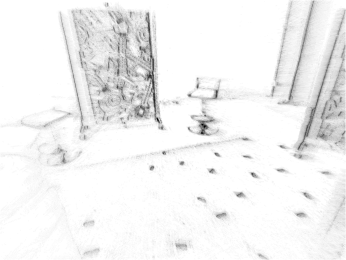}}	
		&\gframe{\includegraphics[width=\linewidth]{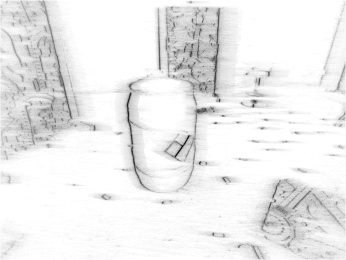}}
		\\
	\end{tabular}
	}
	\caption{\label{fig:shuffled}\emph{Event Simultaneity?}
	Fusion across cameras and time with shuffled time intervals ($N_s=2$). 
	Same scenes as \cref{fig:mapping:depthmaps-grid}.}
\end{figure*}

%% file: chapters/04_experim_dsec.tex
\subsection{Experiments on DSEC Driving Dataset}
\label{sec:experim:dsec}
\input{floats/fig_DSEC}

We also give results on sequences from the driving dataset DSEC \cite{Gehrig21ral}.
\Cref{fig:dsec} shows qualitative results. 
Driving scenarios are challenging for event-based sensors because forward motions typically produce considerably fewer events in the center of the image (where apparent motion is small) than in the periphery.
Forward motion is also not particularly amenable for 3D reconstruction methods compared to sideways motions. %
Nevertheless, our stereo method shows notable results on this dataset. 
As expected, more 3D points are recovered in the periphery than in the center of the image, except when the car is turning.

Quantitative results are summarized in \cref{tab:dsec}.
Because the amount of events recorded by the VGA-resolution Prophesee Gen3 cameras is exorbitant, the experiments are carried out on a subset of the dataset. 
We test Alg.~\ref{alg:fusion:stereo} and the baselines on 635 million events.
Similarly to the results on MVSEC, \cref{tab:dsec} indicates that Alg.~\ref{alg:fusion:stereo} has higher accuracy than ESVO and EMVS 
(at least \SI{18.65}{\percent} better in mean absolute error, and \SI{42.3}{\percent} in median absolute error).
ESVO has marginally higher inlier values (\SI{1.45}{\percent} for $\delta<1.25$).
The monocular method produces the largest errors (the mean and median absolute errors are 1.7-2.8$\times$ larger than those of stereo Alg.~\ref{alg:fusion:stereo}), and also a larger number of bad pixels and outliers.
We also apply a morphological dilation filter (MF), with similar conclusions as in \cref{tab:sota:mvsec}: the number of reconstructed points triples while the accuracy remains better than ESVO's.
Finally, we notice that the ground truth depth of DSEC is sparser than that of MVSEC. 
This is due to the increased pixel resolution (VGA size) and the fact that LiDAR points do not fill as many camera pixels (percentage-wise) as in lower resolution cameras.
Please see the accompanying video for a visual comparison between our method, ESVO, EMVS and GT.
\input{floats/tab_DSEC}

%% file: floats/fig_DSEC.tex
\def\figWidth{0.19\linewidth}
\begin{figure*}[t]
	\centering
    {\small
    \setlength{\tabcolsep}{1pt}
	\begin{tabular}{
	>{\centering\arraybackslash}m{0.4cm} 	
	>{\centering\arraybackslash}m{\figWidth} 
	>{\centering\arraybackslash}m{\figWidth} 
	>{\centering\arraybackslash}m{\figWidth} 
	>{\centering\arraybackslash}m{\figWidth} 
	>{\centering\arraybackslash}m{\figWidth}}

		\rotatebox{90}{\makecell{Depth (Alg~\ref{alg:fusion:stereo})}}
		&\gframe{\includegraphics[width=\linewidth]{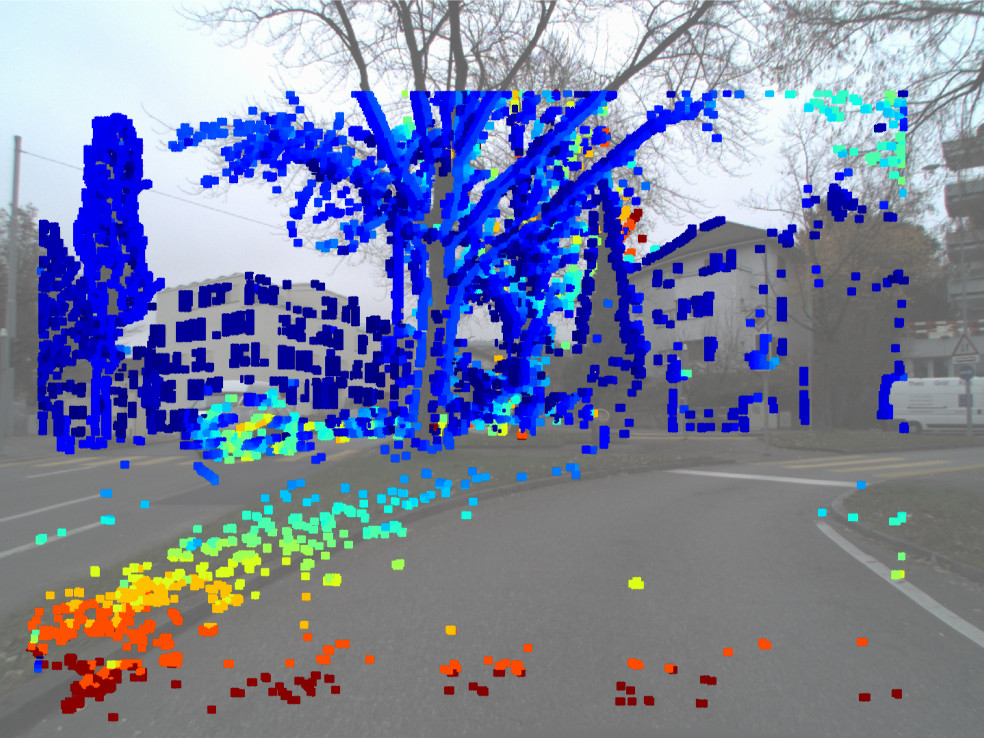}}
		&\gframe{\includegraphics[width=\linewidth]{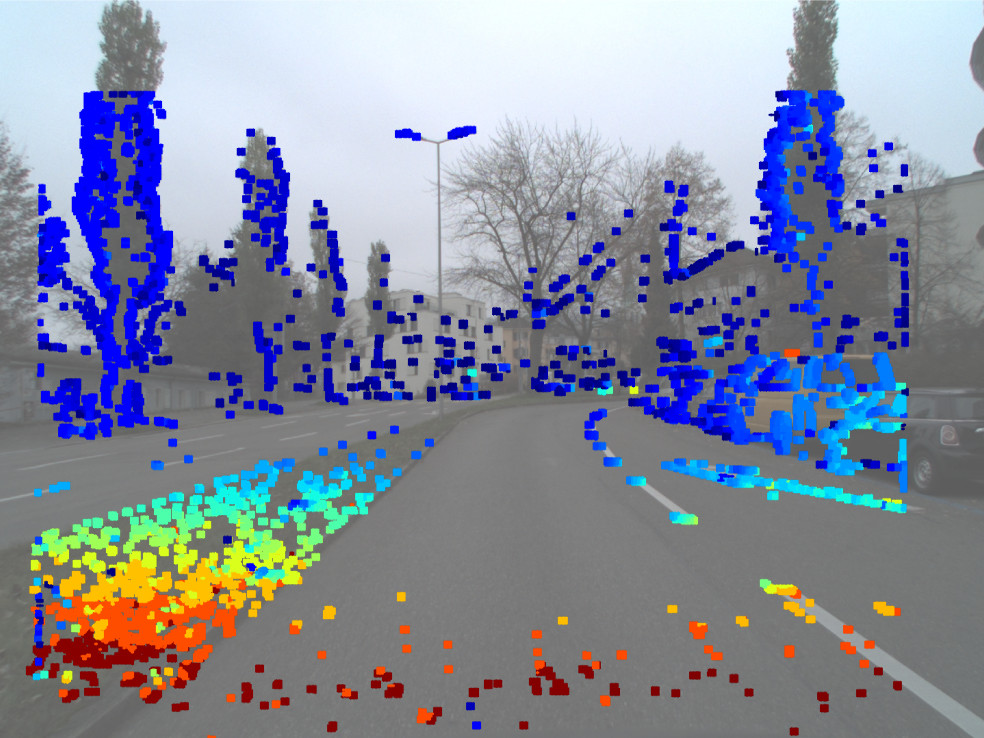}}
		&\gframe{\includegraphics[width=\linewidth]{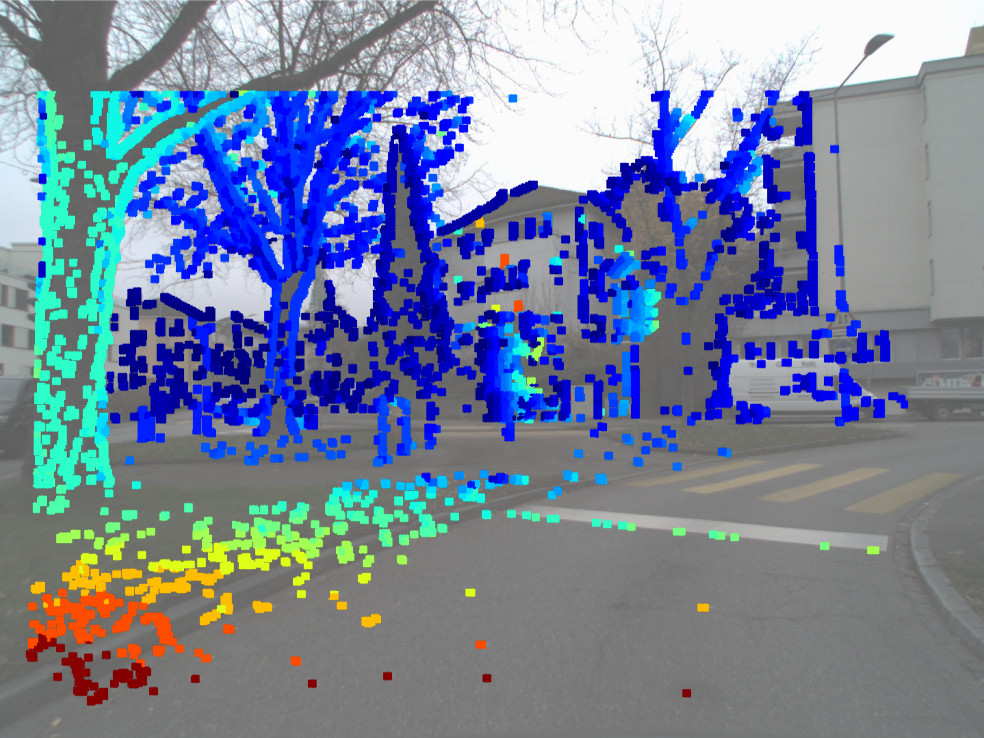}}
 		&\gframe{\includegraphics[width=\linewidth]{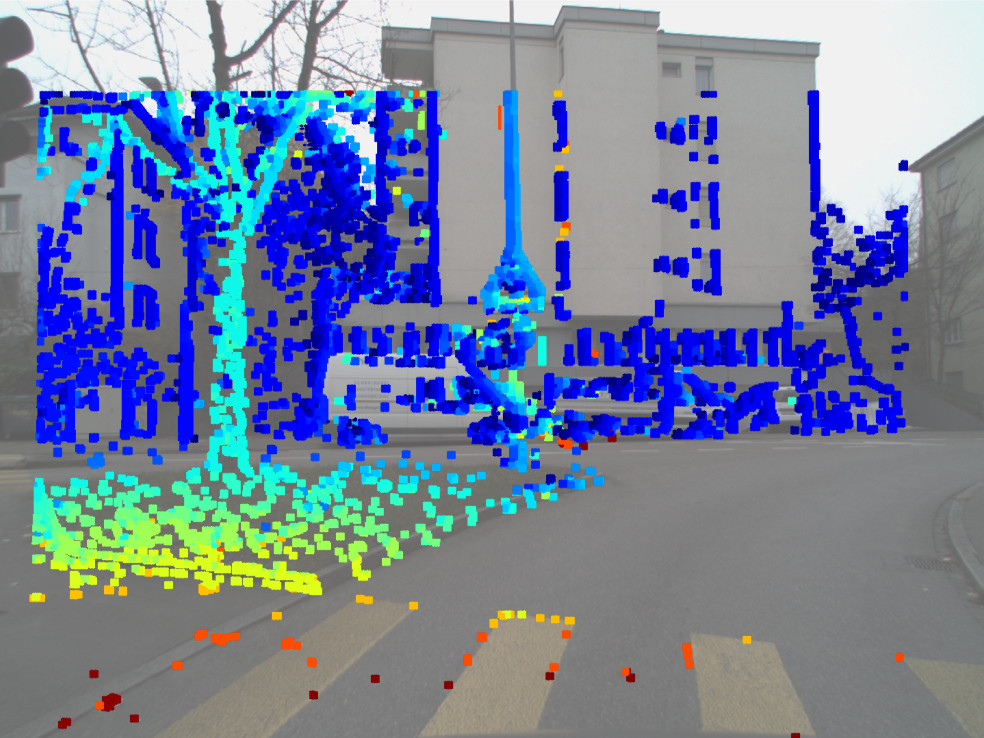}}
 		&\gframe{\includegraphics[width=\linewidth]{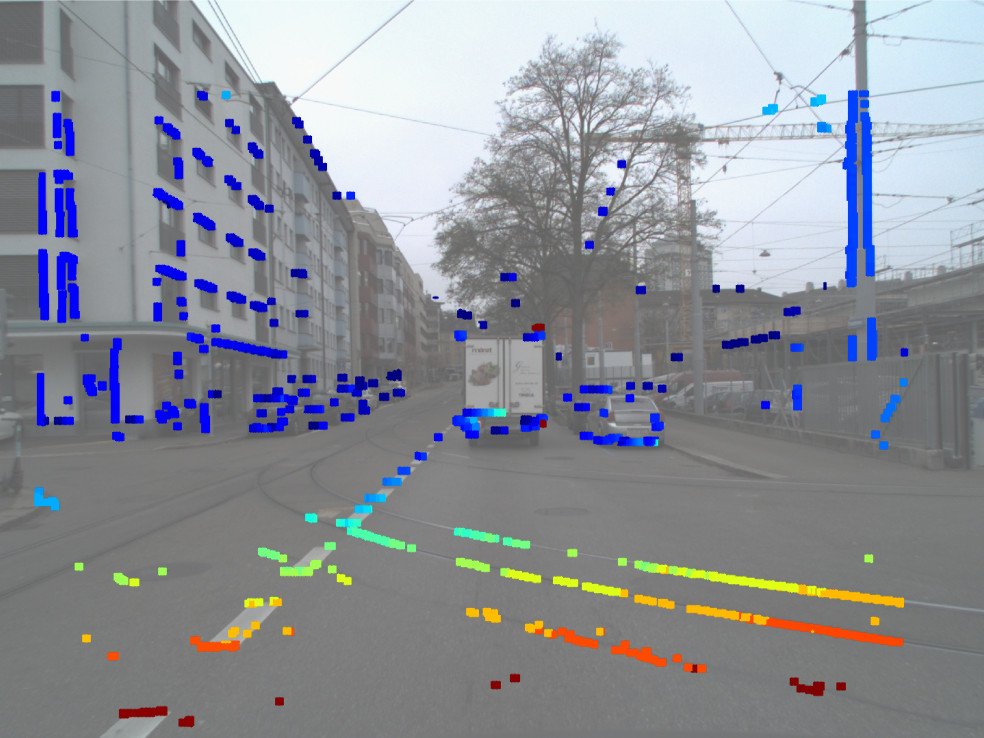}}
		\\
 		
 		\rotatebox{90}{\makecell{Confidence map}}
 		&\gframe{\includegraphics[width=\linewidth,trim={2cm 1cm 2cm 2cm},clip]{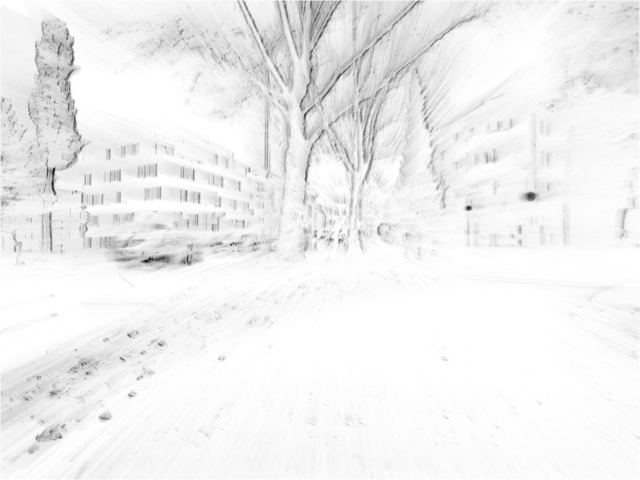}}
		&\gframe{\includegraphics[width=\linewidth,trim={2cm 1cm 2cm 2cm},clip]{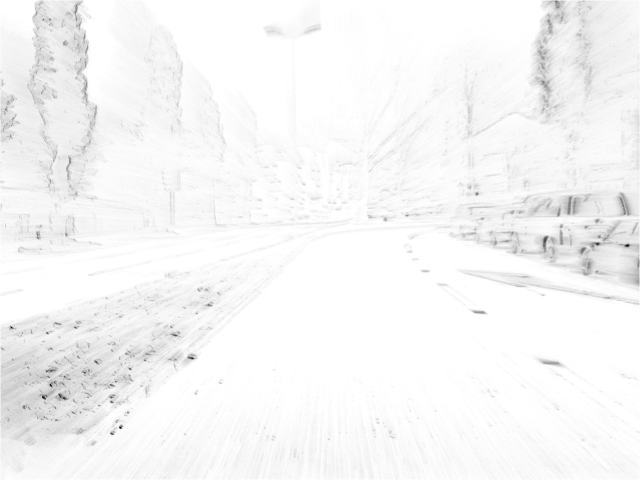}}
		&\gframe{\includegraphics[width=\linewidth,trim={2cm 1cm 2cm 2cm},clip]{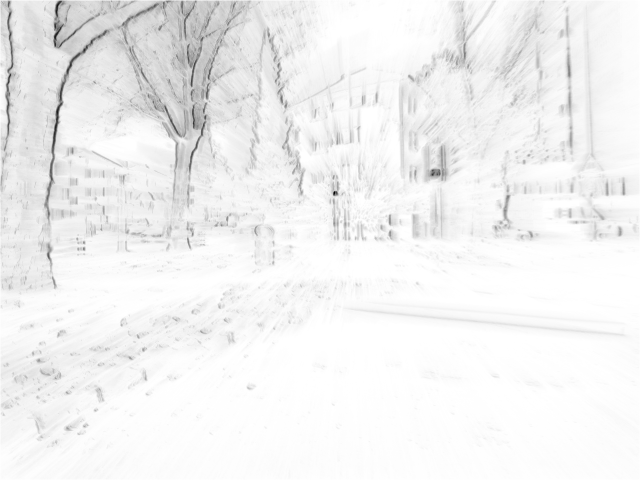}}
 		&\gframe{\includegraphics[width=\linewidth,trim={2cm 1cm 2cm 2cm},clip]{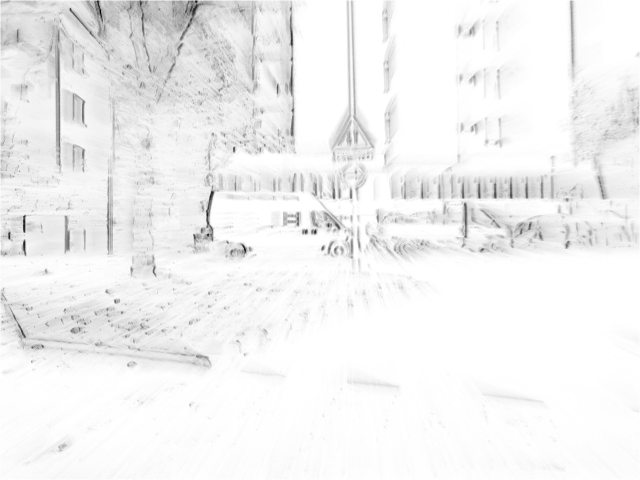}}
 		&\gframe{\includegraphics[width=\linewidth,trim={2cm 1cm 2cm 2cm},clip]{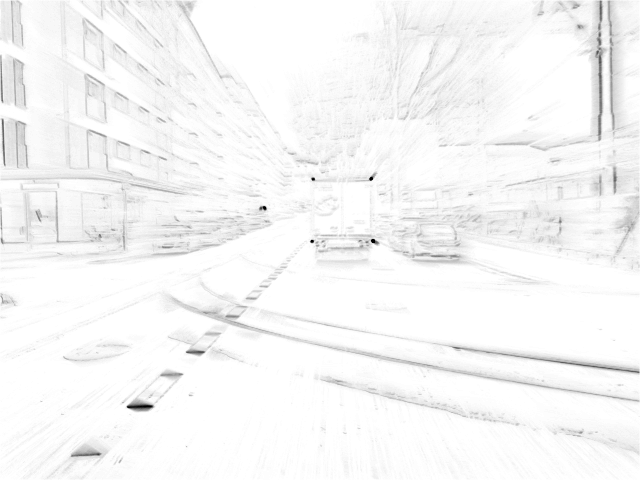}}
 		\\
	\end{tabular}
	}
	\caption{\label{fig:dsec}\emph{Results on DSEC data} \cite{Gehrig21ral}.
	Top row: Semi-dense depth maps (overlaid on color frames) estimated by Alg.~\ref{alg:fusion:stereo} on event packets of 200~\si{\milli\second}. 
	Depth is color-coded from red (close) to blue (far), in the range 4--200~\si{\meter}.
	Bottom row: confidence maps.
	}
\end{figure*}

%% file: floats/tab_DSEC.tex
\begin{table*}[t]
\centering
\begin{adjustbox}{width=\textwidth}
\setlength{\tabcolsep}{4pt}
\begin{tabular}{ll*{2}{S[table-format=1.2,table-number-alignment=center]}
*{1}{S[table-format=2.2,table-number-alignment=center]}
*{1}{S[table-format=1.2,table-number-alignment=center]}
*{5}{S[table-format=2.2,table-number-alignment=center]}
*{1}{S[table-format=1.2,table-number-alignment=center]}}
\toprule 
&Algorithm & \text{Mean Err} & \text{Median Err} & \text{bad-pix} & \text{SILog Err} & \text{AErrR} & \text{log RMSE} & \text{$\delta < 1.25$} & \text{$\delta < 1.25^2$} & \text{$\delta < 1.25^3$} & \text{\#Points}\\
& & \text{[m] $\downarrow$} & \text{[m] $\downarrow$} & \text{[\%] $\downarrow$} & \text{$\times 100 \downarrow$} & \text{[\%] $\downarrow$} & \text{$\times 100 \downarrow$} & \text{[\%] $\uparrow$} & \text{[\%] $\uparrow$} & \text{[\%] $\uparrow$} & \text{[million] $\!\uparrow$}\\
\midrule

& EMVS \cite{Rebecq18ijcv} (monocular) & 5.640576854 &	2.5161228 & 13.6829248 &	13.23340662 &	25.52277207 &	36.48934331 &	72.55970302 &	87.12135933 &	93.56143676 &	1.30717\\
& ESVO\cite{Zhou20tro} &	3.884085092 &	1.556005116 & 12.07867269 &	9.234173631 &	18.887026 &	30.80280662 &	\bfseries 84.52904261 &	\bfseries 92.57009852 &	\bfseries 95.63327073 &	\unum{3.40018}\\
& $H_c \circ A_t$ (Alg~\ref{alg:fusion:stereo}) & \bfseries 3.269906275 &	\bfseries 0.89702034 & \bfseries 10.75437193 & \bfseries 8.194367664 &	\bfseries 17.48164612 &	\bfseries 28.72803724 &	\unum{83.30346325} &	\unum{91.55994971} &	\unum{95.6170991}	& 1.253593\\
& $H_c \circ A_t$ (Alg~\ref{alg:fusion:stereo}) + MF & \unum{3.506127891} &	\unum{0.9579076767} &		\unum{11.81205097} &	\unum{8.893312443} &	\unum{18.8377646} &	\unum{29.99016074} &	81.71788985 &	90.68331057 &	95.07114855 &	\bfseries 3.830694\\
\bottomrule
\end{tabular}
\end{adjustbox}
\caption{Quantitative evaluation on the driving dataset DSEC (zurich04a sequence) with maximum ground truth depth 50~\si{\meter}. 
The methods are evaluated on 35s of stereo data, consisting of 635 million events and containing 350 ground truth depth maps.
Each depth map is computed using 0.2~\si{\s} of event data ($\approx3.5$ million events).
ESVO is executed fusing two depth maps generated at 10~\si{\Hz} (LiDAR rate), i.e., 0.2~\si{\s} of event data. 
MF: morphological filter. 
\label{tab:dsec}
}
\end{table*}

%% file: chapters/04_experim_tumvie.tex
\subsection{Experiments on TUM-VIE Dataset}
\label{sec:experim:tumvie}

We present depth estimation results using Alg.~\ref{alg:fusion:stereo} %
on the TUM-VIE dataset \cite{Klenk21iros}, the first public visual-inertial dataset with 1 Megapixel stereo event cameras (Prophesee Gen4 \cite{Finateu20isscc}). 
To the best of our knowledge, our work is the first to provide results on this new event-based dataset 
(the original paper presented the data but did not evaluate it on any event-based algorithm). 
Our experiments have served as a means to debug and fix the dataset.
Since the dataset has no ground truth depth, we only present qualitative results.

\Cref{fig:tumvie} presents results in indoor and outdoor sequences.
Indoor sequences recorded in a room have ground truth poses given by a motion capture (mocap) system (columns 1 and~2).
The indoor scene depth is small relative to the camera baseline (\SI{11.84}{\cm}). 
Nevertheless, our method does a notable job in recovering 3D structure, with reduced number of outliers and clean the depth maps. 
Having set the reference view of the fused DSI on one camera trajectory, the large baseline makes the event rays back-projected from the other camera appear nearly parallel, which does not favor fusion.

For sequences recorded outside the mocap room, we computed ground truth poses using Basalt's VIO \cite{Usenko20ral} on the stereo frames and IMU. 
The last two columns of \cref{fig:tumvie} depict the performance of %
our stereo method on such sequences.
The space-sweeping method to build the DSIs works best with sideways translations that produce parallax necessary for the convergence of the back-projected rays.
However, majority of the non-mocap sequences comprise forward camera motions, which contribute little parallax and also produce fewer events; hence they are not amenable for 3D reconstruction. %
The forward motion and the lower quality of camera poses lead to an overall poorer reconstruction quality compared to the sequences in the mocap room. 
On the other hand, the stereo setup is able to exploit the camera baseline as additional parallax for 3D reconstruction.
In the accompanying video we provide a visual comparison between our method and ESVO on the TUM-VIE dataset, 
showing that our method performs better at higher resolutions than ESVO.

We noticed that the \emph{calibrationA} sequences (\emph{skate-easy}, \emph{desk2}) produced better results than the \emph{calibrationB} sequences, which leads us to believe that the calibration errors were significant in the latter.
We also compared the results of our method with two different sources of ground truth poses (the mocap system and Basalt), 
and observed no significant differences in the depth- and confidence maps.
Hence, we concluded that ($i$) the poses from Basalt may be considered as accurate as the mocap for short time intervals (e.g., 0.5~\si{\second}, with $\approx$10M events), 
and ($ii$) there is robustness to noise: our methods provide reasonable depth estimates using poses from a VIO (non-mocap) algorithm.

Event cameras offer advantages over frame-based cameras to handle HDR scenes, as demonstrated in \cref{fig:hdr}. 
In the first row, the grayscale frame is overexposed in the outdoor area, whereas the events capture the floor tiles and garden scene well.
In the second row, the end of the corridor is underexposed in the frame, whereas the events capture the whole scene.
Our stereo method correctly estimates depth in all regions due to the HDR capabilities of the events. 

\input{floats/fig_tumvie_combined}
\input{floats/fig_HDR}

%% file: floats/fig_tumvie_combined.tex
\def\figWidth{0.24\linewidth}
\begin{figure*}[t]
	\centering
    {\small
    \setlength{\tabcolsep}{1pt}
	\begin{tabular}{
	>{\centering\arraybackslash}m{0.4cm} 
	>{\centering\arraybackslash}m{\figWidth} 
	>{\centering\arraybackslash}m{\figWidth}
	>{\centering\arraybackslash}m{\figWidth}
	>{\centering\arraybackslash}m{\figWidth}}
 	 
 	 & \multicolumn{2}{c}{With ground truth poses}
  	 & \multicolumn{2}{c}{Without ground truth poses}\\
 	 \cmidrule(l{1mm}r{1mm}){2-3} \cmidrule(l{1mm}r{1mm}){4-5}
        & \emph{6dof} & \emph{desk2} & \emph{skate-easy} & \emph{bike-easy}\\

		\rotatebox{90}{\makecell{Depth (Alg~\ref{alg:fusion:stereo})}}
		&\gframe{\includegraphics[width=\linewidth]{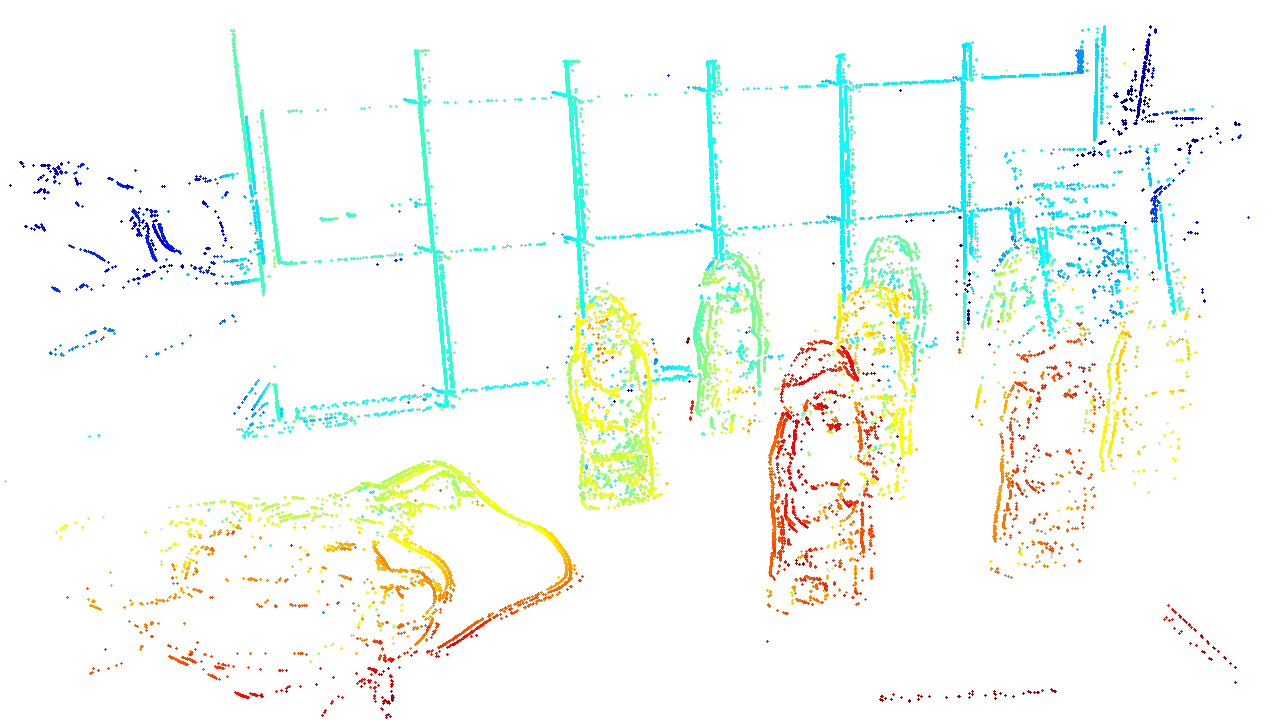}}
		&\gframe{\includegraphics[width=\linewidth]{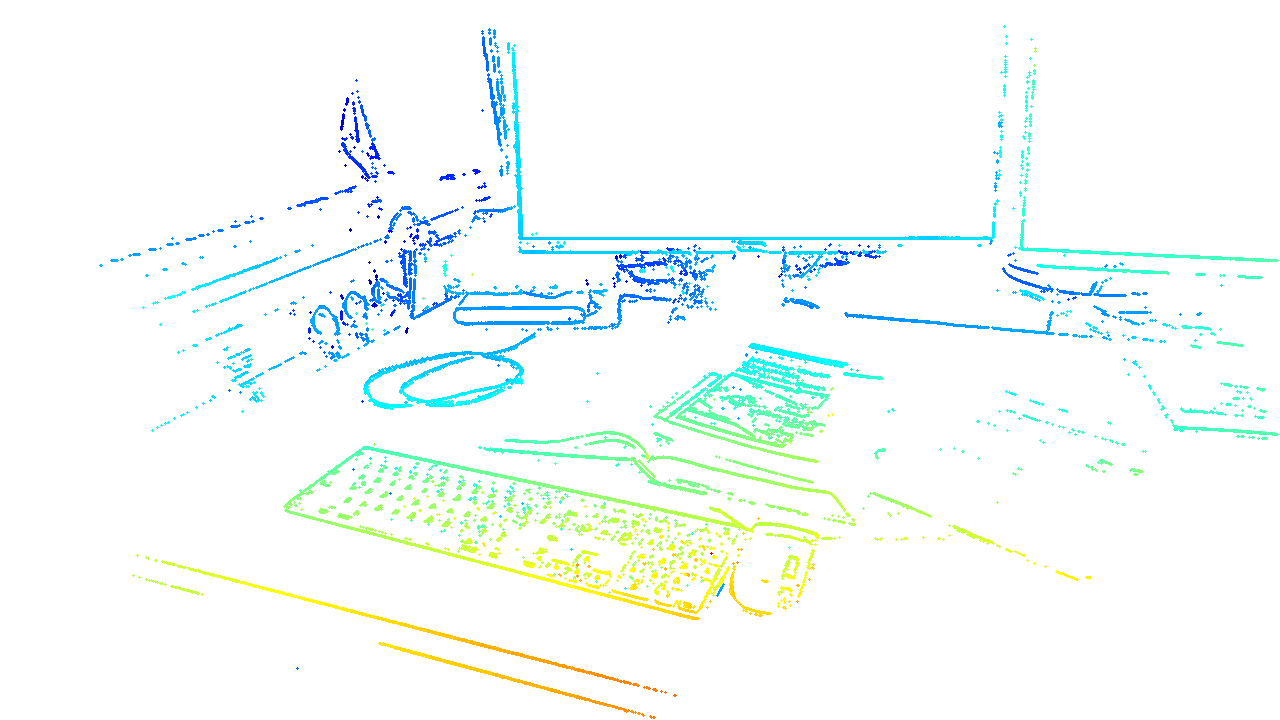}}
		&\gframe{\includegraphics[width=\linewidth]{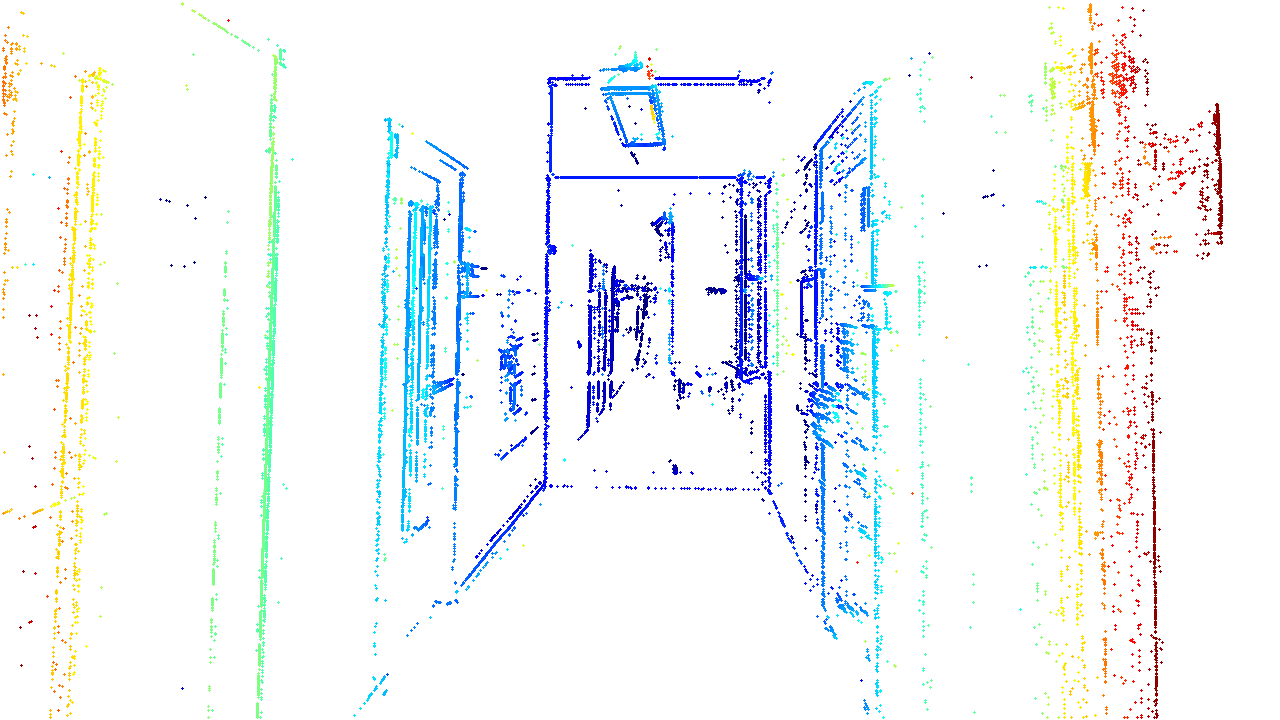}}
		&\gframe{\includegraphics[width=\linewidth]{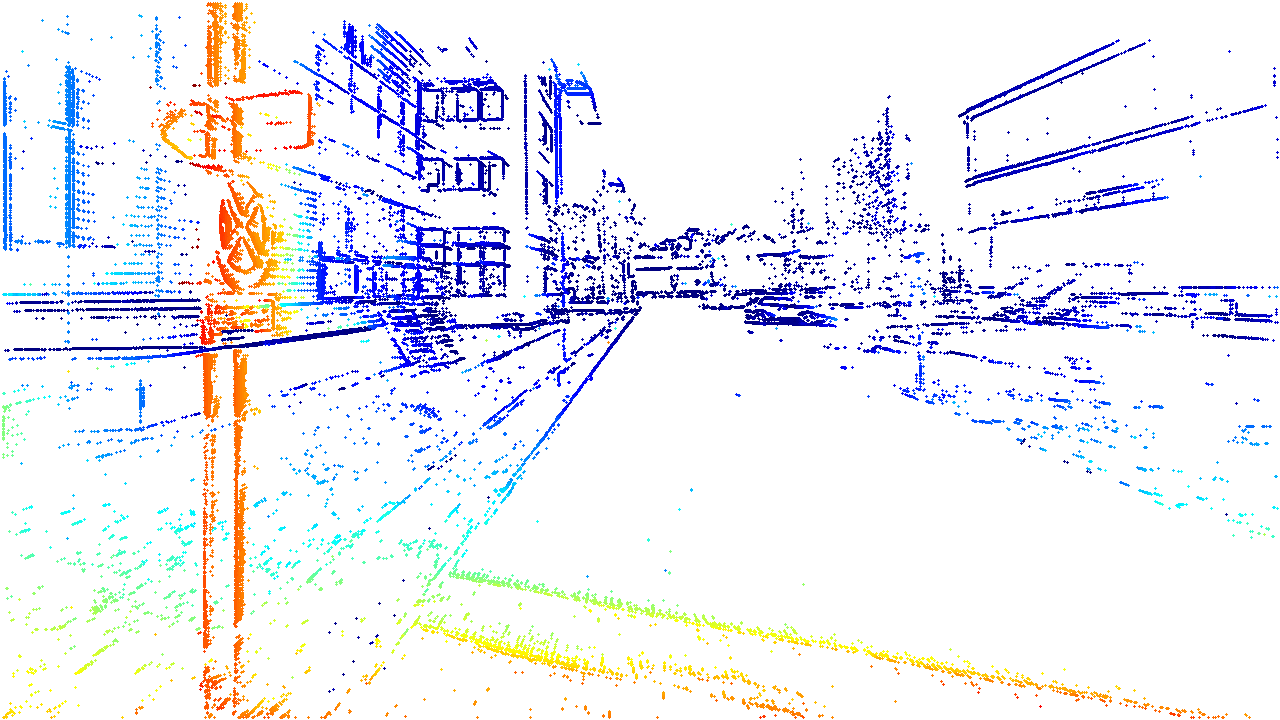}}
		\\
		
		\rotatebox{90}{\makecell{Fused confidence}}
		&\gframe{\includegraphics[width=\linewidth]{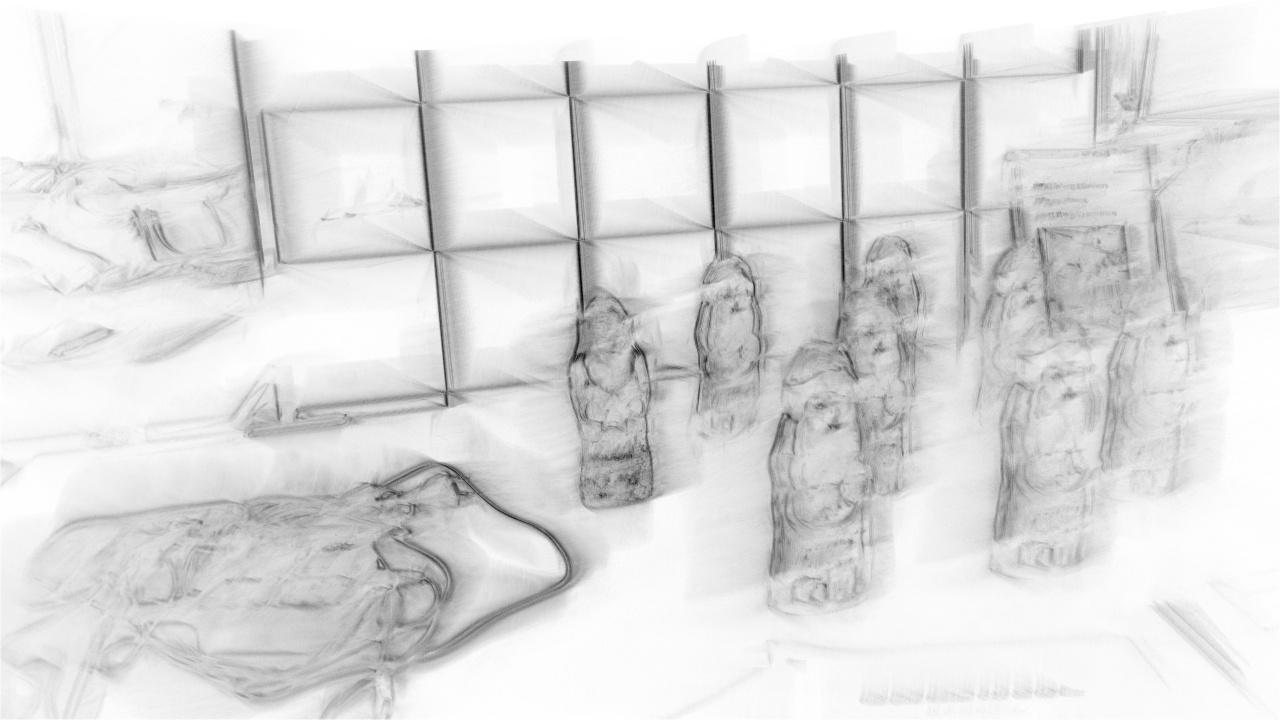}}
		&\gframe{\includegraphics[width=\linewidth]{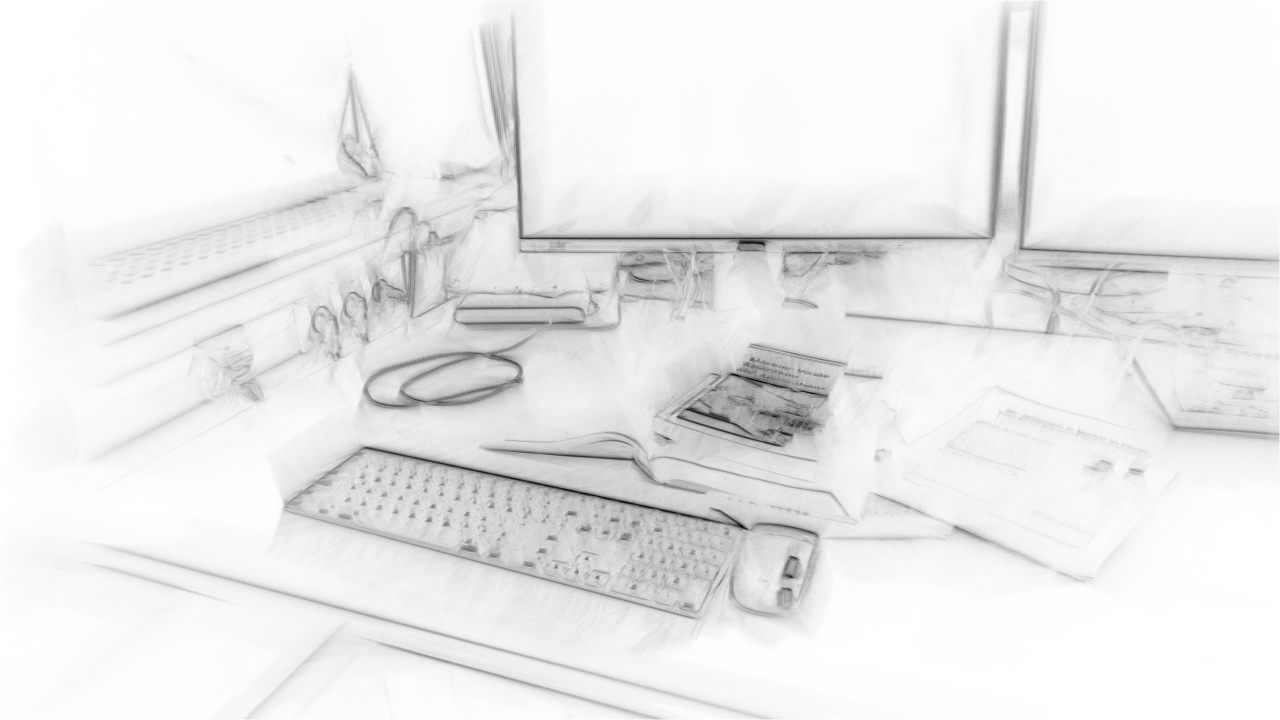}}
		&\gframe{\includegraphics[width=\linewidth]{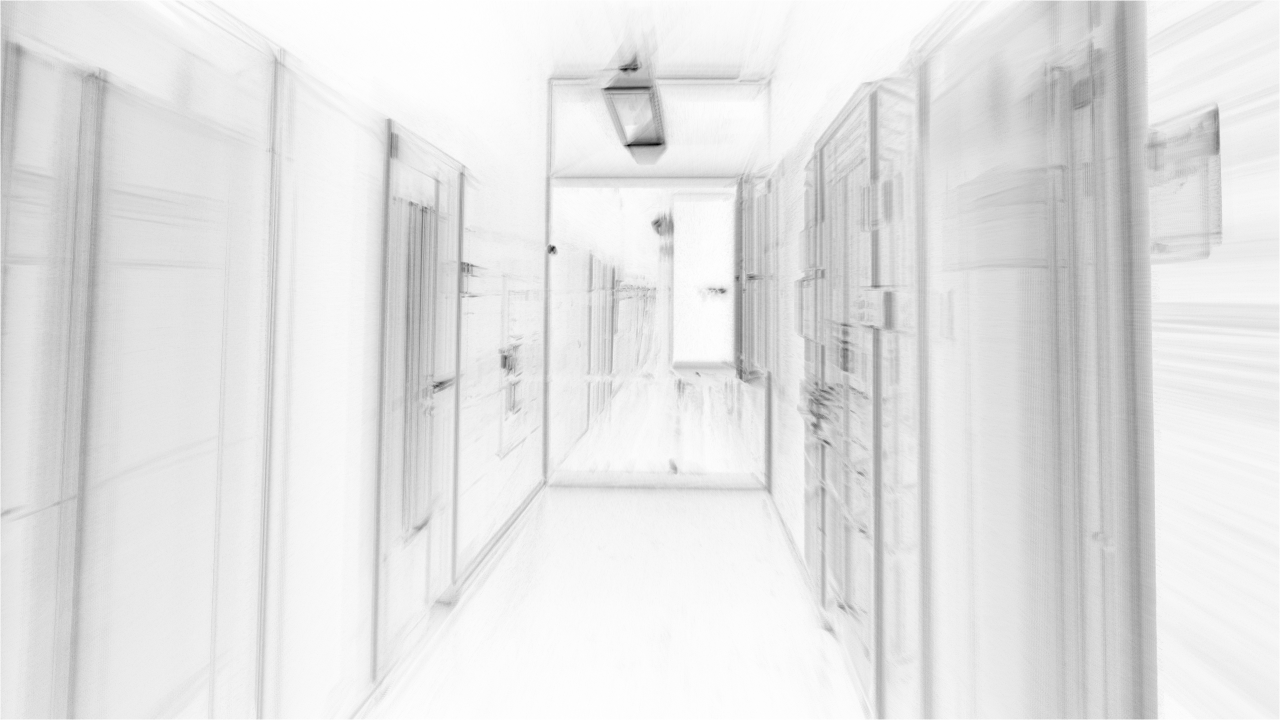}}
		&\gframe{\includegraphics[width=\linewidth]{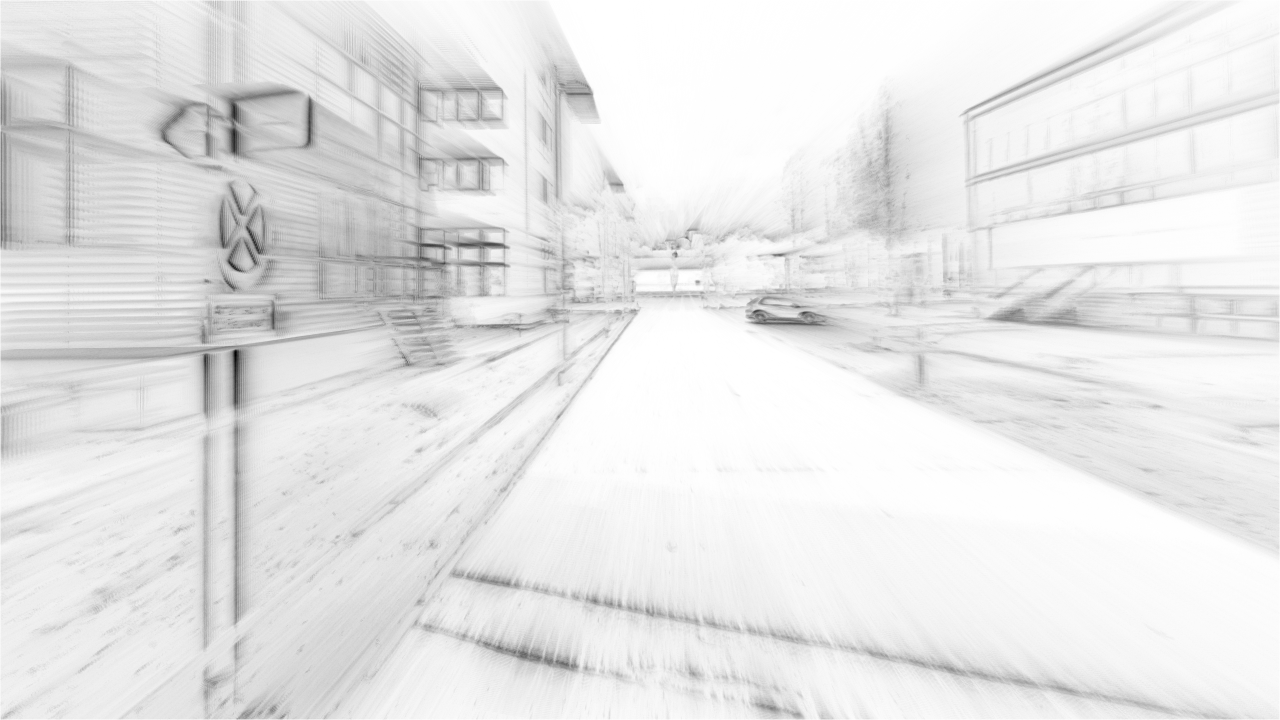}}		
		\\

	\end{tabular}
	}
	\caption{\label{fig:tumvie} \emph{Depth estimation using 1Mpix event cameras.} 
	Depth estimated using Alg~\ref{alg:fusion:stereo} 
	on \SI{0.5}{\second} intervals of data from TUM-VIE \cite{Klenk21iros}. 
	Depth maps are color-coded from red (near) to blue (far).
	The range is 0.45--4 \si{\meter} for indoor sequences in the motion-capture room (\emph{6dof}, \emph{desk2}),
	1--20 \si{\meter} for the corridor sequence \emph{skate-easy}
	and 3--200 \si{\meter} for the outdoor sequence \emph{bike-easy}.
	}
\end{figure*}

%% file: floats/fig_HDR.tex
\def\figWidth{0.243\textwidth}
\begin{figure*}[t]
	\centering
    {\small
    \setlength{\tabcolsep}{2pt}
	\begin{tabular}{
	>{\centering\arraybackslash}m{\figWidth}
	>{\centering\arraybackslash}m{\figWidth}
	>{\centering\arraybackslash}m{\figWidth}
	>{\centering\arraybackslash}m{\figWidth}}
	     Frame (left camera) & Events (left camera) & Confidence map & Depth map\\
		\gframe{\includegraphics[trim={0px 130px 0 200px},clip,width=\linewidth]{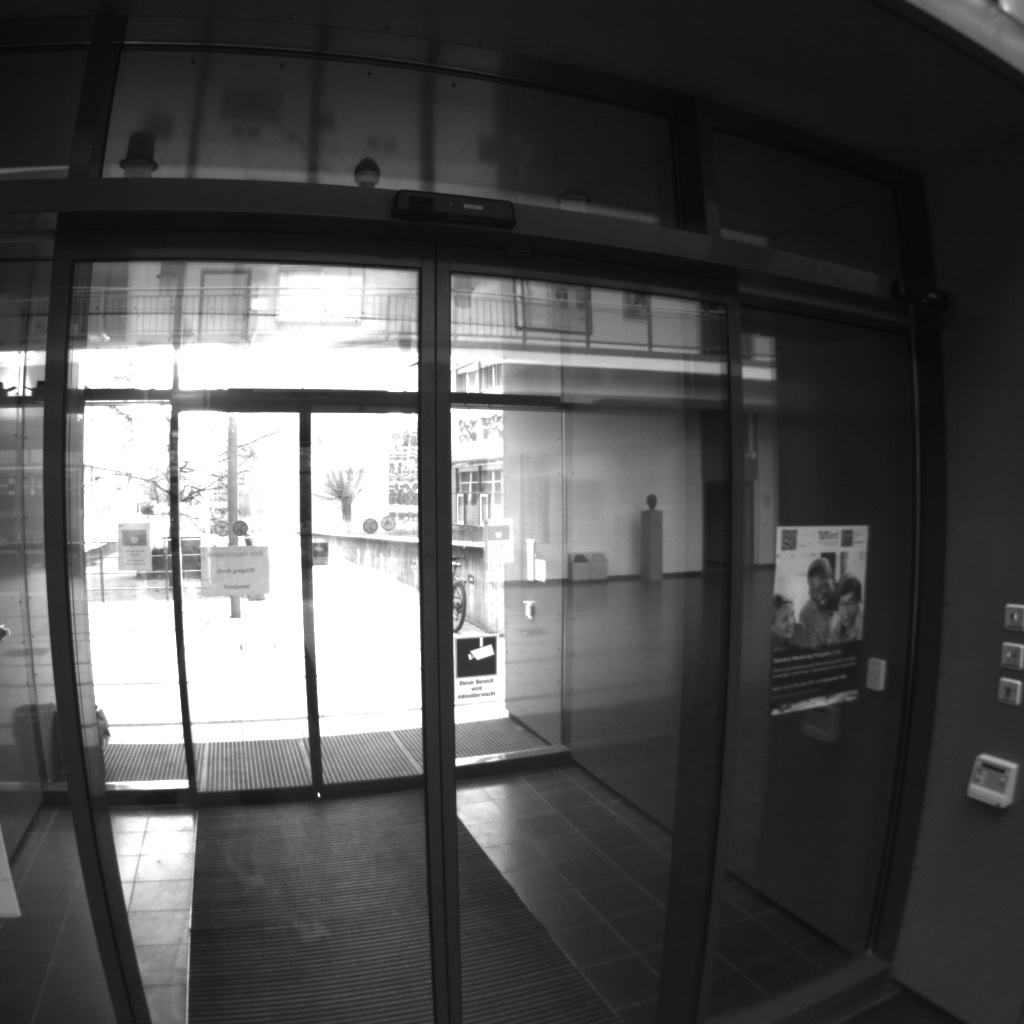}}
		&\gframe{\includegraphics[trim={0px 0 150px 0},clip,width=\linewidth]{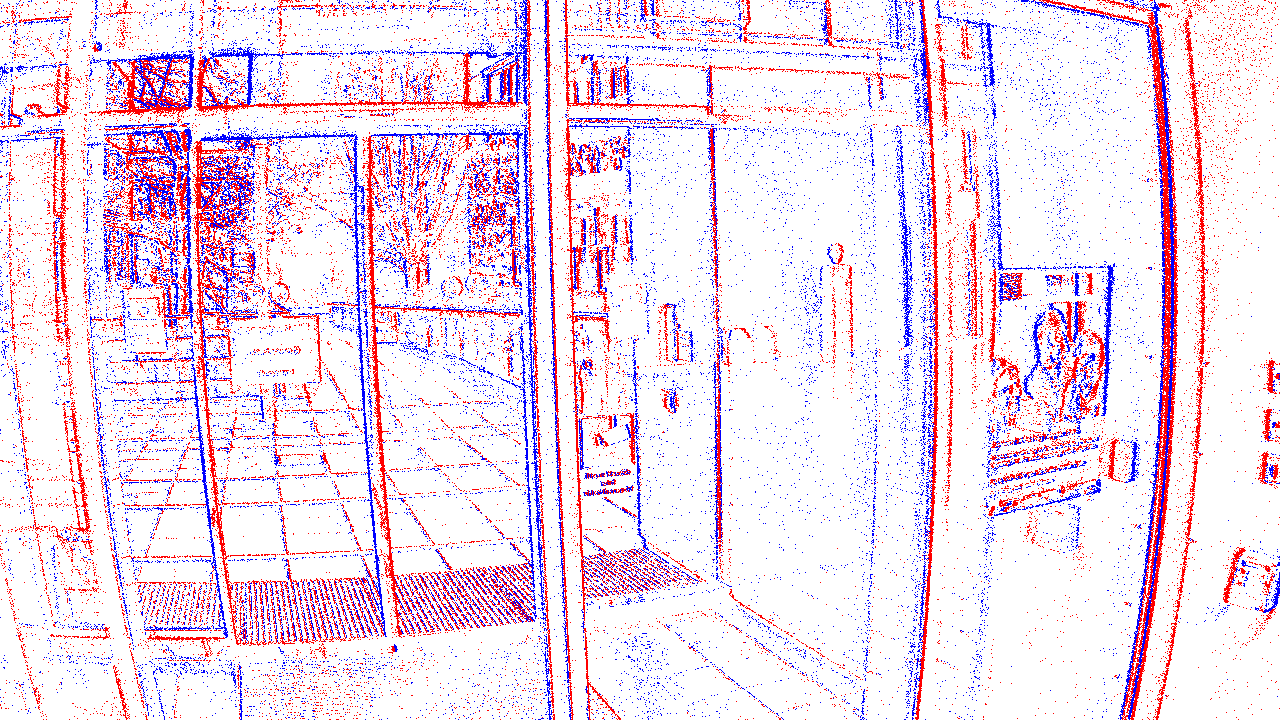}}
        &\gframe{\includegraphics[trim={0px 0 200px 0},clip,width=\linewidth]{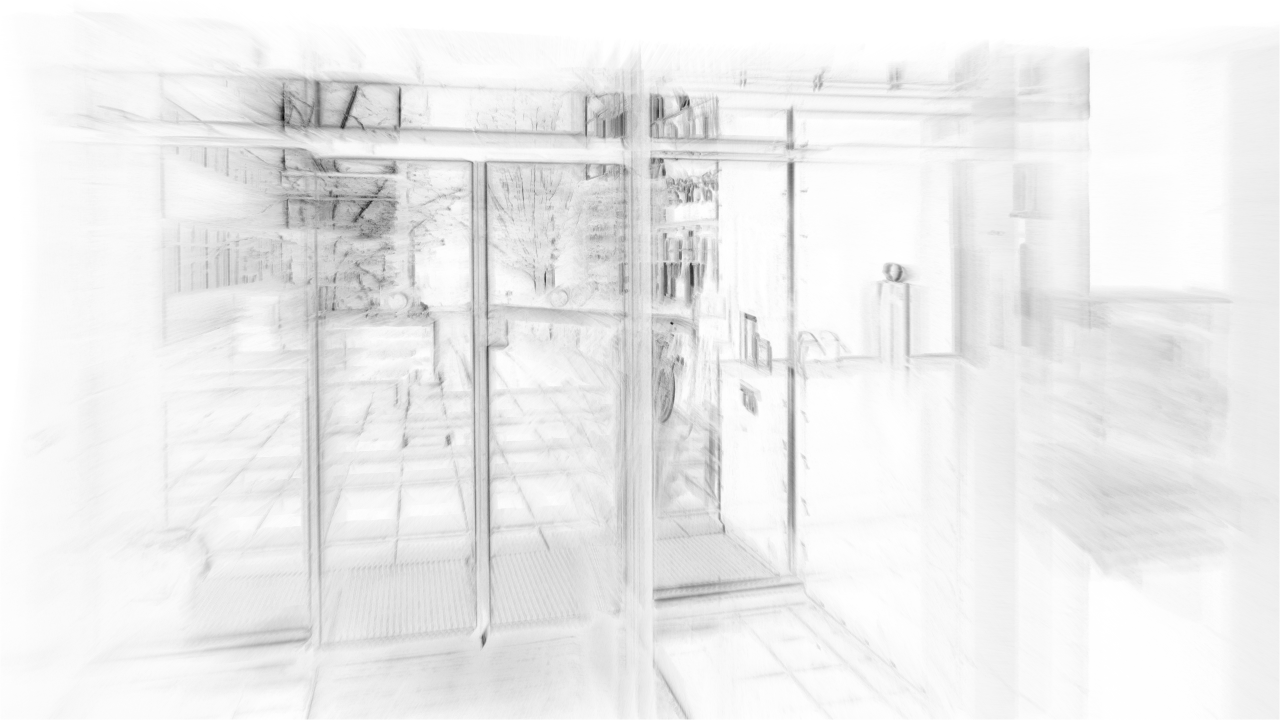}}
		&\gframe{\includegraphics[trim={0px 0 150px 0},clip,width=\linewidth]{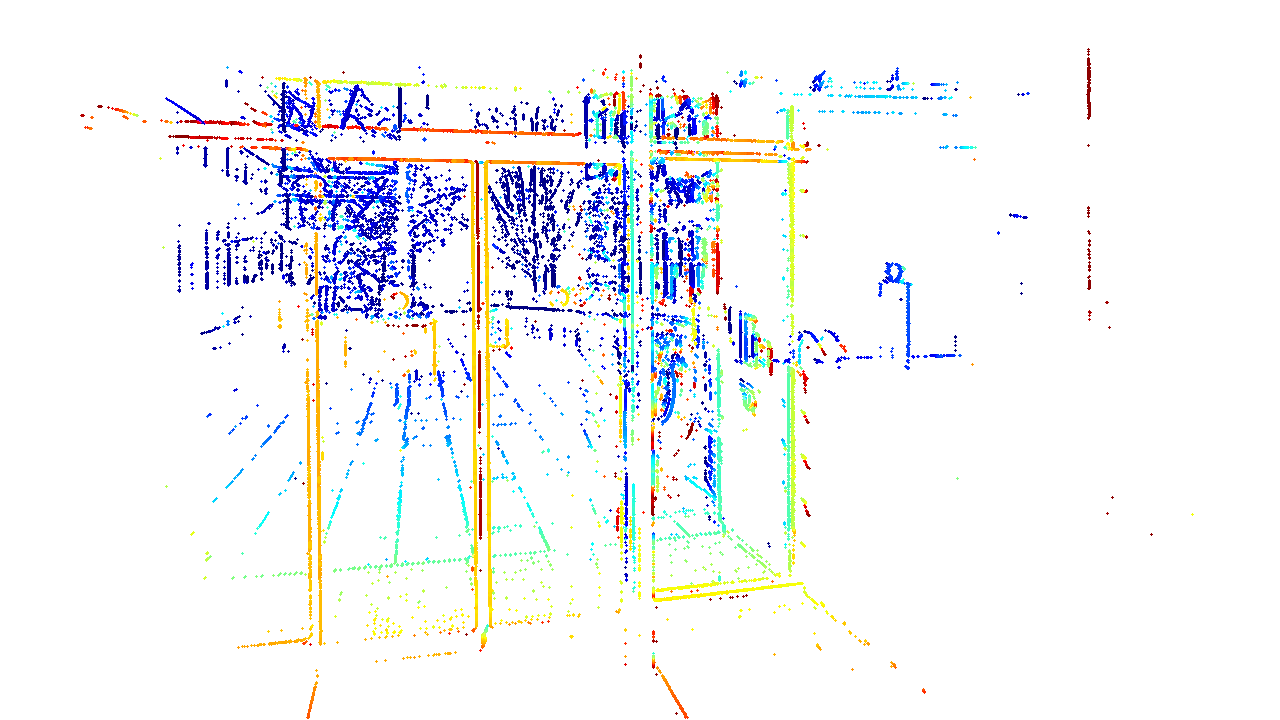}}
		\\
		
		\gframe{\includegraphics[trim={0px 150px 0 300px},clip,width=\linewidth]{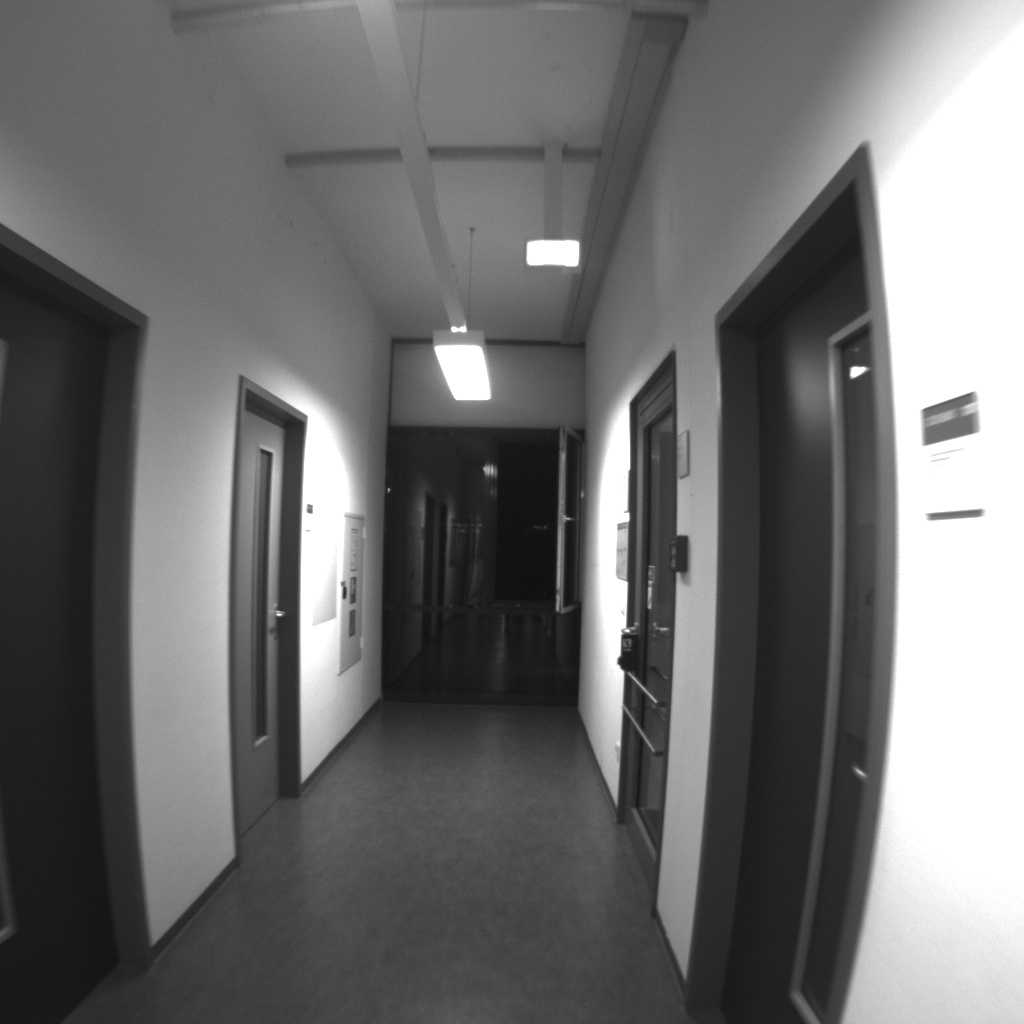}}
		&\gframe{\includegraphics[trim={0px 0 0 0},clip,width=\linewidth]{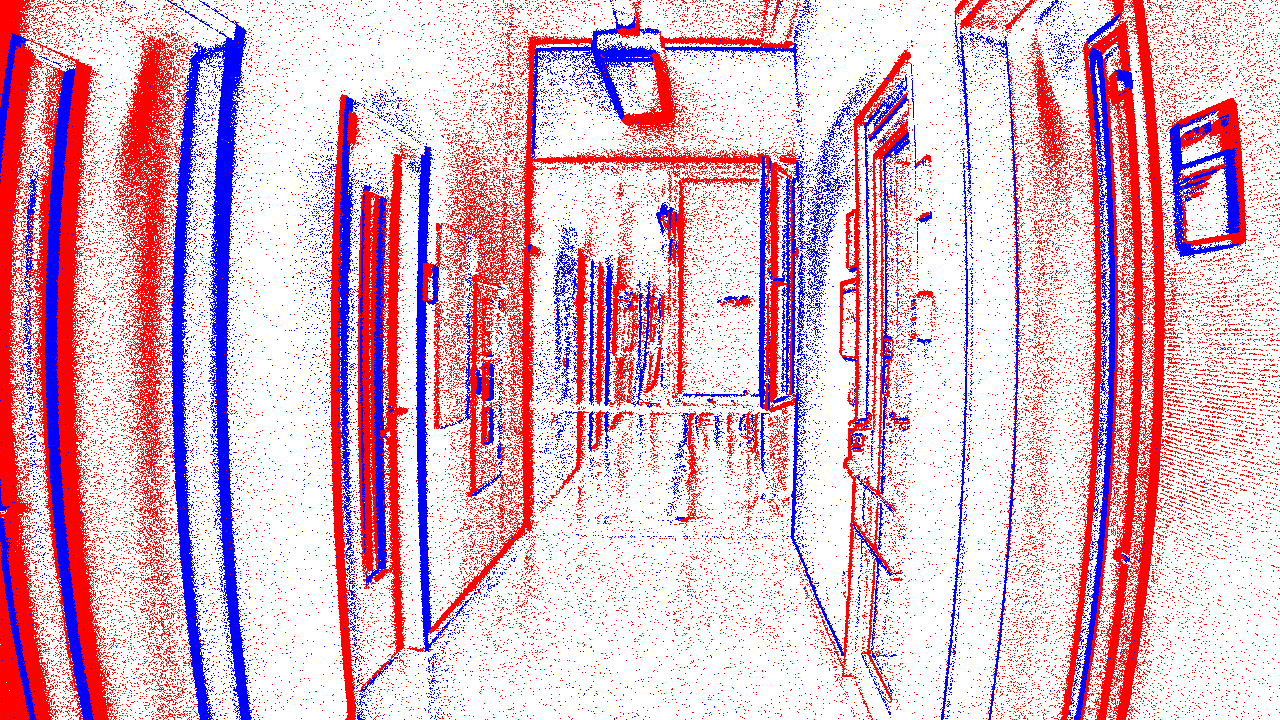}}
		&\gframe{\includegraphics[trim={0px 0 0 0},clip,width=\linewidth]{images/tumvie/skate-easy-1/15.000000confidence_map_fused_2.png}}
        &\gframe{\includegraphics[trim={0px 0 0 0},clip,width=\linewidth]{images/tumvie/skate-easy-1/15.000000inv_depth_colored_dilated_fused_2_w.png}}
		\\
	\end{tabular}
	}
	\caption{\label{fig:hdr}
	\emph{HDR scenes.} Output of Alg.~\ref{alg:fusion:stereo} on HDR scenes from the TUM-VIE dataset. 
 Unlike frame-based cameras, event cameras can perceive both under- and over-exposed regions of the scene well, leading to good depth estimation throughout.
	}
\end{figure*}

%% file: chapters/04_experim_multires.tex
\input{floats/fig_multires_stack}
\subsection{Effect of Varying the Sensor's Spatial Resolution}
\label{sec:experim:multires}

So far, results on datasets from three different resolutions have been presented: DAVIS346 (MVSEC), Prophesee Gen3 (DSEC) and Prophesee Gen4 (TUM-VIE).
However, the scenes are all different, some do not have ground truth depth, and calibration errors may influence the results.
To analyze the response of our stereo Alg.~\ref{alg:fusion:stereo} to varying pixel resolutions under controlled conditions (same scene, same FOV, etc.), we generate events using a simulator (ESIM \cite{Rebecq18corl}).
We use the textured scene \textit{flying\_room}, with a camera baseline of 20~\si{\centi\meter} and the OpenGL renderer. 
The scene is visualized in \cref{fig:multires} (bottom right).
Inspired by MVSEC, DSEC and TUM-VIE datasets, three sensor resolutions are tested: $320\times 240$~pix, $640\times 480$~pix and $1280\times 960$~pix, respectively.

We run Alg.~\ref{alg:fusion:stereo} on stereo events generated during the same time duration (increasing the sensor resolution increases the number of events generated within the same time duration: roughly 0.25 Mev, 2.25 Mev and 14.3 Mev for the three above resolutions, respectively).
\Cref{fig:multires} compares qualitatively the semi-dense depth maps and confidence maps recovered by our method on the three sensor resolutions inputs. 
The observations from these images are complemented by the quantitative analysis in \cref{fig:mutires:quantitative}.
\Cref{fig:multires:error} demonstrates quantitatively that depth errors decrease as the sensor resolution increases, for both monocular and stereo cases. 
Stereo fusion reduces the error almost by half with respect to the monocular case.
Next, we also analyze the density of the reconstructions.
\Cref{fig:mutires:quantitative} also reports the precision, recall and F1-score plots for the depth maps in \cref{fig:multires}. 
A clear trend is observed: the precision increases with increasing camera resolution (lower errors, like in \cref{fig:multires:error}). 
For the highest resolution tested ($1280\times 960$ pix), the precision reached $\approx 97~\si{\percent}$ within  4~\si{\centi\meter} of depth error tolerance. 
However, the recall curves state that a lower resolution may allow us to recover more 3D points in the scene (\SI{6}{\percent} vs. \SI{2}{\percent}), at the expense of bigger depth errors. 
Since the depth outputs are highly sparse due to the nature of event data, the F1-score is heavily skewed by the low values of recall.

%% file: floats/fig_multires_stack.tex
\def\figWidth{0.24\linewidth}
\begin{figure*}[t]
	\centering
    {\small
    \setlength{\tabcolsep}{1pt}
	\begin{tabular}{
	>{\centering\arraybackslash}m{0.4cm} 
	>{\centering\arraybackslash}m{\figWidth} 
	>{\centering\arraybackslash}m{\figWidth}
	>{\centering\arraybackslash}m{\figWidth}
	>{\centering\arraybackslash}m{\figWidth}}
		& $320 \times 240$
		& $640 \times 480$
		& $1280 \times 960$
		& Ground truth
		\\\addlinespace[.3ex]

		\rotatebox{90}{\makecell{Depth map}}
		&\gframe{\includegraphics[width=\linewidth]{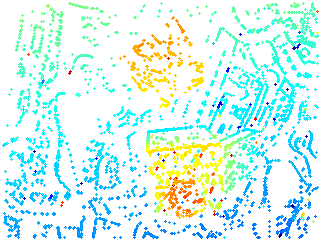}}
		&\gframe{\includegraphics[width=\linewidth]{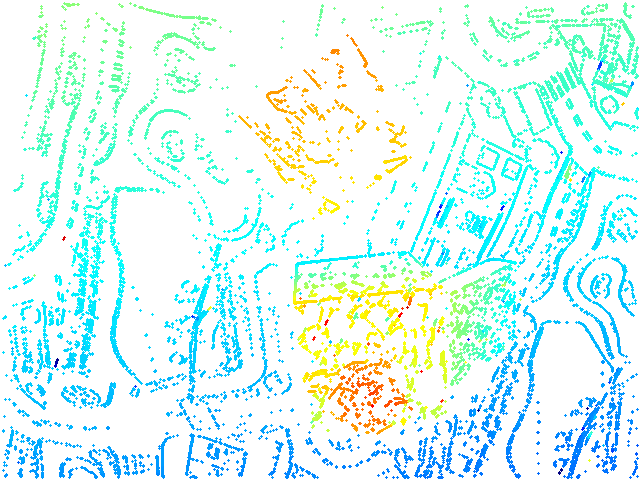}}
		&\gframe{\includegraphics[width=\linewidth]{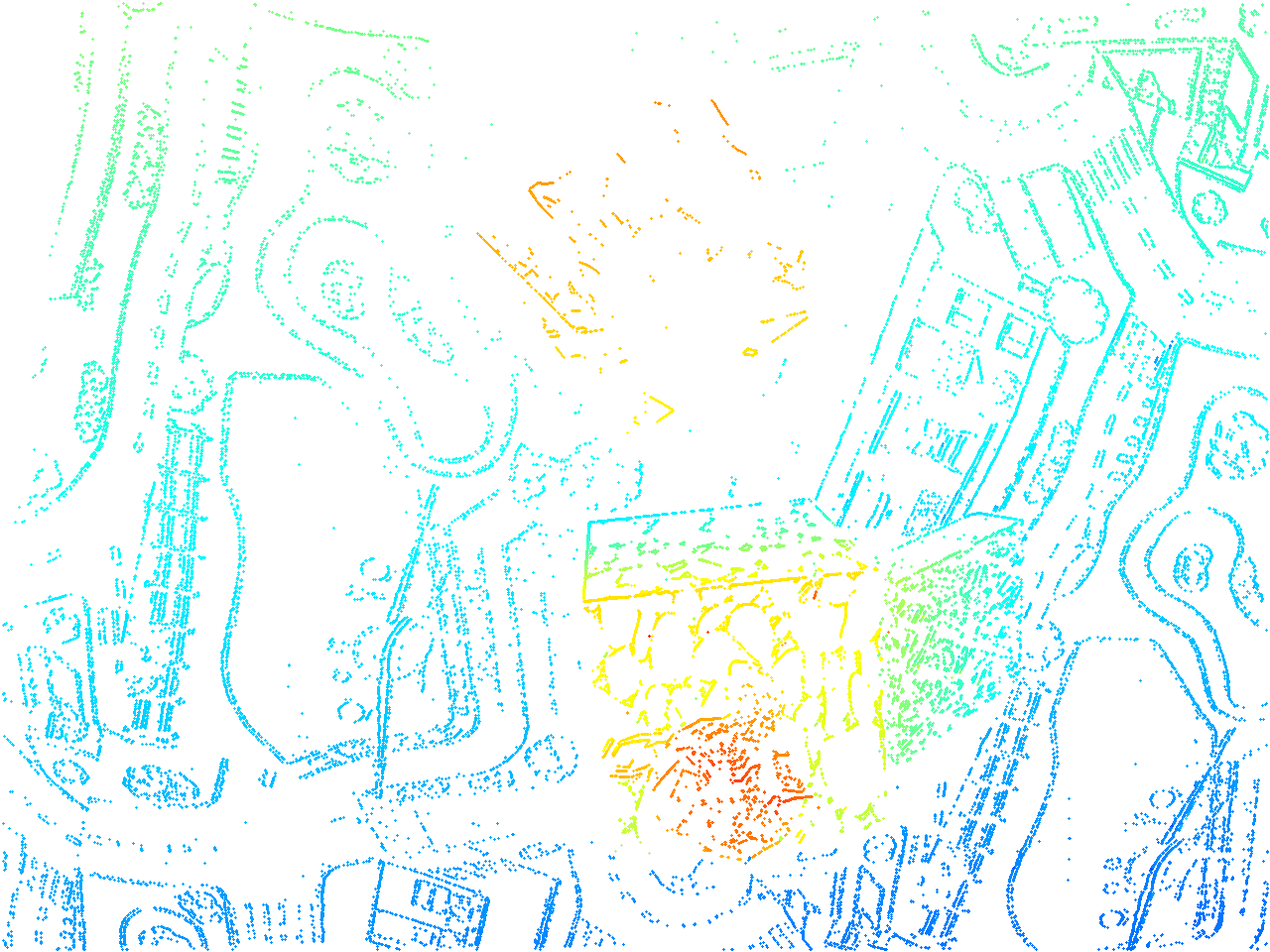}}
		&\gframe{\includegraphics[width=\linewidth]{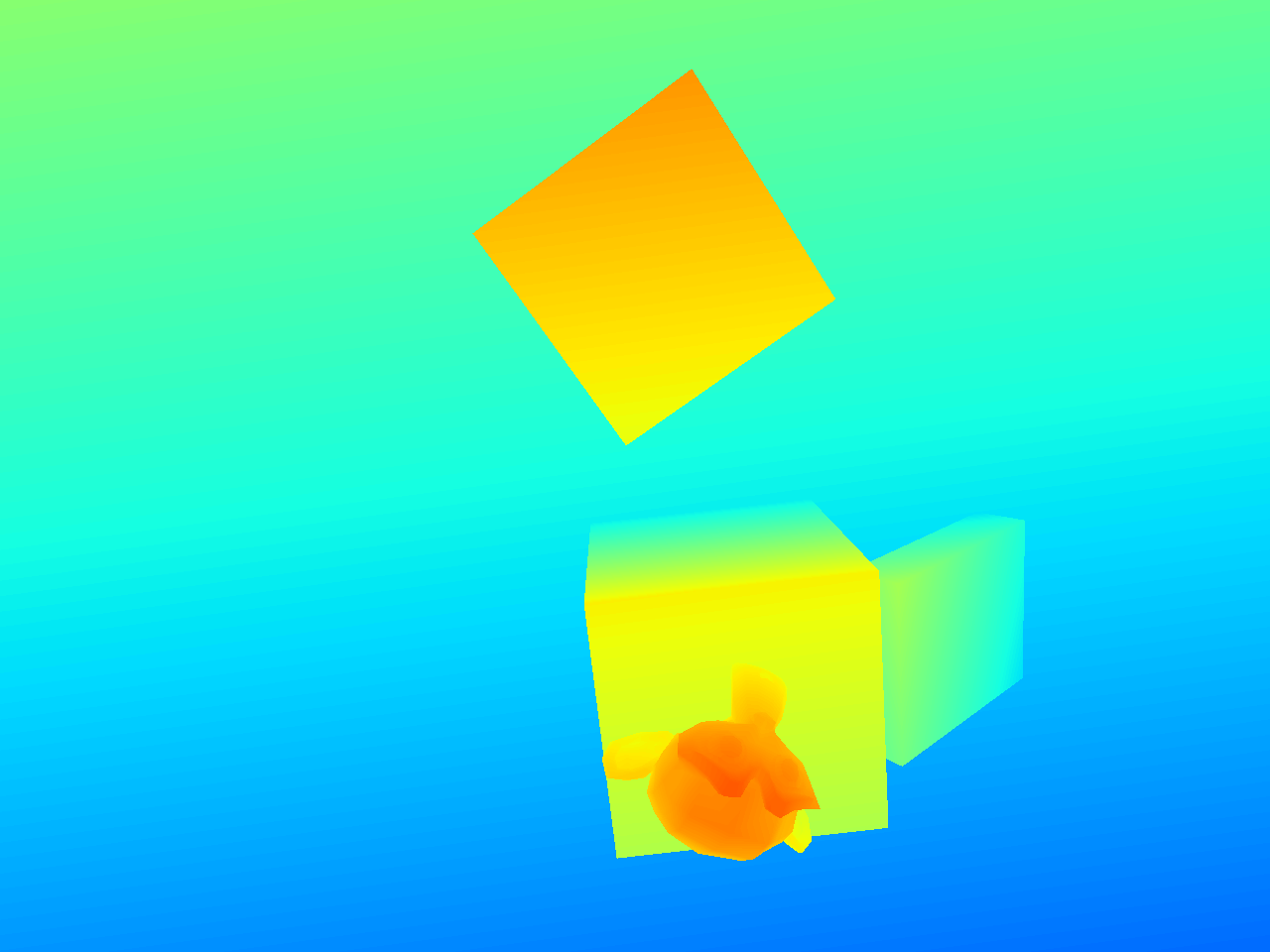}}
		\\
		
		\rotatebox{90}{\makecell{Confidence map}}
		&\gframe{\includegraphics[width=\linewidth]{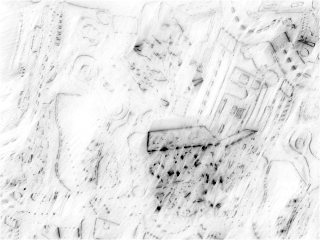}}
		&\gframe{\includegraphics[width=\linewidth]{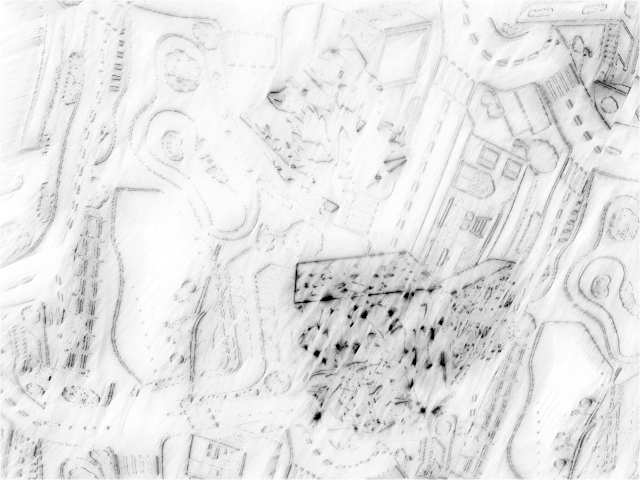}}
		&\gframe{\includegraphics[width=\linewidth]{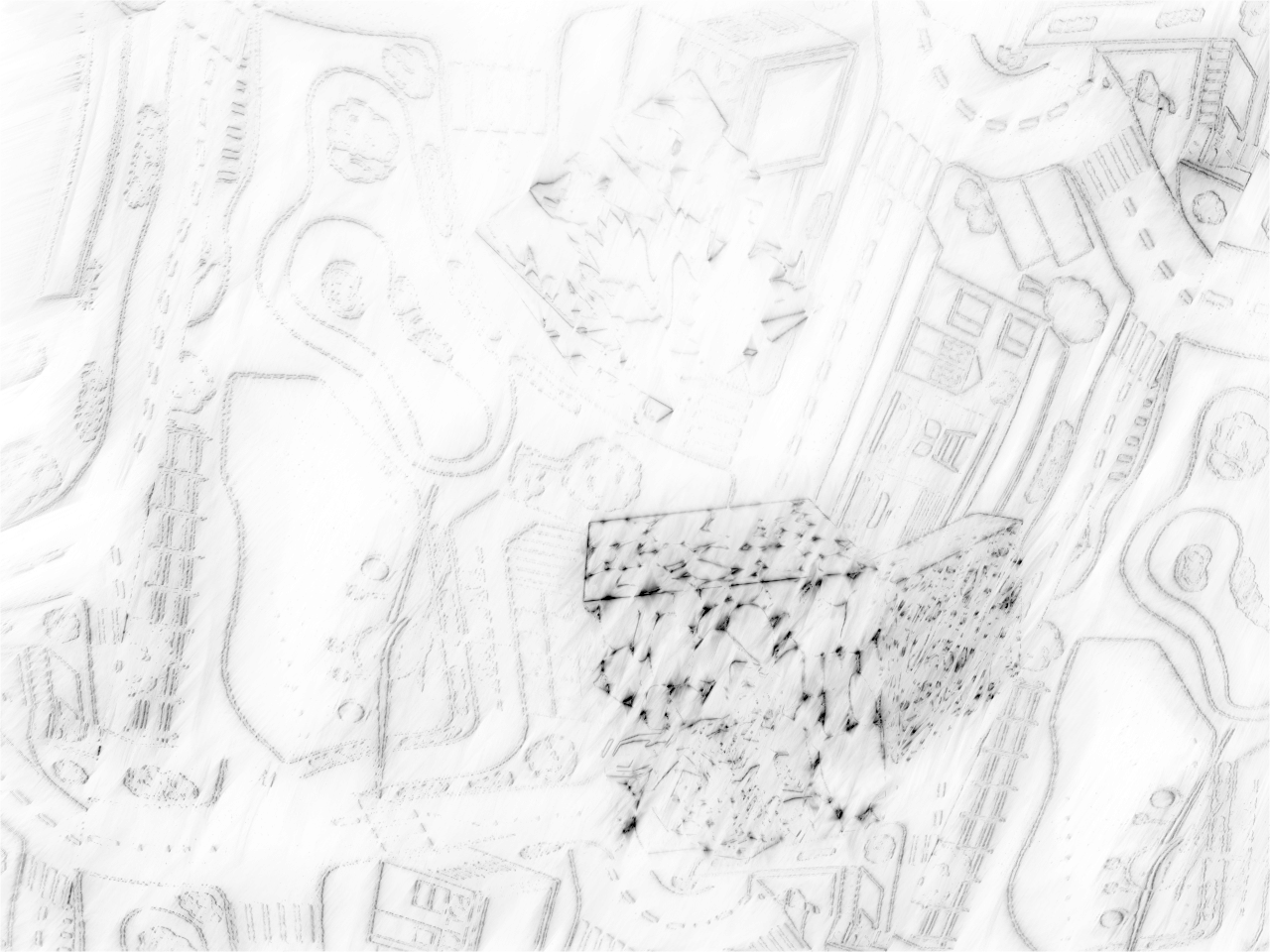}}
	    &\gframe{\includegraphics[trim={58px 0 58px 0},clip,width=\linewidth]{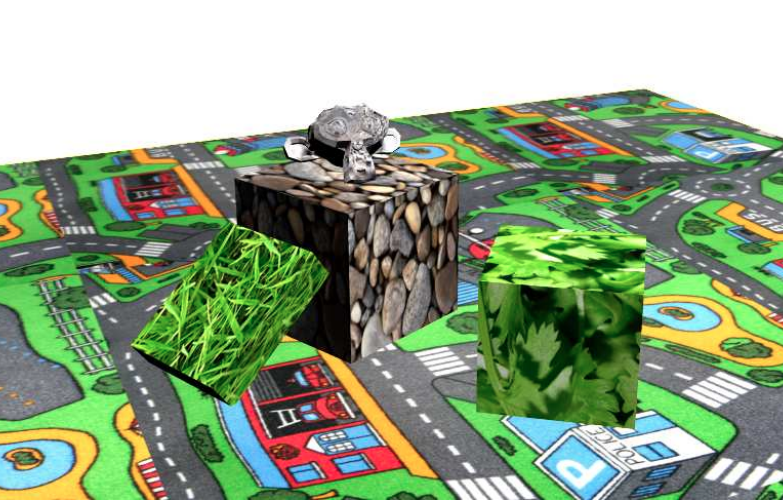}}
		\\
	\end{tabular}
	}
	\caption{\label{fig:multires} \emph{Effect of the sensor's spatial resolution}. 
	Output of Alg.~\ref{alg:fusion:stereo} on events simulated with different camera resolutions.
	The scene --\emph{flying\_room}-- is shown on the top right. 
	Depth maps are colored from red (near) to blue (far) in the range 1.3--4~\si{\meter}.}
\end{figure*}
\begin{figure*}[t]
\def\imgWidth{0.245\linewidth}
\begin{subfigure}{.265\linewidth}
    \includegraphics[trim={0mm 0mm 9mm 10mm},clip,width=\linewidth]{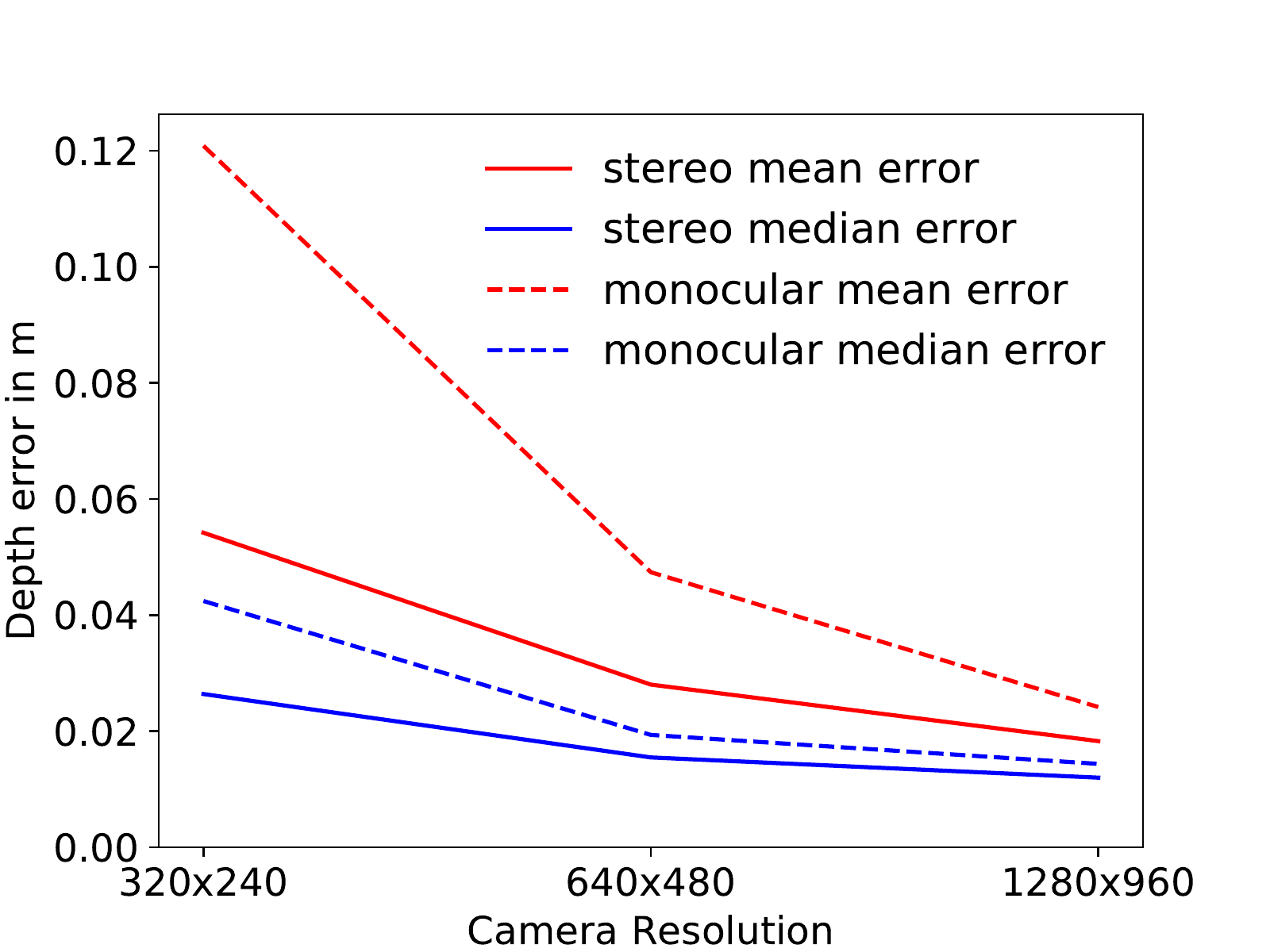}%
    \caption{Depth error vs. resolution\label{fig:multires:error}}
\end{subfigure}%
\begin{subfigure}{\imgWidth}
    \includegraphics[trim={6mm 0 13mm 7mm},clip,width=\linewidth]{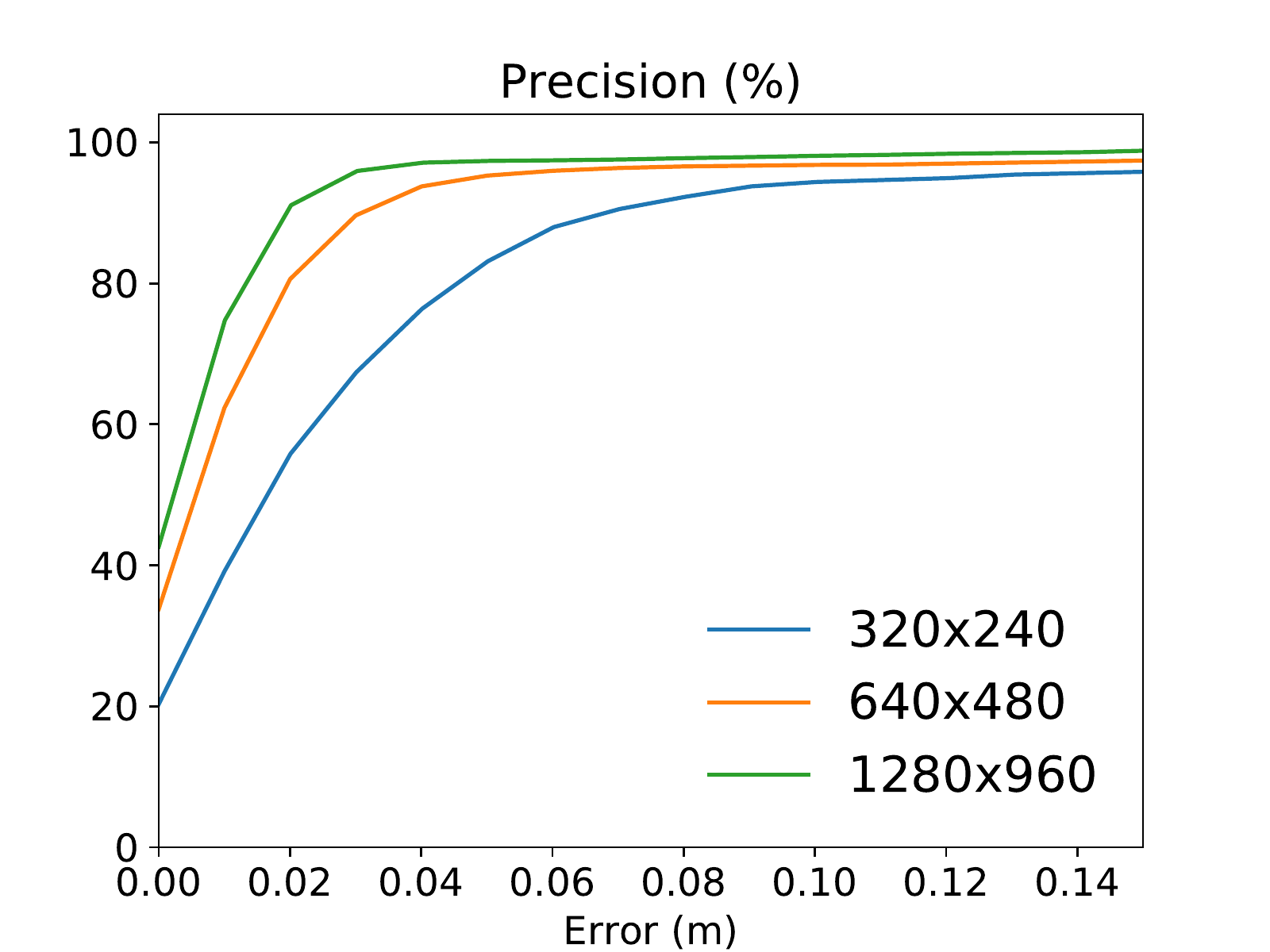}%
    \caption{Precision (\%)\label{fig:multires:precision}}
\end{subfigure}%
\begin{subfigure}{\imgWidth}
    \includegraphics[trim={6mm 0 13mm 7mm},clip,width=\linewidth]{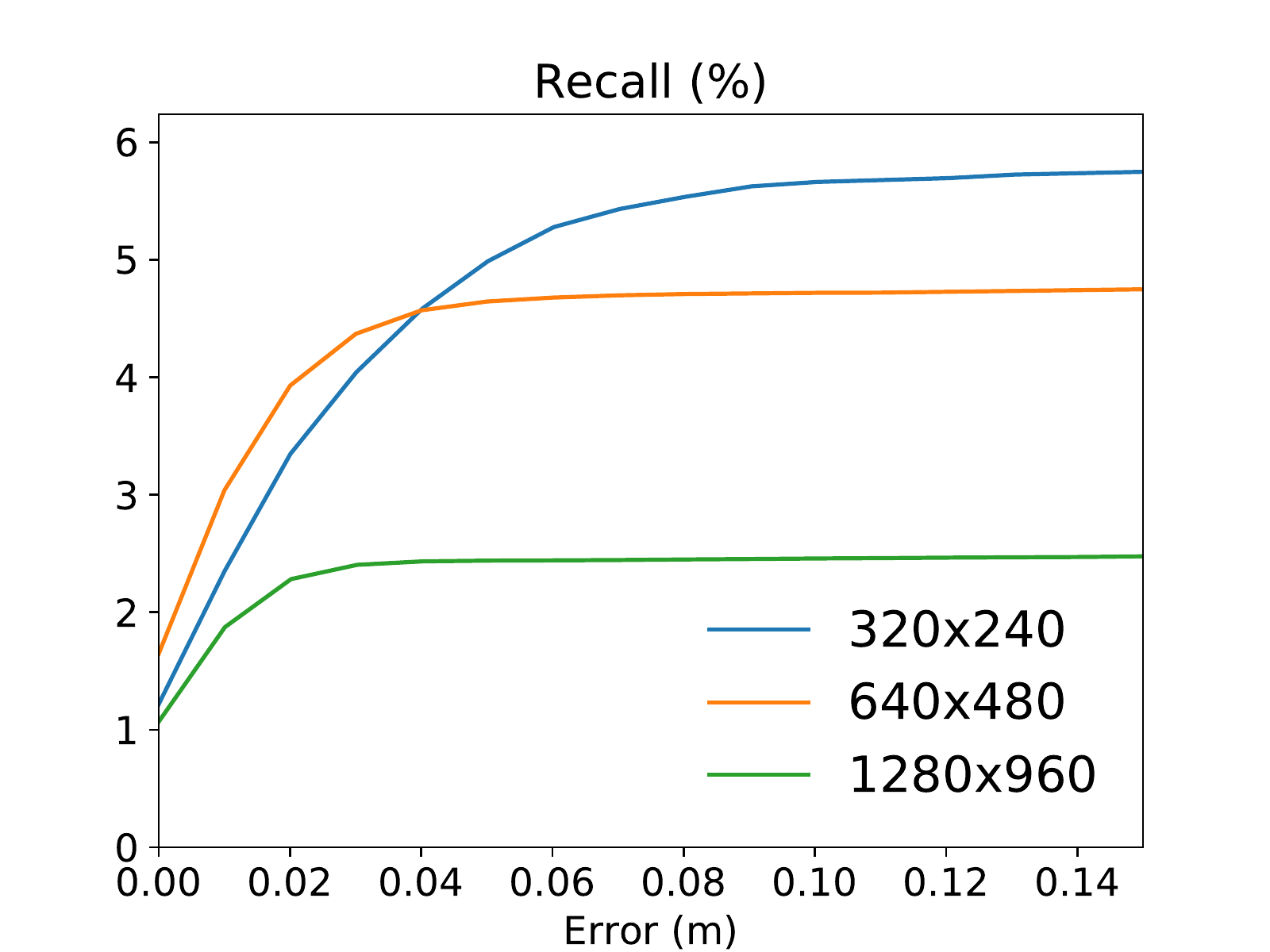}%
    \caption{Recall (\%)\label{fig:multires:recall}}
\end{subfigure}%
\begin{subfigure}{\imgWidth}
    \includegraphics[trim={6mm 0 13mm 7mm},clip,width=\linewidth]{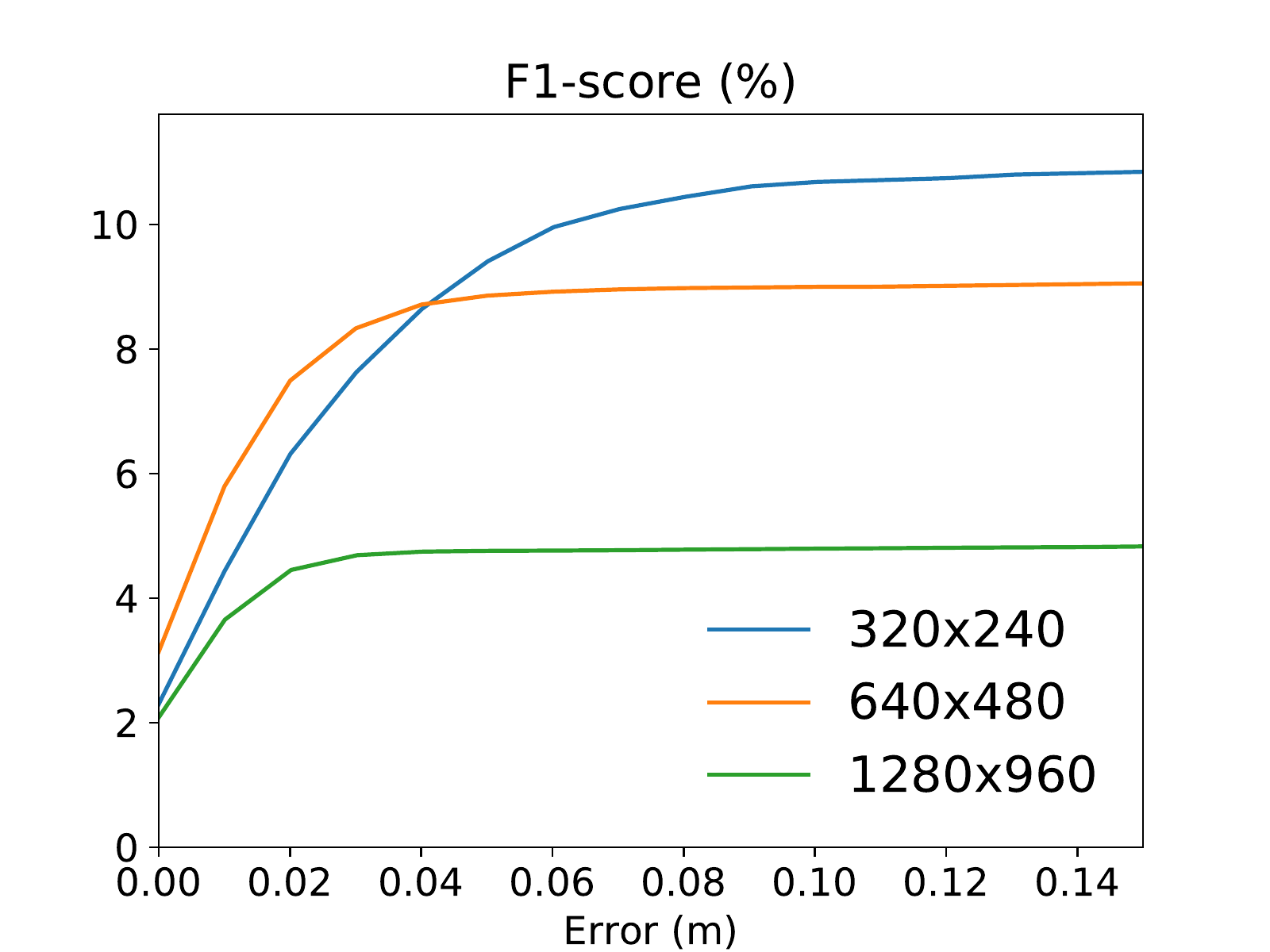}%
    \caption{F1-score (\%)\label{fig:multires:fscore}}
\end{subfigure}
    \caption{\label{fig:mutires:quantitative}
    (a) Depth errors across various camera resolutions for monocular \cite{Rebecq18ijcv} and stereo Alg.~\ref{alg:fusion:stereo} methods. 
    Maximum scene depth is \SI{2.7}{\meter}.
    (b)-(d) Precision, recall and F1-score for depth estimated using Alg.~\ref{alg:fusion:stereo} on scenes of different resolutions.
    }
\end{figure*}

%% file: chapters/04_experim_analysis.tex
\subsection{Fusing More than Two Event Cameras}
\label{sec:experim:morethantwocams}
\input{floats/fig_evimo2_combined}
We also test our method on sequences from the EVIMO2 dataset~\cite{Burner22evimo2}, recorded with a \emph{trinocular} event-camera rig consisting of a Samsung DVS Gen3~\cite{Son17isscc} and two Prophesee CD Gen3 event cameras~\cite{propheseeevk}, all with $640\times 480$ pix. 
The field of views (FOVs) of the cameras have a narrow overlap due to the way they are arranged in the sensor rig (Prophesee event cameras are in portrait mode, whereas the Samsung DVS, in the middle, is in landscape mode).
We set the central camera (Samsung DVS) as the reference one.

\Cref{fig:evimo2} shows the depth- and confidence maps from each camera separately and for the fused DSI during two motions (two pairs of columns): a normal one and a fast motion, with retinal speeds between 1900 pix/s for objects in the far end and 3500 pix/s for objects close to the camera.
The resulting semi-dense depth map obtained from the fused DSI inherit the above-mentioned narrow FOV overlap of the cameras. 
Overall, the fused depth map suppresses noise that would otherwise appear as very prominent outliers in the individual depth maps.
This experiment shows that \emph{our method naturally fuses multiple cameras} with linear complexity, i.e., without handling them by pairs, as prior works do.

\subsection{Runtime}
\label{sec:experim:runtime}

\input{floats/tab_runtime}

Complementing the complexity analysis in \cref{sec:complexity}, \cref{tab:runtime} presents the average time taken in each step (DSI creation, DSI fusion, argmax and AGT thresholding) over 100 sample runs, on a laptop with an Intel i7-10510U 8-core CPU. 
We consider typical numbers from a stereo DAVIS346 configuration \cite{Taverni18tcsii}.
For comparison, we also report the numbers obtained with only one of the cameras (monocular setup \cite{Rebecq18ijcv}).
DSI creation takes the longest time as it is a complex transformation and depends on the number of input events.
We did not find major runtime differences in our implementation of the fusion functions tested \eqref{eq:fusionfunc:A}--\eqref{eq:fusionfunc:max}.

The numbers in \cref{tab:runtime} agree with complexity formulas \eqref{eq:complexityEMVS}--\eqref{eq:complexityAlgTwo}.
The DSI creation runtime of the stereo methods is roughly twice ($N_c=2$) that of the monocular method (520 vs. 234 ms). 
The temporal fusion step with $N_s=2$ sub-intervals is twice as expensive as that with one interval (98 vs. 51 ms).
The arg max and AGT steps are the same for all methods, thus no major runtime differences are observed.

\input{chapters/04_sensitivity_contrast_threshold}

%% file: floats/fig_evimo2_combined.tex
\def\figWidth{0.22\linewidth}
\begin{figure}[t]
	\centering

\begin{subfigure}{0.6\linewidth}
    {\includegraphics[trim={110px 0 110px 60px},clip,width=\linewidth]{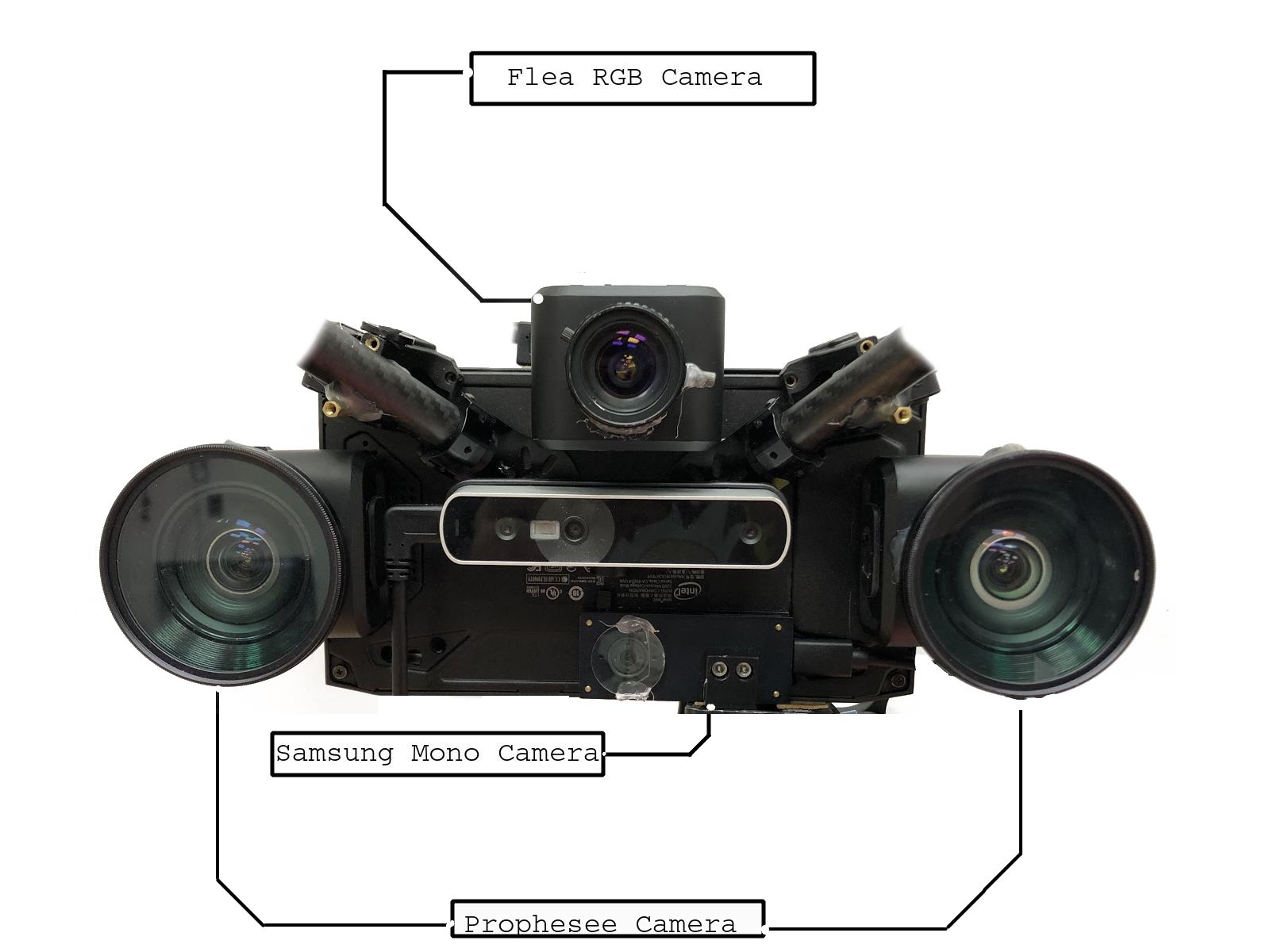}}
    \caption{\label{fig:evimorig}Trinocular rig from EVIMO2v1.
    Used with permission \cite{Burner22evimo2}, 2019.}
\end{subfigure}
\vspace{3ex}
	
    {\small
    \setlength{\tabcolsep}{2pt}
	\begin{tabular}{
	>{\centering\arraybackslash}m{0.3cm} 
	>{\centering\arraybackslash}m{\figWidth}
	>{\centering\arraybackslash}m{\figWidth} 
	>{\centering\arraybackslash}m{\figWidth} 
	>{\centering\arraybackslash}m{\figWidth}}
        & RGB Image (Flea3) & Events (Samsung)
        & RGB Image (Flea3) & Events (Samsung)
        \\\addlinespace[0.2ex]
        &\gframe{\includegraphics[width=\linewidth]{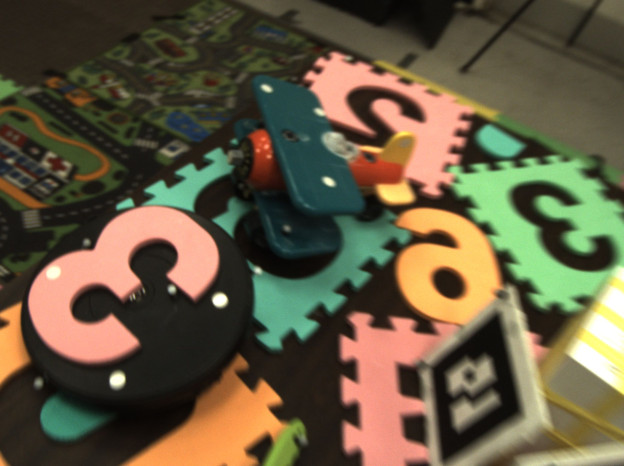}}	
        &\gframe{\includegraphics[width=\linewidth]{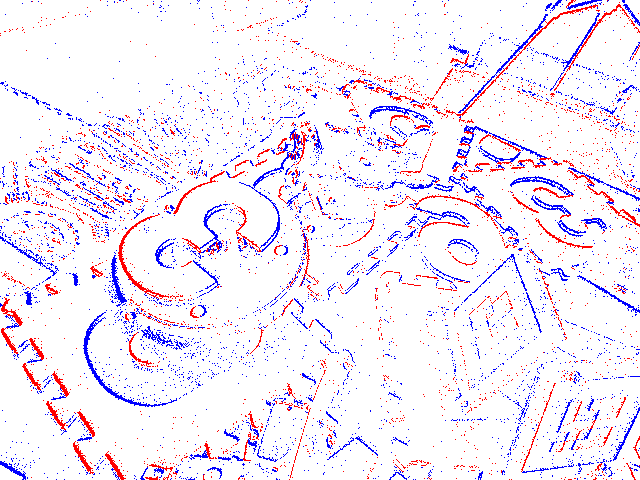}}
        &\gframe{\includegraphics[width=\linewidth]{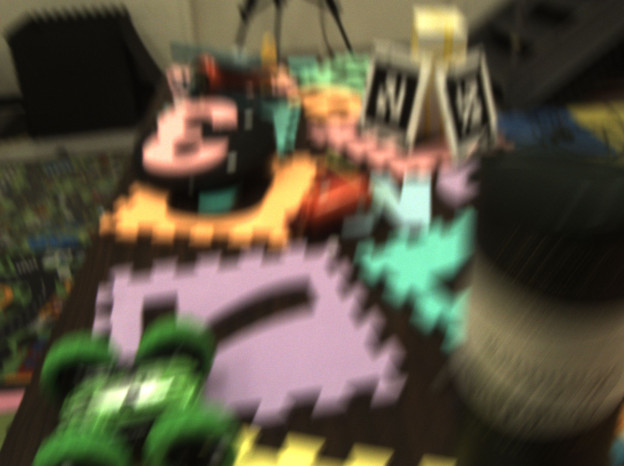}}	
        &\gframe{\includegraphics[width=\linewidth]{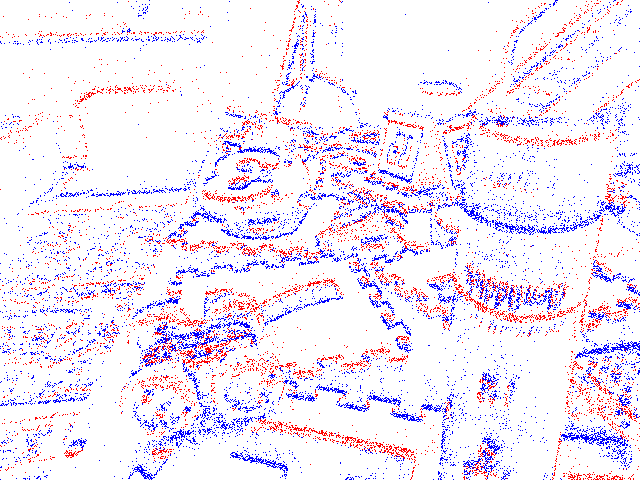}}
        \\\addlinespace[2ex]
        
        & Conf. Map & Inv. Depth
        & Conf. Map & Inv. Depth\\\addlinespace[0.2ex]        
        \rotatebox{90}{\makecell{Prophesee L}}		
		&\gframe{\includegraphics[trim={6cm 3cm 5cm 3cm},clip,width=\linewidth]{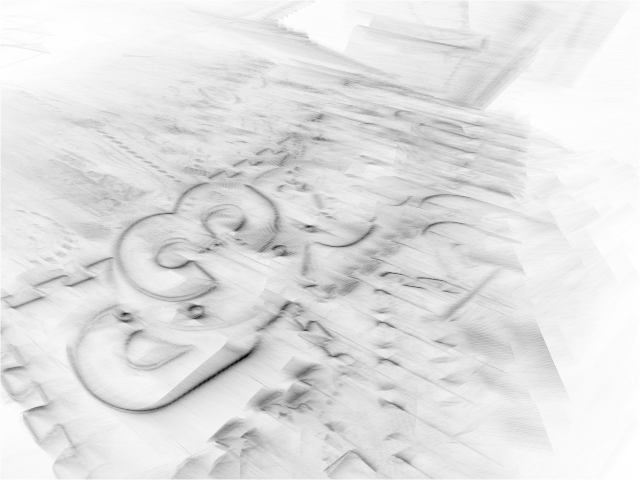}}
		&\gframe{\includegraphics[trim={6cm 3cm 5cm 3cm},clip,width=\linewidth]{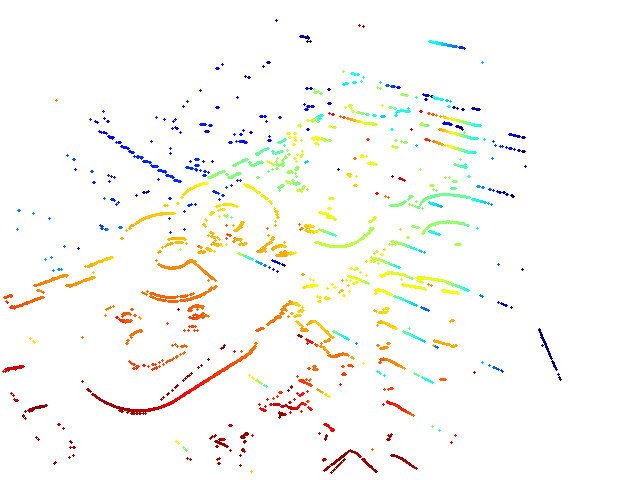}}	
		&\gframe{\includegraphics[trim={6cm 3cm 5cm 3cm},clip,width=\linewidth]{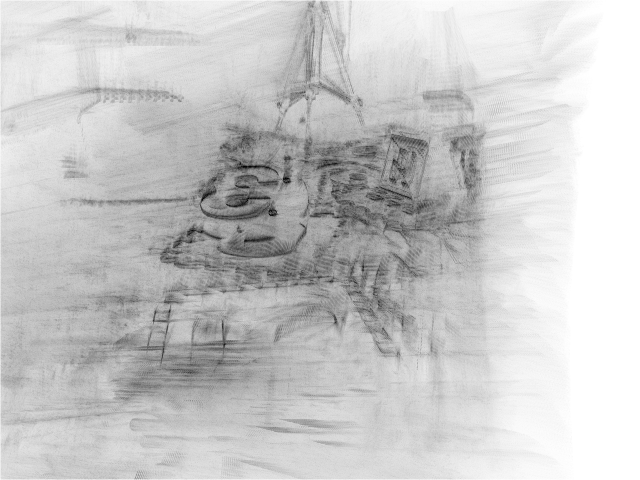}}
		&\gframe{\includegraphics[trim={6cm 3cm 5cm 3cm},clip,width=\linewidth]{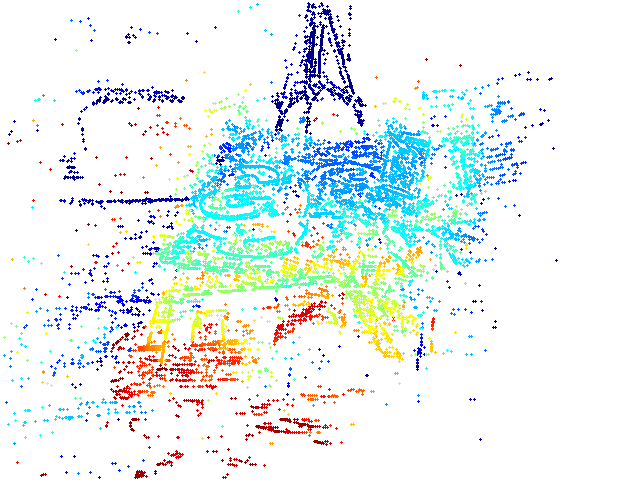}}	
		\\
		
		\rotatebox{90}{\makecell{Samsung}}
        &\gframe{\includegraphics[trim={6cm 3cm 5cm 3cm},clip,width=\linewidth]{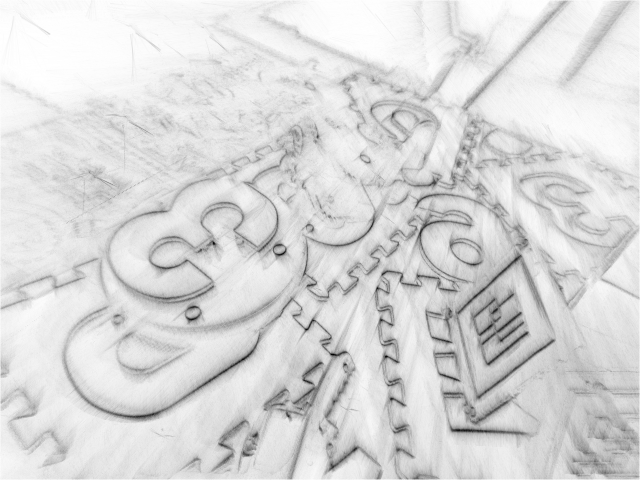}}		
        &\gframe{\includegraphics[trim={6cm 3cm 5cm 3cm},clip,width=\linewidth]{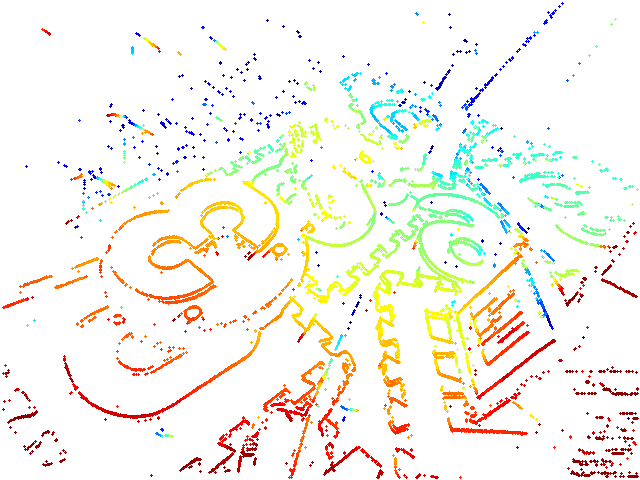}}
        &\gframe{\includegraphics[trim={6cm 3cm 5cm 3cm},clip,width=\linewidth]{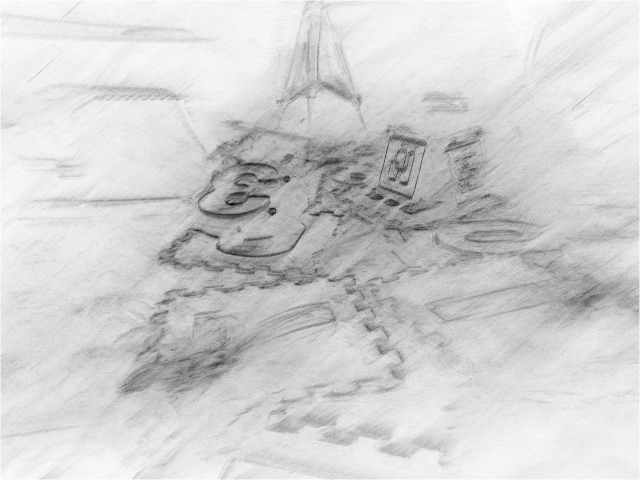}}		
        &\gframe{\includegraphics[trim={6cm 3cm 5cm 3cm},clip,width=\linewidth]{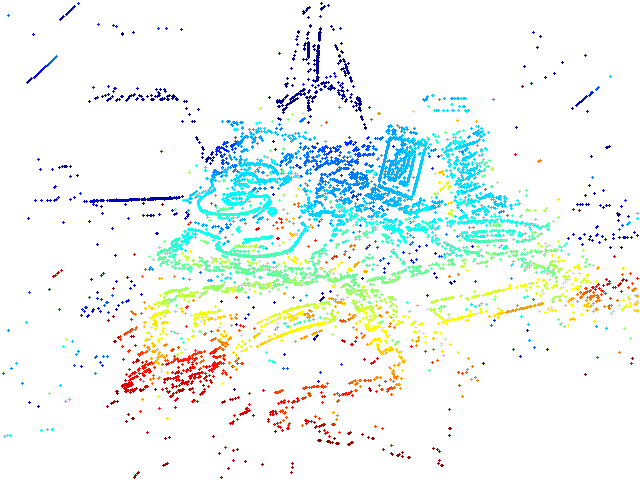}}
        \\

		\rotatebox{90}{\makecell{Prophesee R}}	
		&\gframe{\includegraphics[trim={6cm 3cm 5cm 3cm},clip,width=\linewidth]{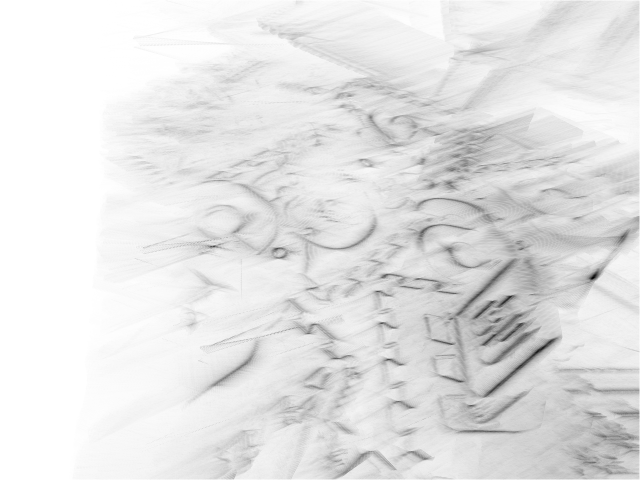}}
		&\gframe{\includegraphics[trim={6cm 3cm 5cm 3cm},clip,width=\linewidth]{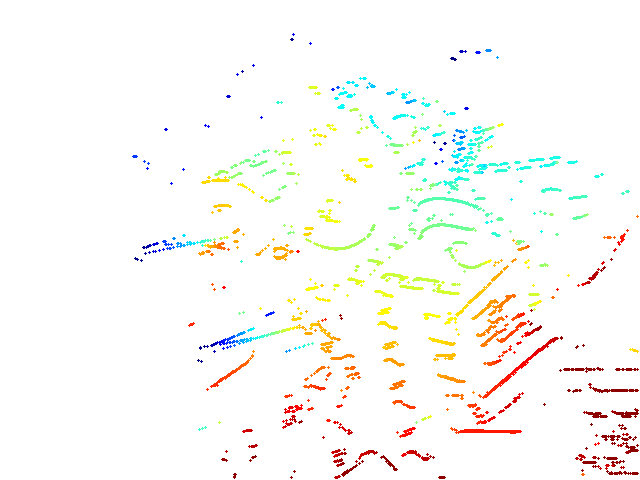}}	
		&\gframe{\includegraphics[trim={6cm 3cm 5cm 3cm},clip,width=\linewidth]{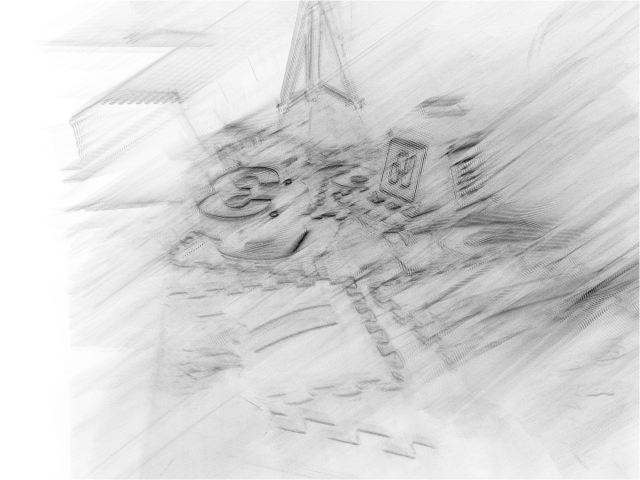}}
		&\gframe{\includegraphics[trim={6cm 3cm 5cm 3cm},clip,width=\linewidth]{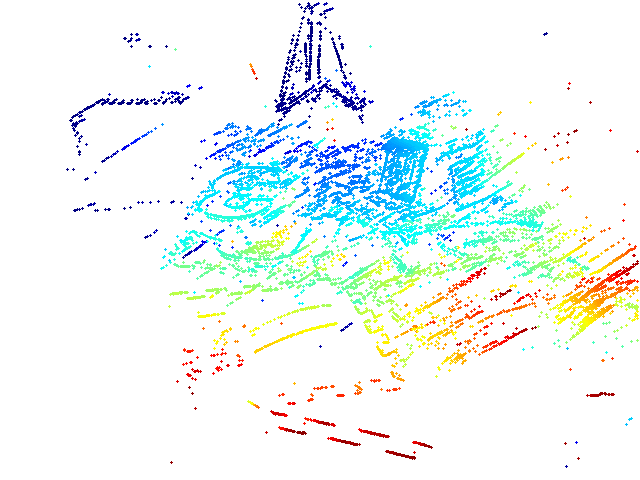}}	
		\\
		
		\rotatebox{90}{\makecell{\textbf{Fused}}}	
		&\gframe{\includegraphics[trim={6cm 3cm 5cm 3cm},clip,width=\linewidth]{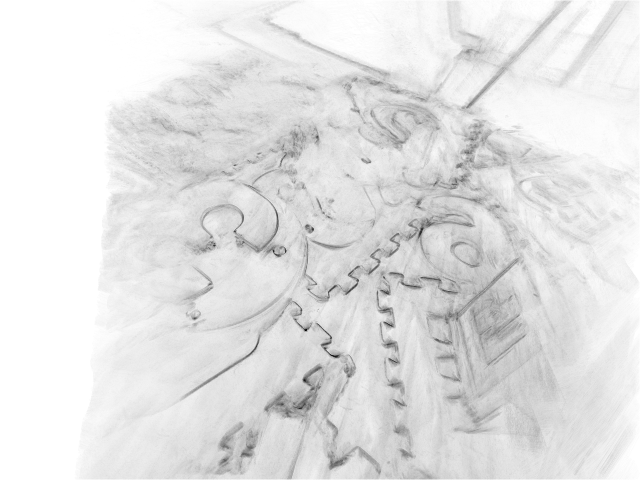}}
		&\gframe{\includegraphics[trim={6cm 3cm 5cm 3cm},clip,width=\linewidth]{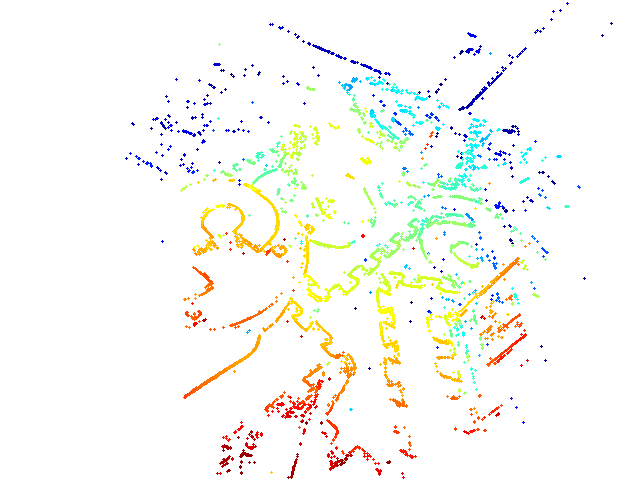}}	
		&\gframe{\includegraphics[trim={6cm 3cm 5cm 3cm},clip,width=\linewidth]{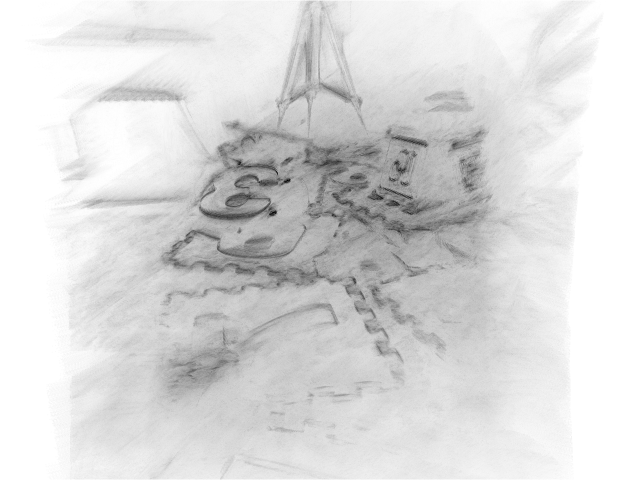}}
		&\gframe{\includegraphics[trim={6cm 3cm 5cm 3cm},clip,width=\linewidth]{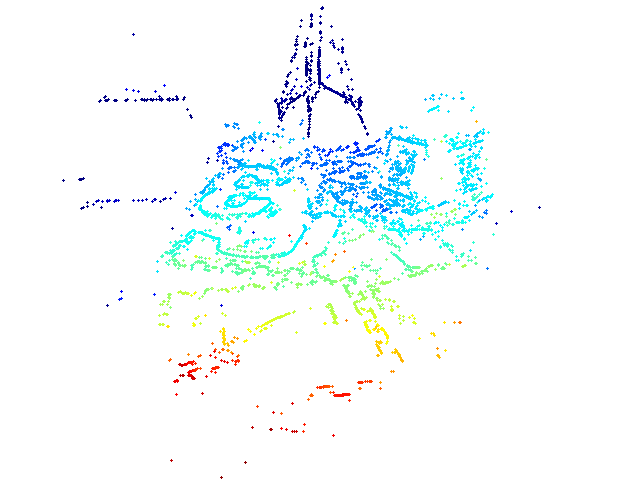}}	
		\\
        
	\end{tabular}
	}
	\caption{\label{fig:evimo2}
	\emph{Trinocular event-camera fusion.} 
	Output of Stereo Alg.~\ref{alg:fusion:stereo} on the SFM sequence 03\_00 from the EVIMO2 dataset \cite{Burner22evimo2}. %
	The first row shows an RGB image of the scene 
	and the events (from the Samsung DVS~\cite{Son17isscc}) over a short time duration, 
	displayed in red/blue according to polarity.
	The following rows show the output of our stereo method using the three event cameras in the dataset.
	}
\end{figure}

%% file: floats/tab_runtime.tex
\begin{table}[t]
\centering
\caption{Runtime comparison of the different steps of three algorithms.
Parameters: $N_{e}\approx 235$ kev, 
$N_{c}=2$, $N_{p}\approx 90$k (DAVIS346), 
$N_{Z}=100$, $N_{k}=25$ ($5\times 5$), $N_{s}=2$ for Alg.~\ref{alg:fusion:time:stereo} (here, $A_t\circ H_c$). 
Time is given in \si{\milli\second}.
}
\label{tab:runtime}
\begin{tabular}{lccc}
\toprule 
 & EMVS~\cite{Rebecq18ijcv} & Alg.~\ref{alg:fusion:stereo} & Alg.~\ref{alg:fusion:time:stereo}\\
\midrule
DSI creation & 233.96 & \multicolumn{2}{c}{520.01}\\
DSI fusion & -- & 51.09 & 98.01\\
arg max & 36.34 & \multicolumn{2}{c}{35.62}\\
AGT & 0.28 & \multicolumn{2}{c}{0.28}\\
\bottomrule
\end{tabular}
\end{table}

%% file: chapters/04_sensitivity_contrast_threshold.tex
\input{floats/fig_contrast_threshold_stack}
\subsection{Effect of Varying the Contrast Sensitivity}
\label{sec:experim:varyingcontrastthreshold}

Due to the sparse nature of event data, which is largely controlled in the camera by the contrast sensitivity threshold, it is interesting to analyze the performance of our 3D reconstruction method for various sparsity levels.
To this end, we ran Alg.~\ref{alg:fusion:stereo} on stereo events simulated using ESIM \cite{Rebecq18corl} with five levels of event generation contrast threshold $\theta=\{0.05, 0.1, 0.2, 0.4, 0.8\}$. 
A small contrast threshold implies that a small change in brightness is sufficient to trigger events, and thus leads to the generation of many events from edges and textures in the scene (i.e., high sensor sensitivity). 
\Cref{fig:contrast_thresh} illustrates the 3D reconstruction quality across varying contrast thresholds for the flying room sequence. 
In general, we observe stable reconstructions as the contrast threshold increases, except for the fact that fewer points are recovered (e.g., on the floor) and more noisy points (i.e., outliers) appear.
This is also observed quantitatively in \cref{fig:contrast:error}, where the mean error increases for high contrast thresholds (the larger number of outliers distort the mean error) while the median error remains fairly constant (a sign of ``stability'' if outliers are removed).
Figures~\ref{fig:contrast:precision}--\ref{fig:contrast:fscore} depict the precision, recall and F1-score curves for depth predicted using data from different contrast thresholds. 
We observe that low contrast thresholds provide better precision and recall since they have fewer noisy outliers and recover more 3D points respectively. 
The increase in precision as well as recall comes at the cost of increased computational overhead as more events need to be processed at lower contrast thresholds.

In summary, the synthetic experiments suggest that increasing the event count either by decreasing the contrast threshold or increasing the spatial resolution of the camera improves 3D reconstruction accuracy. 
This comes at the cost of increased computational effort needed to process more input data. 

While event cameras are dominated by those that compute temporal contrast \cite{Lichtsteiner08ssc} (DVS), an ``event'' could have a broader interpretation, such as any meaningful information that decreases demands on bandwidth, memory, and power for data transmission, storage and processing.
This work has analyzed the advantages that event cameras offer for stereo depth estimation. 
Varying more than the temporal contrast is possible with different prototype vision sensors, such as Parallel Processor Arrays (PPAs). 
These sensors embed a processor within each pixel and are thus programmable, enabling a richer family of operations (at the expense of larger pixel sizes or more transistor layers) \cite{Dudek22sciro}.

%% file: floats/fig_contrast_threshold_stack.tex
\def\figWidth{0.19\linewidth}
\begin{figure*}[t]
	\centering
    {\small
    \setlength{\tabcolsep}{1pt}
	\begin{tabular}{
	>{\centering\arraybackslash}m{0.4cm} 
	>{\centering\arraybackslash}m{\figWidth} 
	>{\centering\arraybackslash}m{\figWidth}
	>{\centering\arraybackslash}m{\figWidth} 
	>{\centering\arraybackslash}m{\figWidth}
	>{\centering\arraybackslash}m{\figWidth}}
		& $\theta = 0.05 $
		& $\theta = 0.1 $
		& $\theta = 0.2 $
		& $\theta = 0.4 $
		& $\theta = 0.8 $
		\\\addlinespace[.3ex]

		\rotatebox{90}{\makecell{Depth map}}
		&\gframe{\includegraphics[width=\linewidth]{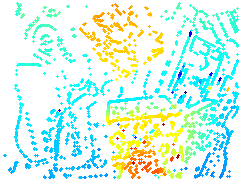}}
		&\gframe{\includegraphics[width=\linewidth]{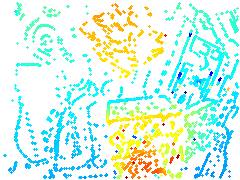}}
		&\gframe{\includegraphics[width=\linewidth]{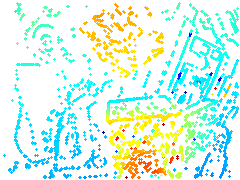}}
		&\gframe{\includegraphics[width=\linewidth]{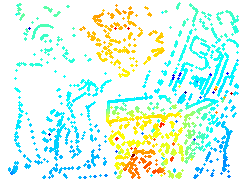}}
		&\gframe{\includegraphics[width=\linewidth]{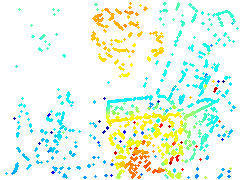}}
		\\
		
		\rotatebox{90}{\makecell{Confidence map}}
		&\gframe{\includegraphics[width=\linewidth]{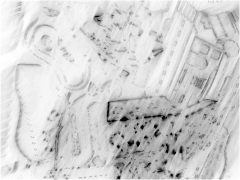}}
		&\gframe{\includegraphics[width=\linewidth]{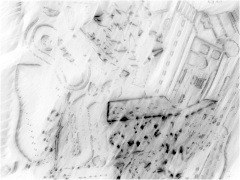}}
		&\gframe{\includegraphics[width=\linewidth]{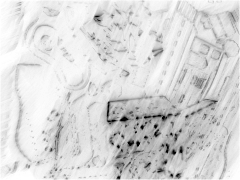}}
		&\gframe{\includegraphics[width=\linewidth]{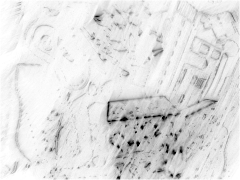}}
		&\gframe{\includegraphics[width=\linewidth]{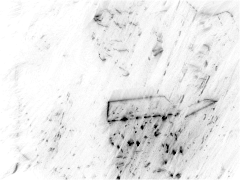}}
		\\

	\end{tabular}
	}
	\captionof{figure}{\label{fig:contrast_thresh}\emph{Effect of varying the camera's contrast threshold}. 
	Output of stereo Alg.~\ref{alg:fusion:stereo} on simulated events generated with varying contrast threshold $\theta$. 
	As $\theta$ increases, the sparsity of recovered points increases and the quality of depth errors decreases. 
	Depth maps are colored from red (near) to blue (far) in the range 1.3--4~\si{\meter}.
	}
\end{figure*}
\begin{figure*}[t] 
\def\imgWidth{0.245\linewidth}
\begin{subfigure}{.265\linewidth}
    \includegraphics[trim={0 0 13mm 13mm},clip,width=\linewidth]{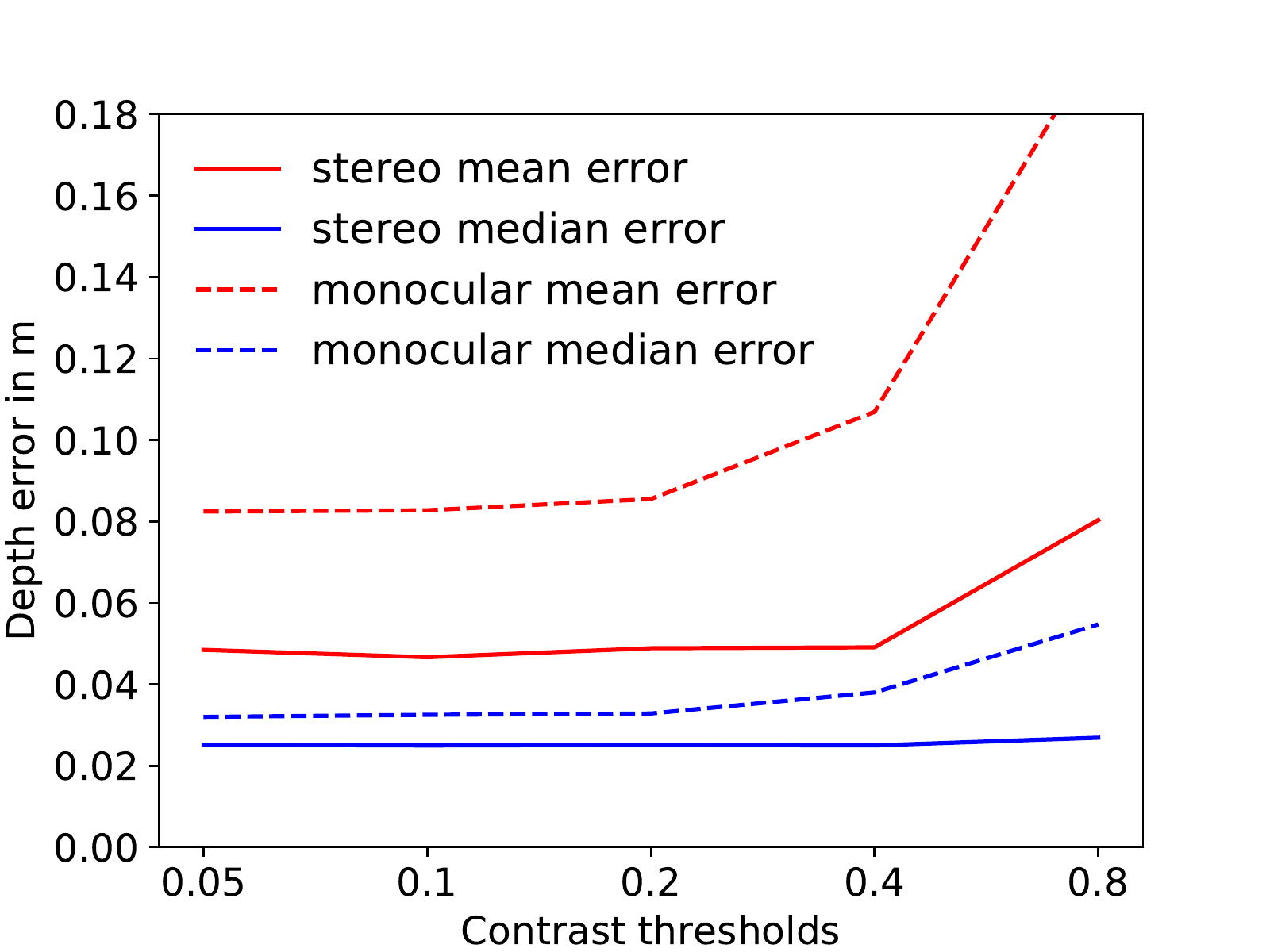}%
    \caption{Depth error vs. contrast thr.\label{fig:contrast:error}}
\end{subfigure}%
\begin{subfigure}{\imgWidth}
    \includegraphics[trim={6mm 0 13mm 7mm},clip,width=\linewidth]{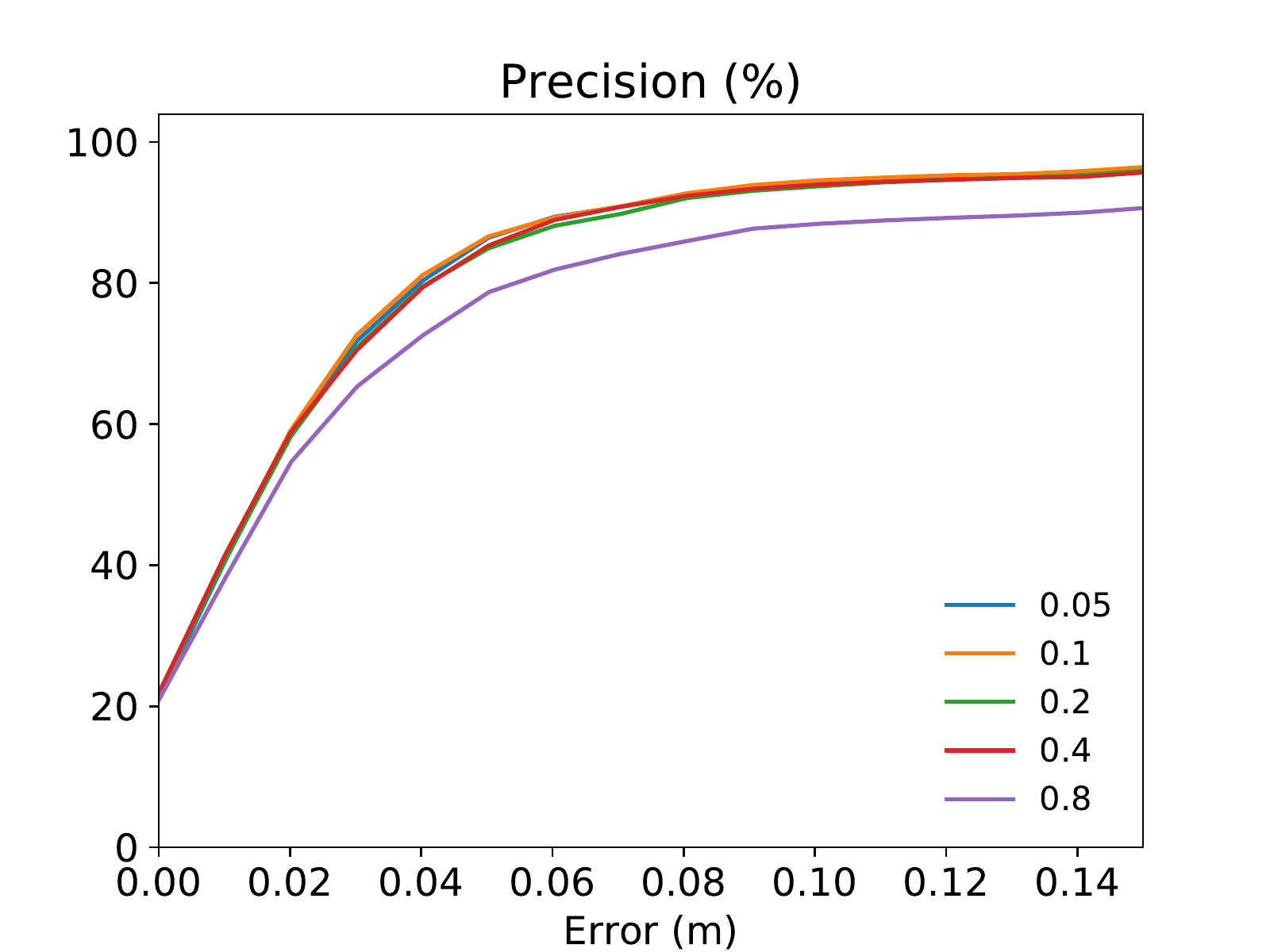}%
    \caption{Precision (\%)\label{fig:contrast:precision}}
\end{subfigure}%
\begin{subfigure}{\imgWidth}
    {\includegraphics[trim={6mm 0 13mm 7mm},clip,width=\linewidth]{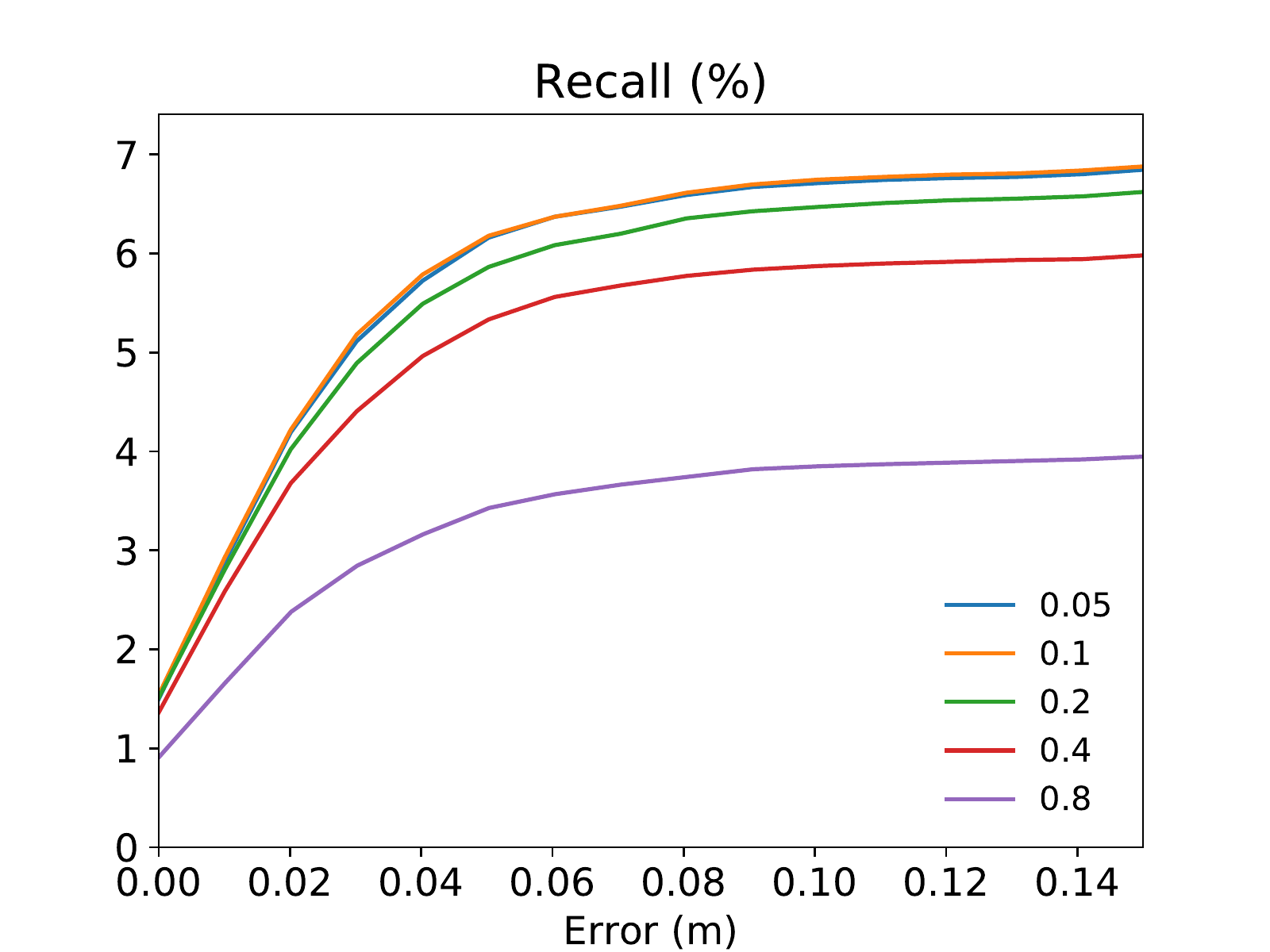}}%
    \caption{Recall (\%)\label{fig:contrast:recall}}
\end{subfigure}%
\begin{subfigure}{\imgWidth}
    \includegraphics[trim={6mm 0 13mm 7mm},clip,width=\linewidth]{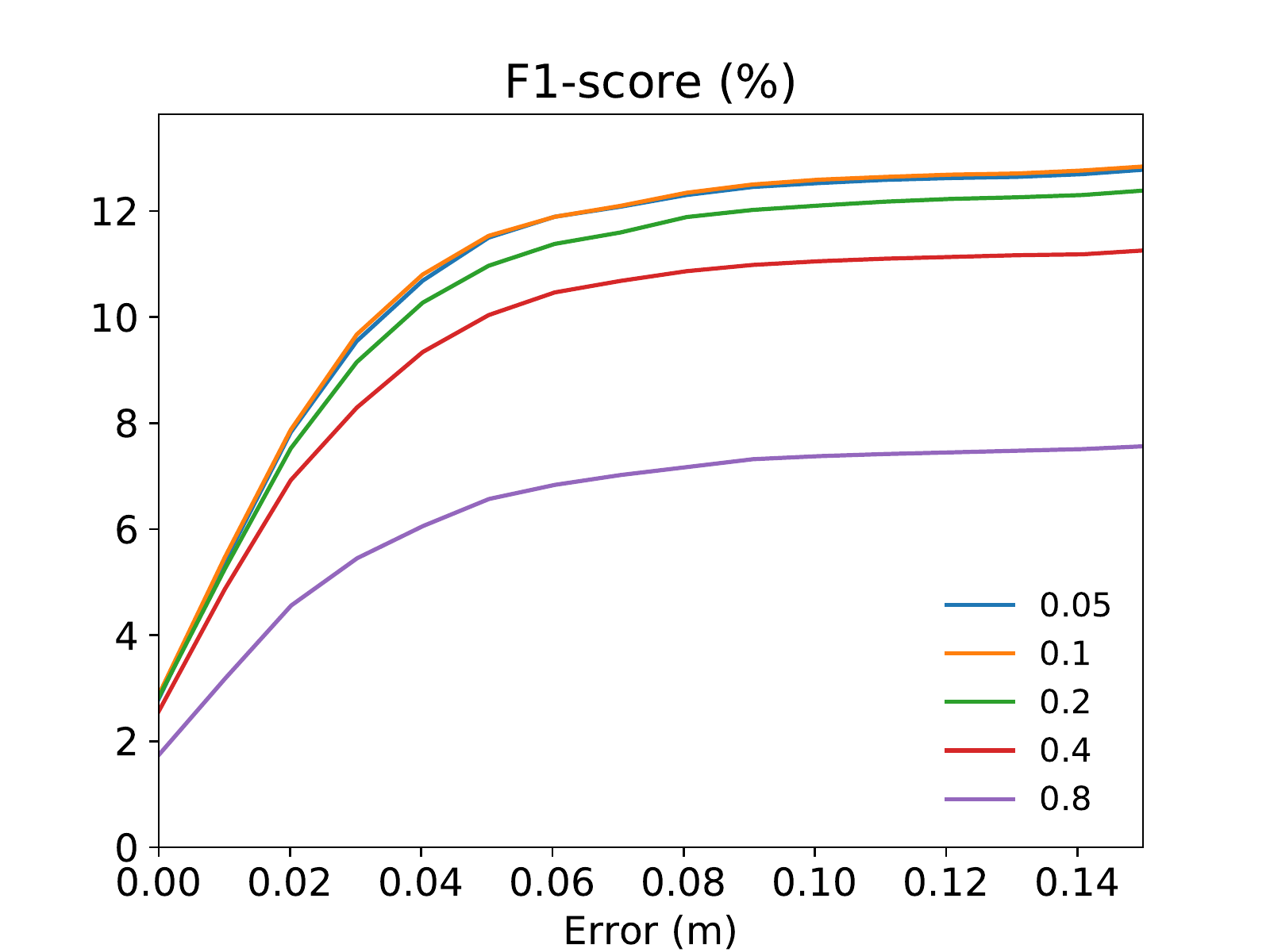}%
    \caption{F1-score (\%)\label{fig:contrast:fscore}}
\end{subfigure}
    \caption{\label{fig:contrast}
    (a) Sensitivity of depth errors with respect to the sensor's contrast threshold. 
    Maximum scene depth is \SI{2.7}{m}.
    (b)-(d) Precision, recall and F1-score for depth estimated using Alg.~\ref{alg:fusion:stereo} using events simulated with different contrast thresholds.
    }
\end{figure*}

%% file: chapters/04_limitations.tex
\subsection{Limitations}
\label{sec:limitations}

Our method assumes, like multi-view 3D reconstruction methods for standard cameras, accurate calibration and camera poses.
These assumptions allowed us to concentrate our efforts on investigating fusion functions for event-based 3D reconstruction under good multi-view alignment conditions.
Calibration is today of good quality using the DAVIS frames or image reconstruction~\cite{Muglikar21cvprw} if frames are not available.
In a full SLAM system, noise in the poses can propagate to the 3D reconstruction module. 
The TUM-VIE experiments, with different pose sources (mocap vs.~VIO algorithm), 
showed the robustness of our method to realistic poses (i.e., noisy as produced by a VIO algorithm).
This is encouraging, as future research on event-based camera tracking could achieve such accuracy and be combined with our stereo method. %
Recent results on the monocular case~\cite{Hidalgo22cvpr} suggests that such accuracy and robustness is achievable combining events and frames.

Another assumption of our method is that events are dominantly triggered by moving edges (brightness constancy). 
Hence, it may fail to estimate depth accurately from events that
are not due to motion, such as those caused by flickering lights.
Such events may be removed during pre-processing \cite{Wang22icra}.

%% file: chapters/05_conclusion.tex
\section{Conclusion}
\label{sec:conclusion}
We presented simple and effective state-of-the-art algorithms for event-based multi-camera 3D reconstruction in SLAM, combining across-camera and temporal fusion.
Thanks to the availability of accurate camera poses, matching within and across event cameras happens \emph{implicitly} in DSI space, which removes the need for \emph{event simultaneity} (explicit data association).
We investigated DSI space fusion methods, and showed how \emph{the same technique unifies temporal and camera fusion}. 
We proposed a spectrum of fusion functions, including summing event rays in multi-camera settings, and objectively analyzed the results they produced. 
Our theoretical design was supported by a comprehensive set of experiments:
testing on five established datasets and a simulator on a variety of scenarios, on millions of input events, while outperforming state of the art methods.
Fusion functions like the $H$-mean are beneficial for fusion because they have strong concavity, are bounded and smooth.
The effect of the parallax given by an additional camera is stronger than the effect of temporal fusion.
Our method works well regardless of the spatial resolution, which is interesting given the increasingly high spatial resolution of event cameras (see the comparisons with ESVO on DSEC and TUM-VIE datasets in the accompanying~video). 

Future research may look into combining the proposed method with an appropriately designed camera tracking algorithm to yield a full event-based stereo visual SLAM pipeline. 
The results on the driving datasets open the door to applying the proposed technique for creating HDR edge maps of a vehicle's surroundings and using it for later processing stages of Spatial AI and mobile intelligent systems \cite{Davison18arxiv}, such as spatial awareness or extraction of semantic information.

%% file: chapters/07_suppl_mat.tex
Tables \ref{tab:sota:mvsec:one} to \ref{tab:sota:mvsec:three} report depth estimation metrics on individual sequences of the MVSEC dataset.
\cref{tab:sota:mvsec} is the average of these per-sequence tables.

\input{floats/tab_sota_mvsec1}
\input{floats/tab_sota_mvsec2}

\input{floats/tab_sota_mvsec3}

%% file: floats/tab_sota_mvsec1.tex
\begin{table*}[t!]
\centering
\begin{adjustbox}{width=\textwidth}
\setlength{\tabcolsep}{3pt}
\begin{tabular}{ll*{1}{S[table-format=2.2,table-number-alignment=center]}
*{4}{S[table-format=1.2,table-number-alignment=center]}
*{4}{S[table-format=2.2,table-number-alignment=center]}
*{1}{S[table-format=1.2,table-number-alignment=center]}}
\toprule 
&Algorithm & \text{Mean Err} & \text{Median Err} & \text{bad-pix} & \text{SILog Err} & \text{AErrR} & \text{log RMSE} & \text{$\delta < 1.25$} & \text{$\delta < 1.25^2$} & \text{$\delta < 1.25^3$} & \text{\#Points}\\
& & \text{[cm] $\downarrow$} & \text{[cm] $\downarrow$} & \text{[\%] $\downarrow$} & \text{$\times 100 \downarrow$} & \text{[\%] $\downarrow$} & \text{$\times 100 \downarrow$} & \text{[\%] $\uparrow$} & \text{[\%] $\uparrow$} & \text{[\%] $\uparrow$} & \text{[million] $\!\uparrow$}\\
\midrule
\multirow{5}{*}{\begin{turn}{90}SOTA\end{turn}} 
& EMVS \cite{Rebecq18ijcv} (monocular)  & 39.37136238 & 14.94999894 & 3.048097665 & 4.721034606 & 13.25202463 & 22.10126025 & 82.02810969 & 93.43219923 & 97.61761158 & 1.21\\
& ESVO \cite{Zhou20tro} & 24.04372049 & 10.20585612 & 2.536969899 & 2.944309631 & 9.761736485 & 17.16740049 & 91.42858847 & 96.52733387 & 98.55429805 & 1.95\\
& ESVO indep. 1s    & 23.39492285 & 10.03367581 & 2.182036623 & 2.794700037 & 9.779762633 & 16.71744558 & 91.57270694 & 96.84331759 & 98.78664347 & 1.41\\
& SGM indep. 1s     & 35.45336664 & 13.61129454 & 5.539006607 & 7.348128796 & 15.0332788 & 27.46238437 & 85.96023654 & 93.50983372 & 96.40167787 & 11.64\\
& GTS indep. 1s     & 700.3735054 & 38.38886006 & 32.51423792 & 79.2616281 & 111.2145715 & 91.44298446 & 54.2713768 & 67.15918595 & 73.3880282 & 0.03\\
\midrule
\multirow{8}{*}{\begin{turn}{90}Ours\end{turn}} 
& $H_c\circ A_t$ (Alg~\ref{alg:fusion:stereo}) & 22.5255442 & 9.721720219 & 1.304463343 & 1.940232641 & 7.911660967 & 14.10815175 & 93.4944119 & 97.49894831 & 99.16836618 & 0.96\\
& $H_c\circ A_t$ (Alg~\ref{alg:fusion:stereo}) + MF & 23.33038256 & 9.89696104 & 1.392539834 & 2.079349574 & 8.124461973 & 14.61445869 & 93.15841335 & 97.28312028 & 99.05228516 & 3.48\\
& $H_c \circ H_t$   & 24.37908197 & 10.81239068 & 1.401730301 & 2.114942149 & 8.591000846 & 14.71321655 & 92.53297259 & 97.38247455 & 99.14749063 & 1.24\\
& $H_t \circ A_c$   & 23.78122167 & 10.36575082 & 1.263473183 & 2.025525259 & 8.359658651 & 14.40645251 & 92.72286954 & 97.53651008 & 99.20798216 & 1.02\\
& $A_c\circ H_t$    & 24.47753341 & 10.77572139 & 1.390238281 & 2.143336953 & 8.620342098 & 14.80962442 & 92.47754174 & 97.34322985 & 99.12938698 & 1.20\\
& $A_c \circ A_t$   & 21.93026528 & 9.351542038 & 1.220587726 & 1.873584337 & 7.752986464 & 13.84683879 & 93.6505991 & 97.6483503 & 99.23250318 & 0.87\\
& $A_t \circ H_c$   & 22.31653623 & 9.630297945 & 1.311233591 & 1.917541224 & 7.851942051 & 14.02090323 & 93.62255866 & 97.51074875 & 99.17440424 & 1.03\\
& $A_t \circ H_c$ + shuffling   & 23.91665966 & 10.60357964 & 1.277721909 & 2.008594962 & 8.403286363 & 14.36027092 & 92.75821546 & 97.55202714 & 99.21243775 & 1.11\\

\bottomrule
\end{tabular}
\end{adjustbox}
\caption{\label{tab:sota:mvsec:one}
Quantitative comparison of the proposed methods with the state of the art. 
\emph{MVSEC flying1}. See \cref{tab:sota:mvsec}.
}
\end{table*}

%% file: floats/tab_sota_mvsec2.tex
\begin{table*}[!ht]
\centering
\begin{adjustbox}{width=\textwidth}
\setlength{\tabcolsep}{3pt}
\begin{tabular}{ll*{1}{S[table-format=2.2,table-number-alignment=center]}
*{4}{S[table-format=1.2,table-number-alignment=center]}
*{4}{S[table-format=2.2,table-number-alignment=center]}
*{1}{S[table-format=1.2,table-number-alignment=center]}}
\toprule 
&Algorithm & \text{Mean Err} & \text{Median Err} & \text{bad-pix} & \text{SILog Err} & \text{AErrR} & \text{log RMSE} & \text{$\delta < 1.25$} & \text{$\delta < 1.25^2$} & \text{$\delta < 1.25^3$} & \text{\#Points}\\
& & \text{[cm] $\downarrow$} & \text{[cm] $\downarrow$} & \text{[\%] $\downarrow$} & \text{$\times 100 \downarrow$} & \text{[\%] $\downarrow$} & \text{$\times 100 \downarrow$} & \text{[\%] $\uparrow$} & \text{[\%] $\uparrow$} & \text{[\%] $\uparrow$} & \text{[million] $\!\uparrow$}\\
\midrule
\multirow{5}{*}{\begin{turn}{90}SOTA\end{turn}} 
& EMVS \cite{Rebecq18ijcv} (monocular)  & 31.41534863 & 13.01253973 & 6.152877683 & 4.561817977 & 13.373451 & 21.79860788 & 84.0673499 & 94.71703438 & 97.87541752 & 1.17\\
& ESVO \cite{Zhou20tro} & 21.3370895 & 8.96599798 & 3.74800638 & 3.482257075 & 9.323901439 & 19.13798664 & 91.60358453 & 95.8756925 & 97.86020885 & 1.89\\
& ESVO indep. 1s    & 20.41544047 & 8.625230618 & 3.499327255 & 3.235895456 & 9.140796331 & 18.35002731 & 92.0280127 & 96.18976373 & 98.19458821 & 1.41\\
& SGM indep. 1s     & 32.93834739 & 8.745162766 & 8.288065659 & 9.498117844 & 15.81916239 & 31.5404858 & 84.40149395 & 92.33314752 & 95.48485172 & 16.95\\
& GTS indep. 1s     & 167.1425728 & 37.22868023 & 43.07730704 & 71.91434618 & 94.77958401 & 86.92928268 & 49.35577463 & 60.54314594 & 67.76221376 & 0.07\\
\midrule
\multirow{8}{*}{\begin{turn}{90}Ours\end{turn}} 
& $H_c\circ A_t$ (Alg~\ref{alg:fusion:stereo}) & 18.20081727 & 8.487537573 & 1.767505319 & 1.784137166 & 8.128384933 & 13.58919896 & 95.52695413 & 98.13105919 & 99.08027501 & 0.65\\
& $H_c\circ A_t$ (Alg~\ref{alg:fusion:stereo}) + MF & 18.57703625 & 8.675972496 & 1.862961766 & 1.808397279 & 8.186908143 & 13.70624079 & 95.26650388 & 98.06730566 & 99.09251827 & 2.42\\
& $H_c \circ H_t$   & 22.47482687 & 10.12687622 & 2.9196898 & 2.459121792 & 9.41165269 & 16.07012978 & 92.66168637 & 97.19262052 & 98.7934966 & 1.39\\
& $H_t \circ A_c$   & 21.36712247 & 9.855306623 & 2.502318527 & 2.260657136 & 9.132095046 & 15.41369604 & 93.41386915 & 97.59878885 & 98.91870291 & 1.18\\
& $A_c\circ H_t$    & 22.28284883 & 10.06370769 & 2.808758982 & 2.400553561 & 9.354631066 & 15.87596563 & 92.80287894 & 97.27508457 & 98.82430615 & 1.34\\
& $A_c \circ A_t$   & 19.04319486 & 8.760938524 & 2.243082528 & 1.990532089 & 8.437067267 & 14.39671124 & 94.68495396 & 97.88781548 & 99.01863174 & 1.01\\
& $A_t \circ H_c$   & 19.82199248 & 8.910974836 & 2.549009317 & 2.085586845 & 8.600561075 & 14.7699477 & 94.1624981 & 97.55107249 & 98.94108747 & 1.17\\
& $A_t \circ H_c$ + shuffling   & 21.659804 & 9.907930607 & 2.599652784 & 2.219718231 & 9.15884291 & 15.27316034 & 93.13517369 & 97.53063926 & 98.92019183 & 1.22\\
\bottomrule

\end{tabular}
\end{adjustbox}
\caption{\label{tab:sota:mvsec:two}
Quantitative comparison of the proposed methods with the state of the art. 
\emph{MVSEC flying2}. See \cref{tab:sota:mvsec}.
}
\end{table*}

%% file: floats/tab_sota_mvsec3.tex
\begin{table*}[!ht]
\centering
\begin{adjustbox}{width=\textwidth}
\setlength{\tabcolsep}{3pt}
\begin{tabular}{ll*{1}{S[table-format=2.2,table-number-alignment=center]}
*{4}{S[table-format=1.2,table-number-alignment=center]}
*{4}{S[table-format=2.2,table-number-alignment=center]}
*{1}{S[table-format=1.2,table-number-alignment=center]}}
\toprule 
&Algorithm & \text{Mean Err} & \text{Median Err} & \text{bad-pix} & \text{SILog Err} & \text{AErrR} & \text{log RMSE} & \text{$\delta < 1.25$} & \text{$\delta < 1.25^2$} & \text{$\delta < 1.25^3$} & \text{\#Points}\\
& & \text{[cm] $\downarrow$} & \text{[cm] $\downarrow$} & \text{[\%] $\downarrow$} & \text{$\times 100 \downarrow$} & \text{[\%] $\downarrow$} & \text{$\times 100 \downarrow$} & \text{[\%] $\uparrow$} & \text{[\%] $\uparrow$} & \text{[\%] $\uparrow$} & \text{[million] $\!\uparrow$}\\
\midrule
\multirow{5}{*}{\begin{turn}{90}SOTA\end{turn}} 
& EMVS \cite{Rebecq18ijcv} (monocular)  & 30.53881128 & 15.08932186 & 2.31357303 & 3.330249026 & 11.59026086 & 18.26740209 & 88.15789028 & 96.45067328 & 98.47499669 & 1.42\\
 
& ESVO \cite{Zhou20tro} &  29.62368052 & 12.61075504 & 3.778104891 & 4.024672842 & 11.49599765 & 20.19621934 & 88.28025413 & 94.8847182 & 97.51792733 & 2.29\\
 
& ESVO indep. 1s    &  24.29024964 & 10.84031385 & 2.813125333 & 3.049568841 & 9.836208376 & 17.53676678 & 91.86569595 & 96.46226415 & 98.16058419 & 1.86\\
 
& SGM indep. 1s     & 37.85577042 & 14.68797862 & 5.330375467 & 8.515982608 & 17.65201827 & 29.45758805 & 85.671693 & 93.31468235 & 96.21100682 & 14.81\\

& GTS indep. 1s & 299.4824869 & 60.66447855 & 39.75406171 & 72.24124387 & 102.7739061 & 88.866648 & 45.03800865 & 58.86272172 & 66.9384409 & 0.08\\
\midrule
\multirow{8}{*}{\begin{turn}{90}Ours\end{turn}} 

& $H_c\circ A_t$ (Alg~\ref{alg:fusion:stereo}) &  19.48989113 & 10.3834481 & 0.9878361933 & 1.431991574 & 7.347378923 & 12.00940611 & 96.08606665 & 98.59641666 & 99.38336662 & 0.82\\

& $H_c\circ A_t$ (Alg~\ref{alg:fusion:stereo}) + MF &20.02439611 & 10.58983243 & 1.023597876 & 1.502869295 & 7.498733122 & 12.2994767 & 95.78721581 & 98.50569848 & 99.36128349 & 3.11\\
 
& $H_c \circ H_t$   &  23.48281337 & 11.98642343 & 1.358318329 & 1.950949877 & 8.583740318 & 14.01311876 & 93.79053529 & 97.90448919 & 99.1410937 & 1.79\\

& $H_t \circ A_c$   & 22.67851744 & 11.82609539 & 1.244020313 & 1.810642569 & 8.349989384 & 13.49578522 & 94.33166309 & 98.13887751 & 99.23956939 & 1.55\\
   
& $A_c\circ H_t$    & 23.2916984 & 11.96870439 & 1.317997954 & 1.907771235 & 8.524299554 & 13.85607342 & 93.93615822 & 97.97410893 & 99.17004442 & 1.72\\
 
 &  $A_c \circ A_t$ & 20.17370508 & 10.68020706 & 1.061889237 & 1.527618354 & 7.589141025 & 12.39750585 & 95.66516001 & 98.47913924 & 99.35894362 & 1.09\\
 
& $A_t \circ H_c$   & 20.60883243 & 10.73634685 & 1.125799117 & 1.586391531 & 7.684260763 & 12.64440855 & 95.39913235 & 98.3471855 & 99.30471363 & 1.23\\
 
& $A_t \circ H_c$ + shuffling   & 22.23696784 & 11.60359773 & 1.174945187 & 1.730918105 & 8.171904232 & 13.20019073 & 94.57741868 & 98.20642482 & 99.2650945 & 1.39\\

\bottomrule
\end{tabular}
\end{adjustbox}
\caption{\label{tab:sota:mvsec:three}
Quantitative comparison of the proposed methods with the state of the art. 
\emph{MVSEC flying3}. See \cref{tab:sota:mvsec}.
}
\end{table*}

%% file: main.bbl

%% file: chapters/biographies.tex
\begin{IEEEbiography}[{\includegraphics[width=1in,height=1.25in,clip,keepaspectratio]{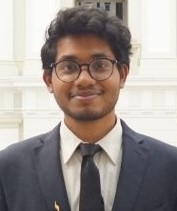}}]{Suman Ghosh} 
is pursuing doctoral studies in the Department of Electrical Engineering and Computer Science at Technische Universit\"at Berlin, Germany. Before that, he worked as Research Fellow at the EDPR group at Istituto Italiano di Tecnologia, Italy. He obtained his Masters degree on Robotics in 2019 from the University of Genoa, Italy, and Warsaw University of Technology, Poland, with full Erasmus Mundus scholarship under the European Master on Advanced Robotics (EMARO+) program. In 2016, he received a Bachelors degree in Electronics and Telecommunication Engineering from Jadavpur University, India. His research interests include (biological and computer) vision, robotics and learning.
\end{IEEEbiography}

\begin{IEEEbiography}[{\includegraphics[width=1in,height=1.25in,clip,keepaspectratio]{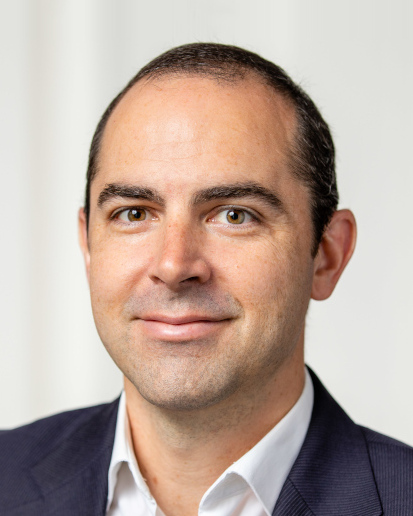}}]{Guillermo Gallego} is Associate Professor at Technische Universit\"at Berlin, Berlin, Germany, in the Dept. of Electrical Engineering and Computer Science, and at the Einstein Center Digital Future, Berlin, Germany.
He is also a Principal Investigator at the Science of Intelligence Excellence Cluster, Berlin, Germany.
He received the PhD degree in Electrical and Computer Engineering from the Georgia Institute of Technology, USA, in 2011, supported by a Fulbright Scholarship.
From 2011 to 2014 he was a Marie Curie researcher with Universidad Politecnica de Madrid, Spain, and from 2014 to 2019 he was a postdoctoral researcher at the Robotics and Perception Group, University of Zurich and ETH Zurich, Switzerland.
He serves as Associate Editor for IEEE Transactions on Pattern Analysis and Machine Intelligence, and for IEEE Robotics and Automation Letters.
His research interests include robotics, computer vision, signal processing, optimization and geometry. 
\end{IEEEbiography}